\documentclass{egpubl}
\usepackage{eg2026}
 
\ConferencePaper        %

\CGFStandardLicense

\usepackage[T1]{fontenc}
\usepackage{dfadobe}  

\biberVersion
\BibtexOrBiblatex
\usepackage[backend=biber,bibstyle=EG,citestyle=alphabetic,backref=true]{biblatex} 
\electronicVersion
\PrintedOrElectronic

\ifpdf \usepackage[pdftex]{graphicx} \pdfcompresslevel=9
\else \usepackage[dvips]{graphicx} \fi

\usepackage{egweblnk}

\title[GTAvatar: Bridging Gaussian Splatting and Texture Mapping for Relightable and Editable Gaussian Avatars]%
      {GTAvatar: Bridging Gaussian Splatting and Texture Mapping for Relightable and Editable Gaussian Avatars}

\author[K. Baert et al.]{
    \parbox{\textwidth}{
        \centering
        Kelian Baert\orcid{0000-0003-1732-4653} \quad
        Mae Younes\orcid{0000-0002-4831-3343} \quad
        Francois Bourel\orcid{0009-0005-3819-735X} \quad
        Marc Christie\orcid{0000-0001-6080-8026} \quad
        Adnane Boukhayma\orcid{0000-0003-0529-7598}
    }
    \\%
    \parbox{\textwidth}{
        \centering
        Univ Rennes, Inria, CNRS, IRISA, France
    }
}

\usepackage{xcolor}
\usepackage{xspace}
\usepackage{multirow}
\usepackage{amsmath}
\usepackage{amssymb}
\usepackage{graphicx}
\usepackage[export]{adjustbox}
\usepackage[table, dvipsnames]{xcolor} %
\usepackage[caption=false]{subfig}
\usepackage{booktabs}
\usepackage{float} %
\usepackage[normalem]{ulem} %

\usepackage{siunitx} %
\newcommand{\ie}{\textsl{i.e.}~}
\newcommand{\eg}{\textsl{e.g.}~}

\newcommand{\cf}{\textsl{cf.}~}

\newcommand{\uv}{uv}

\newcommand{\Jst}{\ensuremath{J_{\text{st}}}}
\newcommand{\Jtri}{\ensuremath{J_{\text{v}}}}
\newcommand{\Jtriinv}{\ensuremath{\Jtri^{\dagger}}}
\newcommand{\Juv}{\ensuremath{J_{\text{uv}}}}
\newcommand{\Jstuv}{\ensuremath{J_{\text{st}\to\text{\uv}}}}

\newcommand{\uvzero}{\ensuremath{\text{uv}_0}}

\newcommand{\rankone}[1]{\cellcolor{orange!30}#1}
\newcommand{\ranktwo}[1]{\cellcolor{yellow!30}#1}

\begin{document}

\teaser{
    \includegraphics[width=\linewidth]{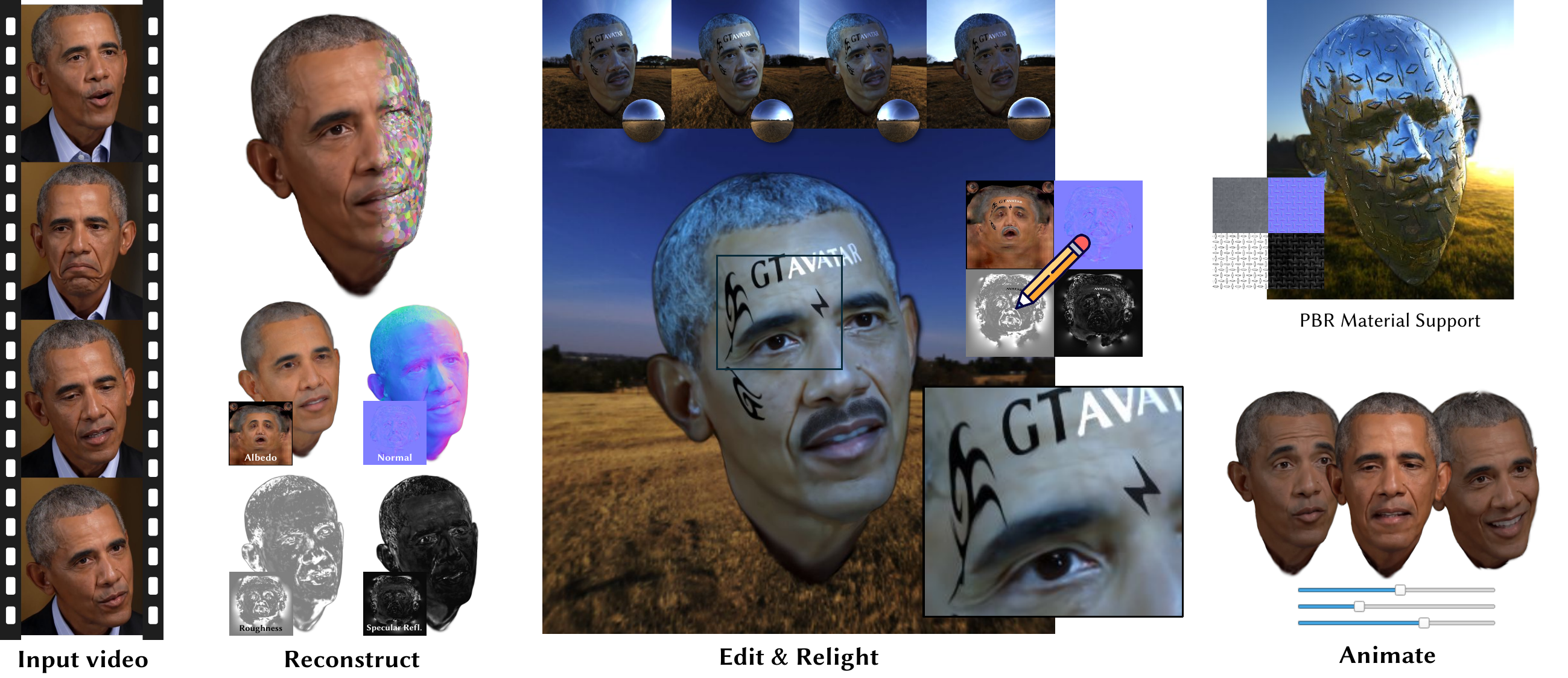}
    \centering
    \caption{\textbf{GTAvatar} broadens applications of monocular Gaussian Splatting head avatars beyond reenactment and relighting, enabling interactive editing of textures for precise control of intrinsic appearance, while preserving training efficiency, rendering speed and visual fidelity. Project page: \url{https://kelianb.github.io/GTAvatar/}
    }
    \label{fig:teaser}
}

\maketitle
\begin{abstract}
Recent advancements in Gaussian Splatting have enabled increasingly accurate reconstruction of photorealistic head avatars, opening the door to numerous applications in visual effects, videoconferencing, and virtual reality. This, however, comes with the lack of intuitive editability offered by traditional triangle mesh-based methods. In contrast, we propose a method that combines the accuracy and fidelity of 2D Gaussian Splatting with the intuitiveness of UV texture mapping. By embedding each canonical Gaussian primitive’s local frame into a patch in the UV space of a template mesh in a computationally efficient manner, we reconstruct continuous editable material head textures from a single monocular video on a conventional UV domain. Furthermore, we leverage an efficient physically based reflectance model to enable relighting and editing of these intrinsic material maps. Through extensive comparisons with state-of-the-art methods, we demonstrate the accuracy of our reconstructions, the quality of our relighting results, and the ability to provide intuitive controls for modifying an avatar’s appearance and geometry via texture mapping without additional optimization.

\printccsdesc   
\end{abstract}

\section{Introduction}

Gaussian Splatting based avatars have revolutionized the capture and rendering of digital humans by delivering an unprecedented level of photorealism with real-time rendering capability, enabling new possibilities for reanimating content from video inputs. However, this realism comes with a trade-off: unlike traditional texture-map-based modeling, Gaussian avatars offer little flexibility for intuitive appearance editing. This gap becomes critical in practice. In film and visual effects, artists routinely adjust the smallest details of a face -- smoothing skin to create a flawless appearance, removing distracting high-frequency features, or sculpting age and character through wrinkles, scars and bruises. In gaming and virtual production, creators seek the same level of control to personalize avatars with tattoos, makeup, or stylized patterns using artist-friendly tools that operate directly on material texture maps. Without such editing capabilities, even the most realistic avatar remains a fixed reproduction, rather than a medium for creative expression.

This limitation stems from the fact that each Gaussian splat carries its own color or material properties in isolation, without the shared structure that texture maps naturally provide. As a result, the surface lacks the local coherence needed to treat it as a continuous canvas, making it difficult -- if not impossible -- to apply edits consistently, whether to a small region of the face or to its overall appearance. Conversely, while mesh-based 3D inverse rendering methods~\cite{nvdiffrec,flare,spark} inherit a coherent defacto UV-domain fit for artist-friendly edits, they can suffer from topological rigidity when representing high-frequency details and are hampered in their reconstruction capacity due to the limitations of differentiable rasterization when handling complex or translucent geometries (\cf Table~\ref{tab:exp-reconstruction}). Furthermore, naively embedding Gaussians in UV domain leads to discontinuous texture maps that are hard to edit successfully, as witnessed in our results (\eg Figure~\ref{fig:exp-ablation-texturing}) and also in~\cite{fate}. The authors of FATE~\cite{fate} proposed a second U-Net neural baking stage to alleviate this issue in the context of non-relightable avatars. 

Our goal is to reconstruct a head avatar from a single monocular RGB video, that can be rendered in real time, relit under arbitrary environment maps, animated with new poses and expressions, and directly edited through texture mapping, yet without the added postprocessing complexity of training a two-stage model or baking lighting information in texture directly.

Following seminal work~\cite{gaussianavatars,surfhead}, we adopt FLAME-anchored~\cite{FLAME} Gaussian splats.
Our key observation is that, unlike vanilla 3DGS~\cite{3DGS}, which operates directly in screen space, the 2DGS variant~\cite{2DGS} defines a local splat coordinate frame and computes ray-splat intersections to query kernels. We leverage this property to map ray-splat intersections into the UV domain via an approximate orthographic projection from the splat tangent space to its corresponding FLAME mesh triangle, effectively mapping each splat plane to a continuous UV patch. We devise an efficient method to compute this mapping using a single matrix multiplication per intersection, which is highly important to maintain the method's efficiency given the very high number of intersections involved in rendering. PBR material attributes are then sampled smoothly from learnable texture maps (albedo, roughness, and specular reflectance) using bilinear filtering. The splat orientations are combined with a residual normal map to produce the final shading normals. We jointly optimize the FLAME parameters, texture maps, 2D Gaussian positional parameters and environment lighting. This enables differentiable expression of outgoing radiance via the Cook-Torrance BRDF \cite{cook1982reflectance} under the split-sum approximation \cite{karis2013real}, using splatted geometry and material G-buffers in a deferred shading framework. Finally, we introduce a novel UV regularization that is crucial for maintaining representation integrity by enforcing alignment of UV coordinates of ray-splat intersections for a given ray. Figure \ref{fig:exp-ablation-texturing} underlines the benefits of our UV mapping and regularization compared to a naive UV embedding strategy based on the projection of splat origins.

\begin{figure*}[h]
    \newcommand{\w}{0.15\linewidth}
    \setlength{\tabcolsep}{0em} %
    \def\arraystretch{0.8}{ %
    \begin{tabular}{c@{\hskip 0.05in}ccc@{\hskip 0.05in}c}
        Reconstruction & \multicolumn{3}{c}{Edit} & Albedo texture
        \\
        \includegraphics[width=\w,valign=c]{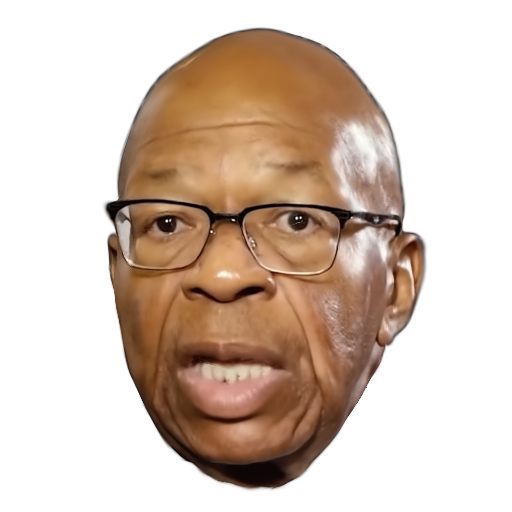} &
        \includegraphics[width=\w,valign=c]{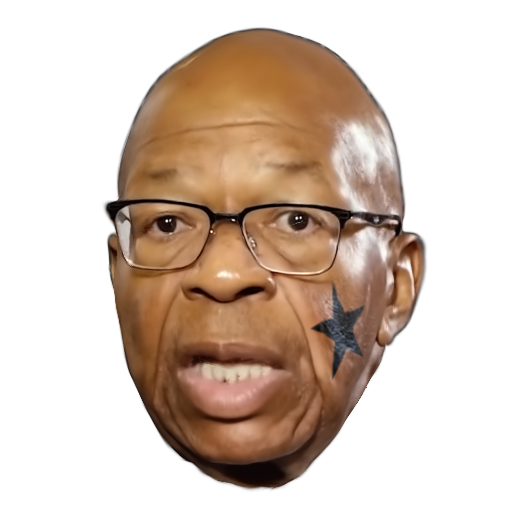} &
        \includegraphics[width=\w,valign=c]{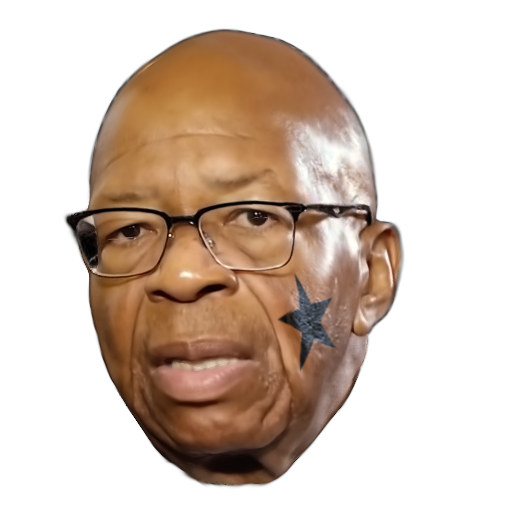} &
        \includegraphics[width=\w,valign=c]{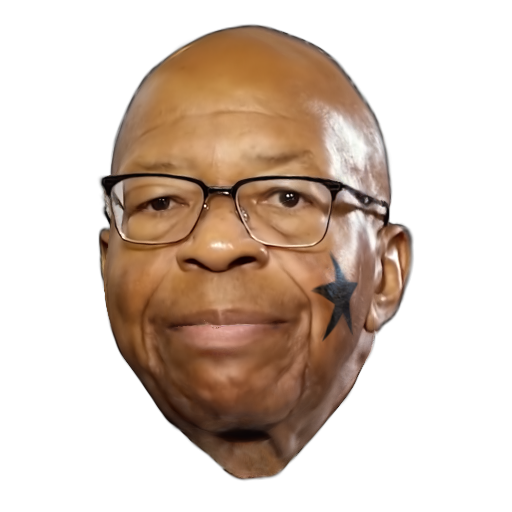} &
        \includegraphics[width=\w,valign=c]{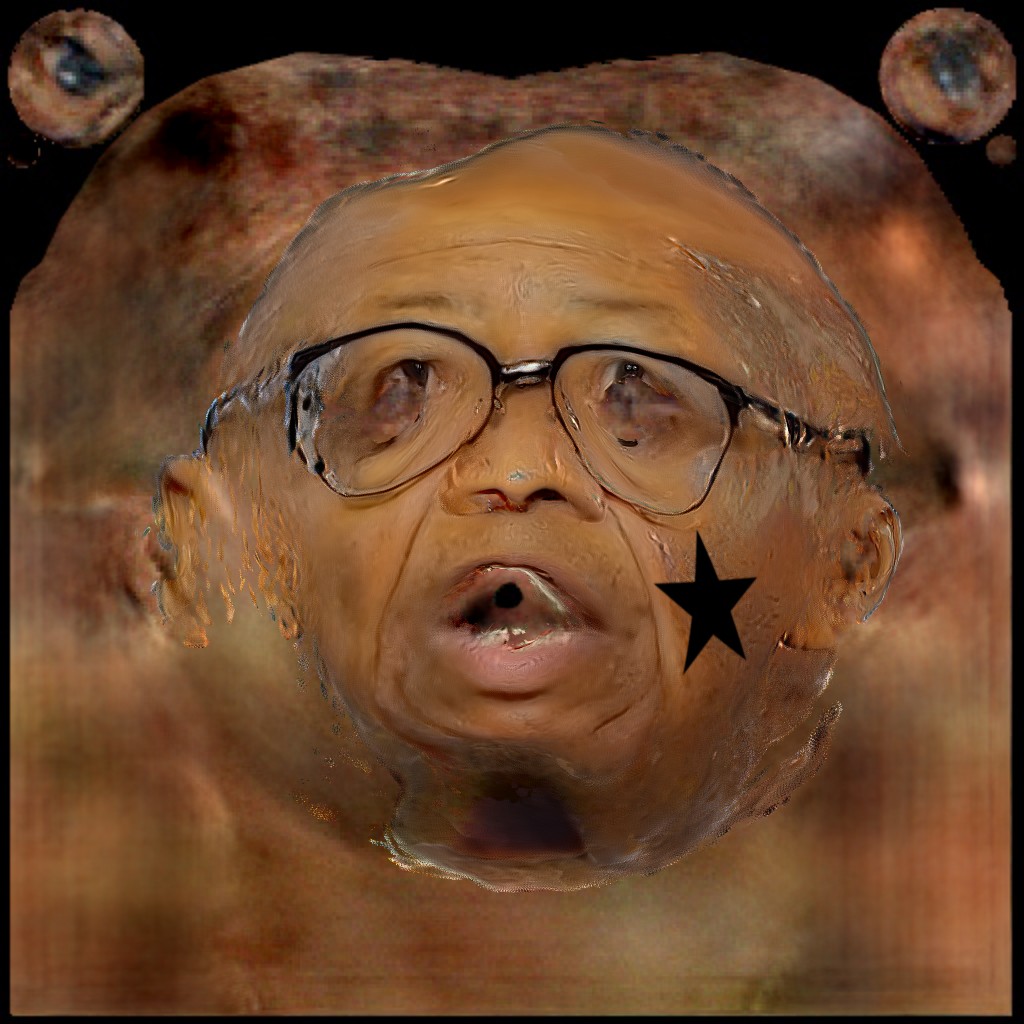} 
        \\
        \includegraphics[width=\w,valign=c]{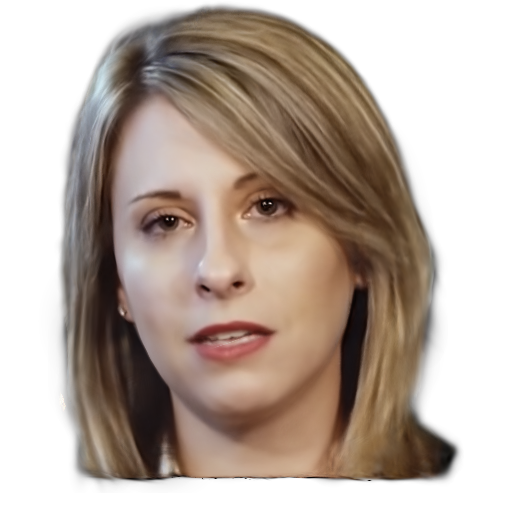} &
        \includegraphics[width=\w,valign=c]{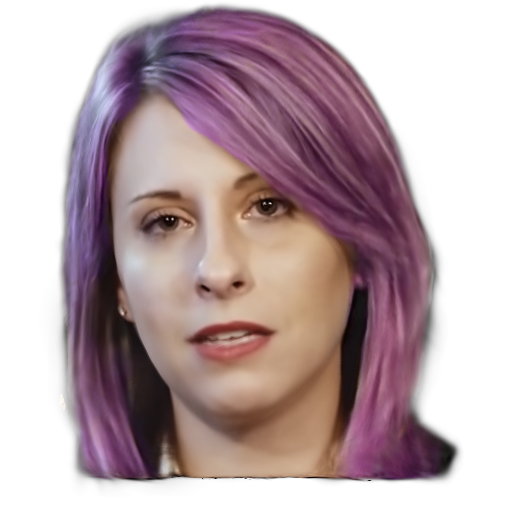} &
        \includegraphics[width=\w,valign=c]{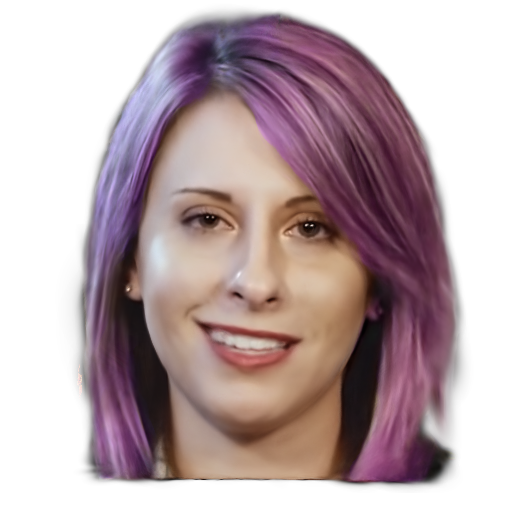} &
        \includegraphics[width=\w,valign=c]{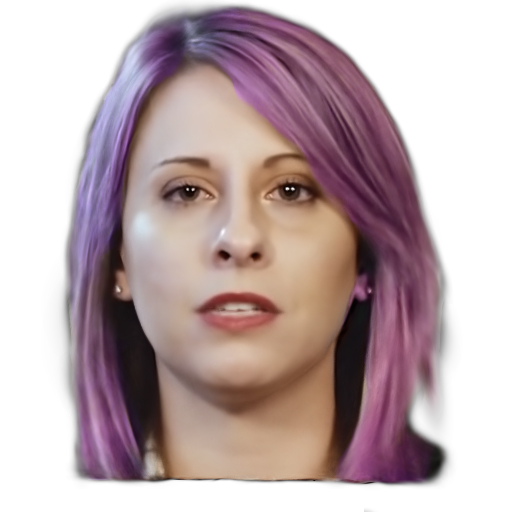} &
        \includegraphics[width=\w,valign=c]{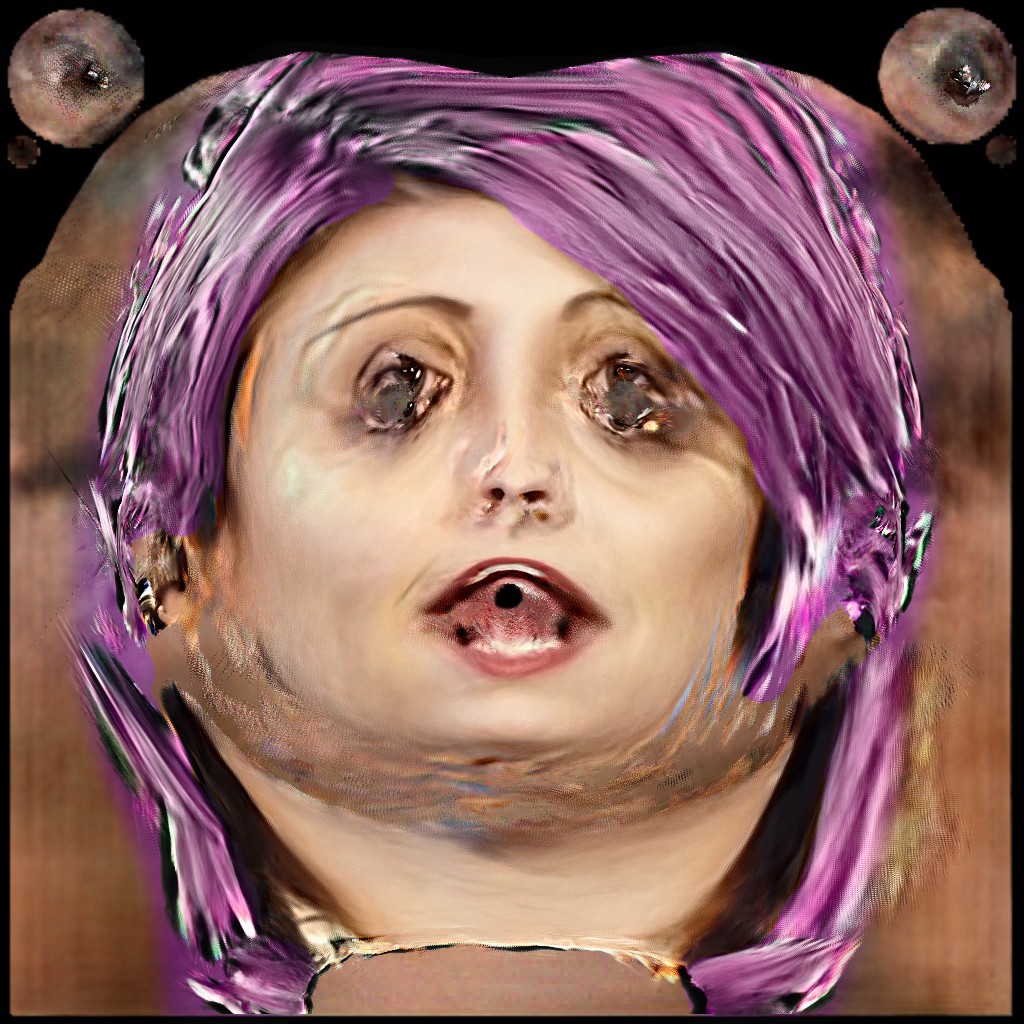}
        \\
        \includegraphics[width=\w,valign=c]{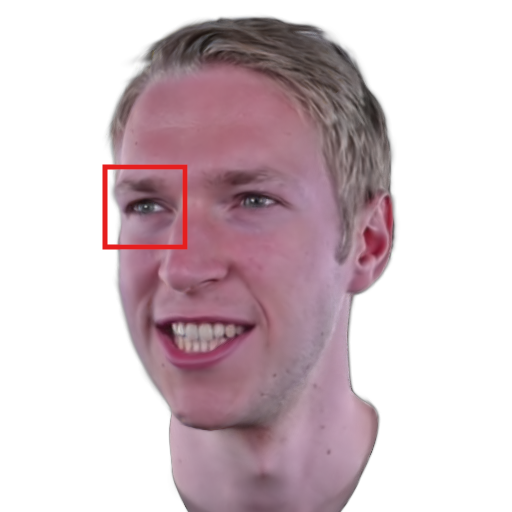} &
        \includegraphics[width=\w,valign=c]{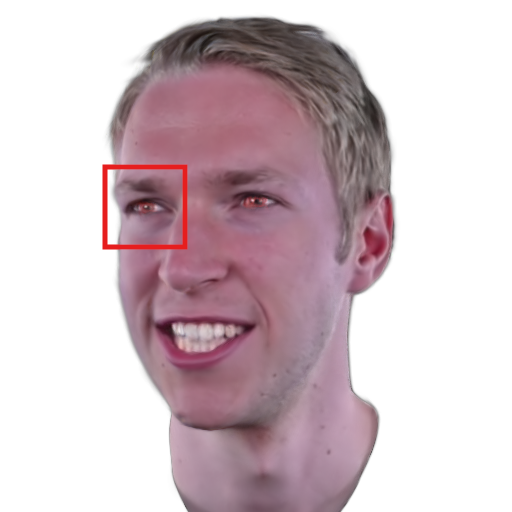} &
        \includegraphics[width=\w,valign=c]{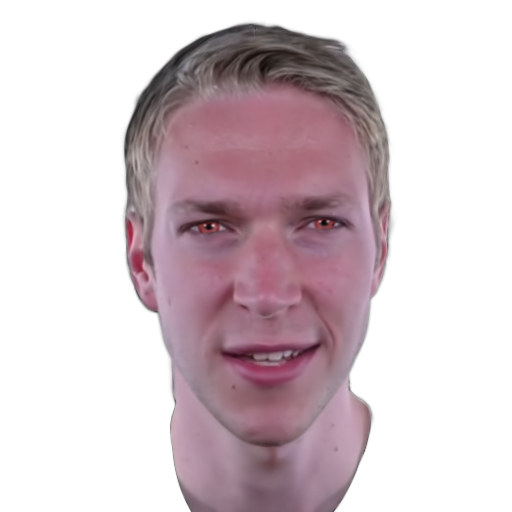} &
        \includegraphics[width=\w,valign=c]{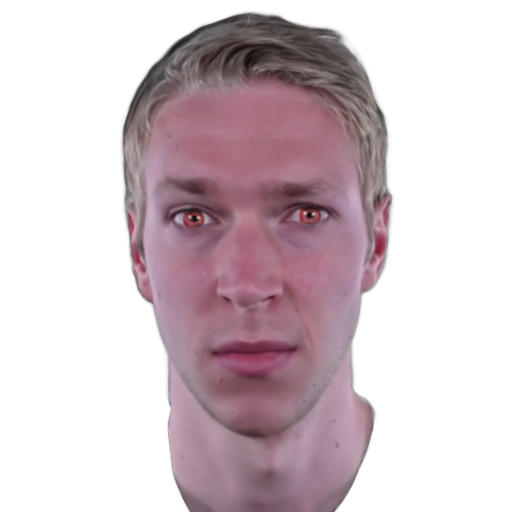} &
        \includegraphics[width=\w,valign=c]{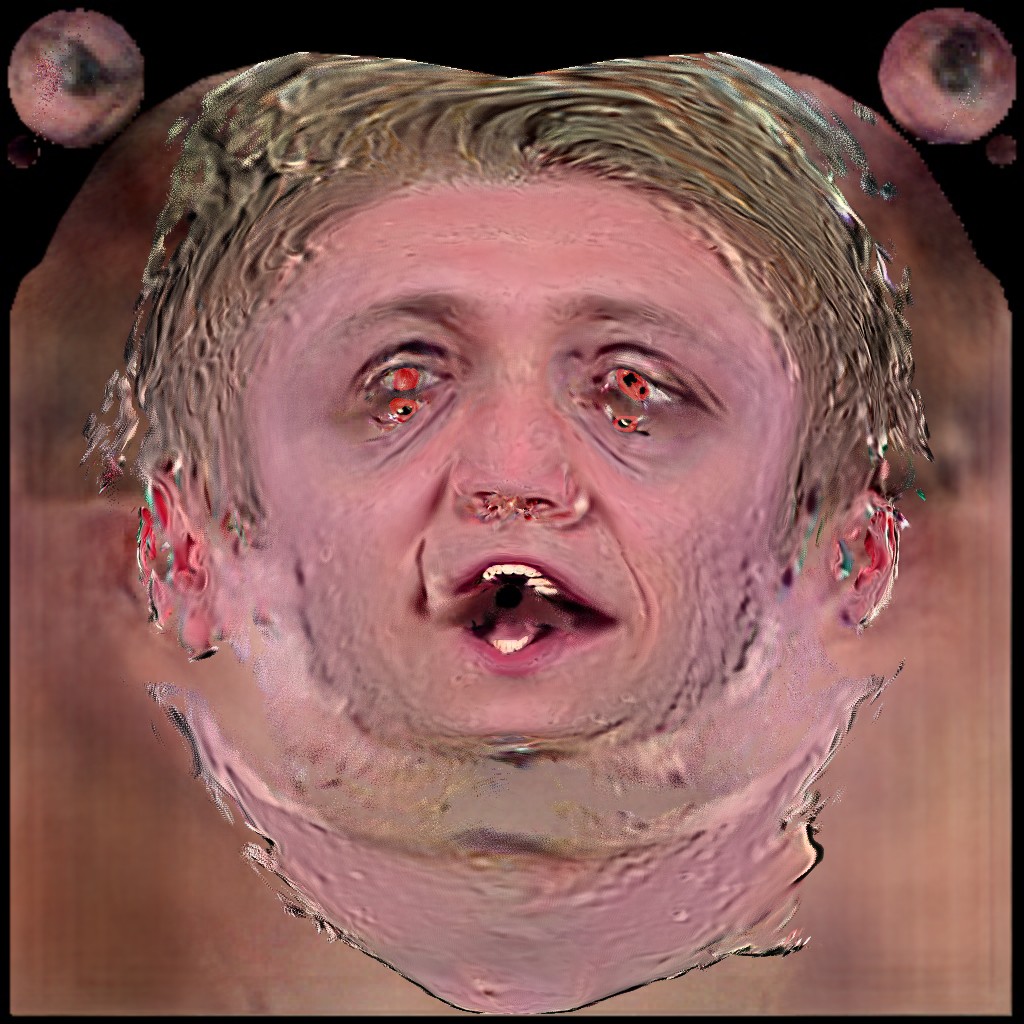}
        \\
        \includegraphics[width=\w,valign=c]{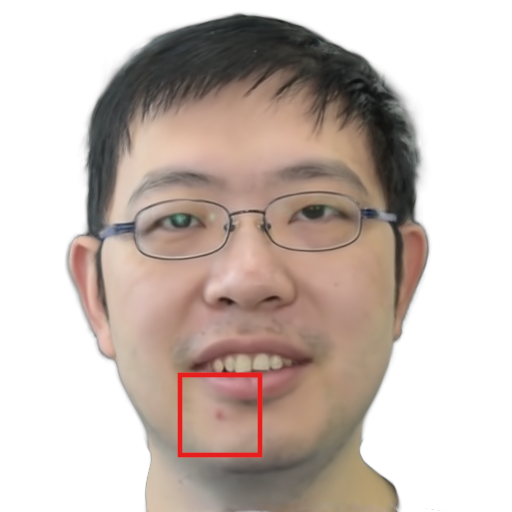} &
        \includegraphics[width=\w,valign=c]{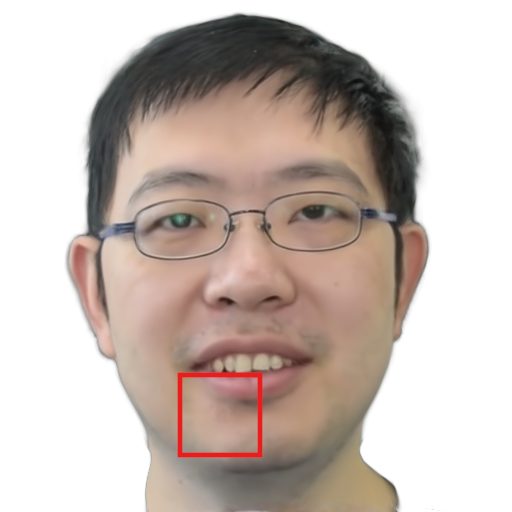} &
        \includegraphics[width=\w,valign=c]{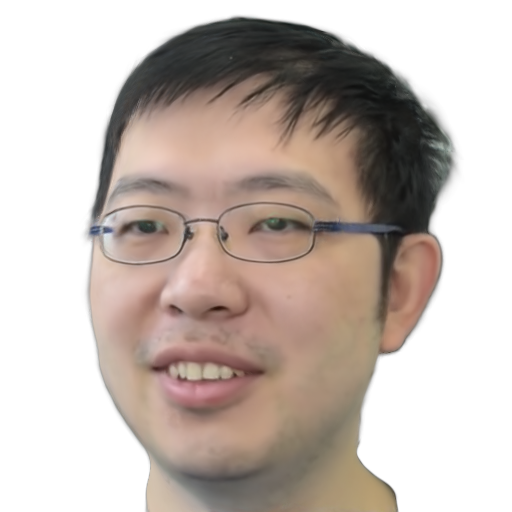} &
        \includegraphics[width=\w,valign=c]{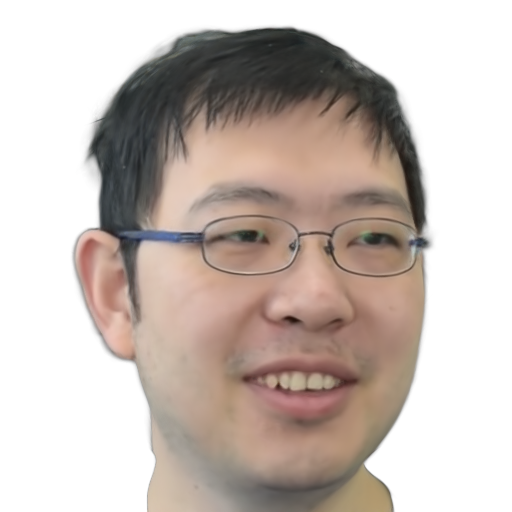} &
        \includegraphics[width=\w,valign=c]{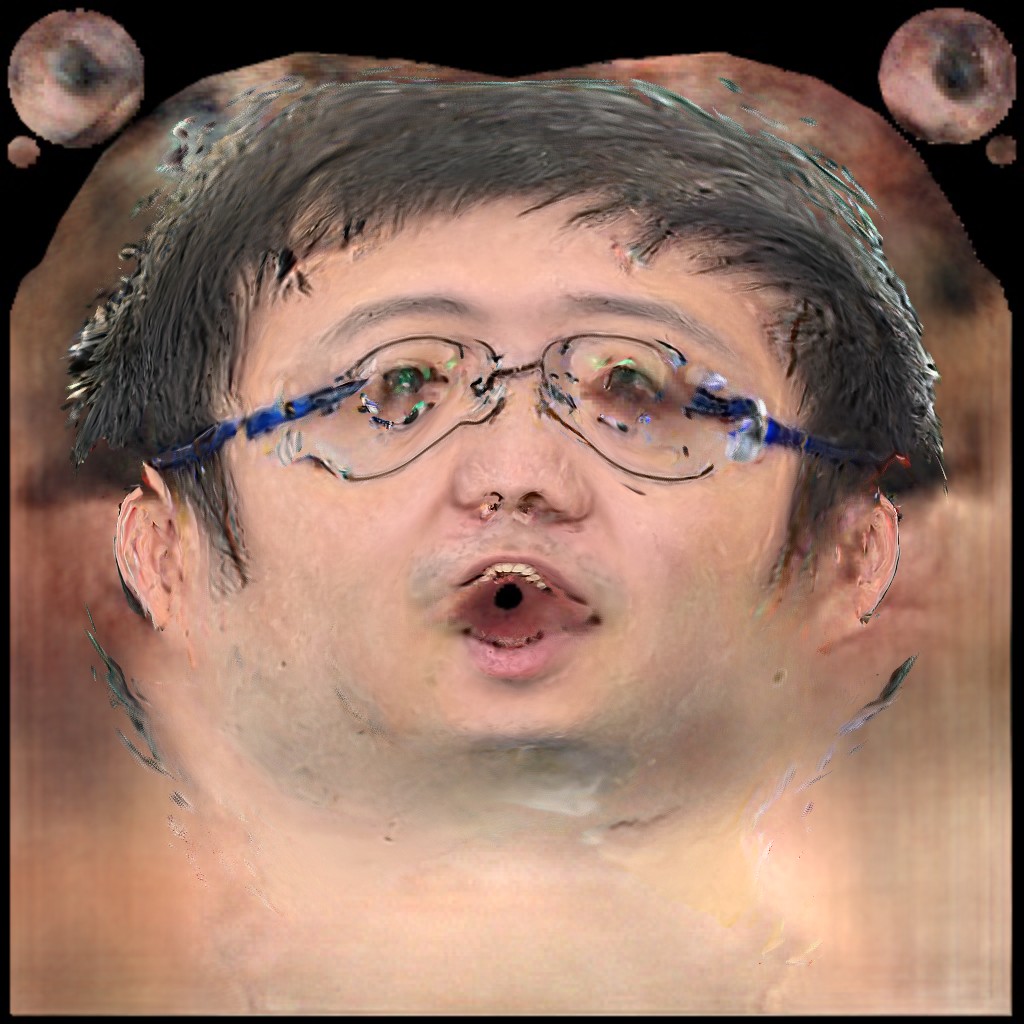}
        \\
        \includegraphics[width=\w,valign=c]{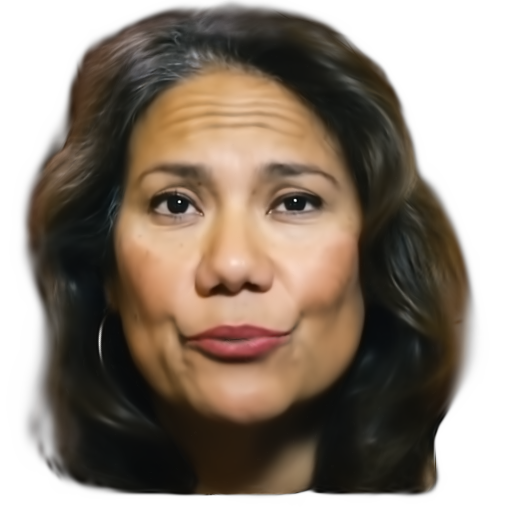} &
        \includegraphics[width=\w,valign=c]{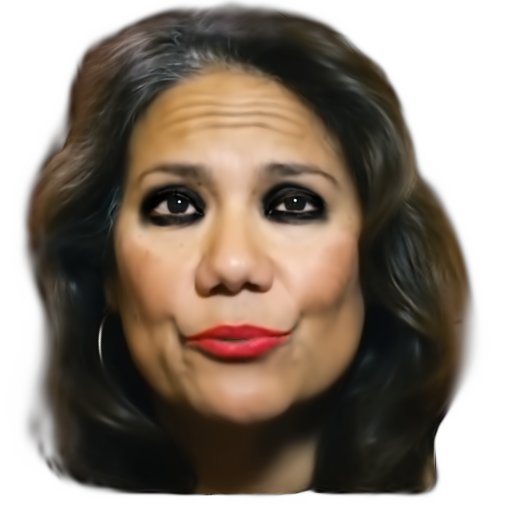} &
        \includegraphics[width=\w,valign=c]{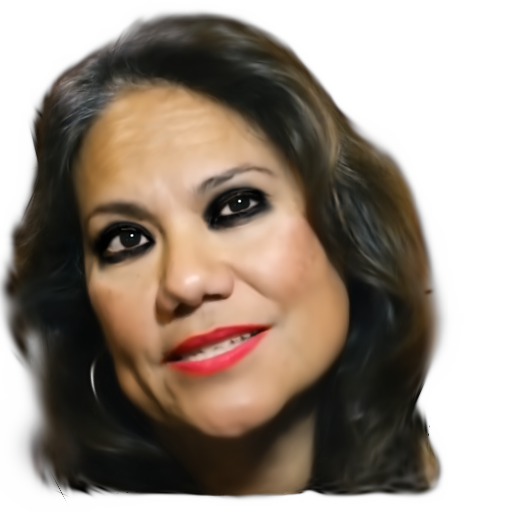} &
        \includegraphics[width=\w,valign=c]{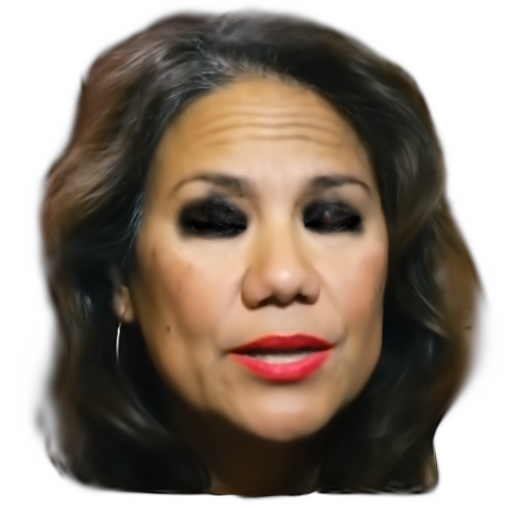} &
        \includegraphics[width=\w,valign=c]{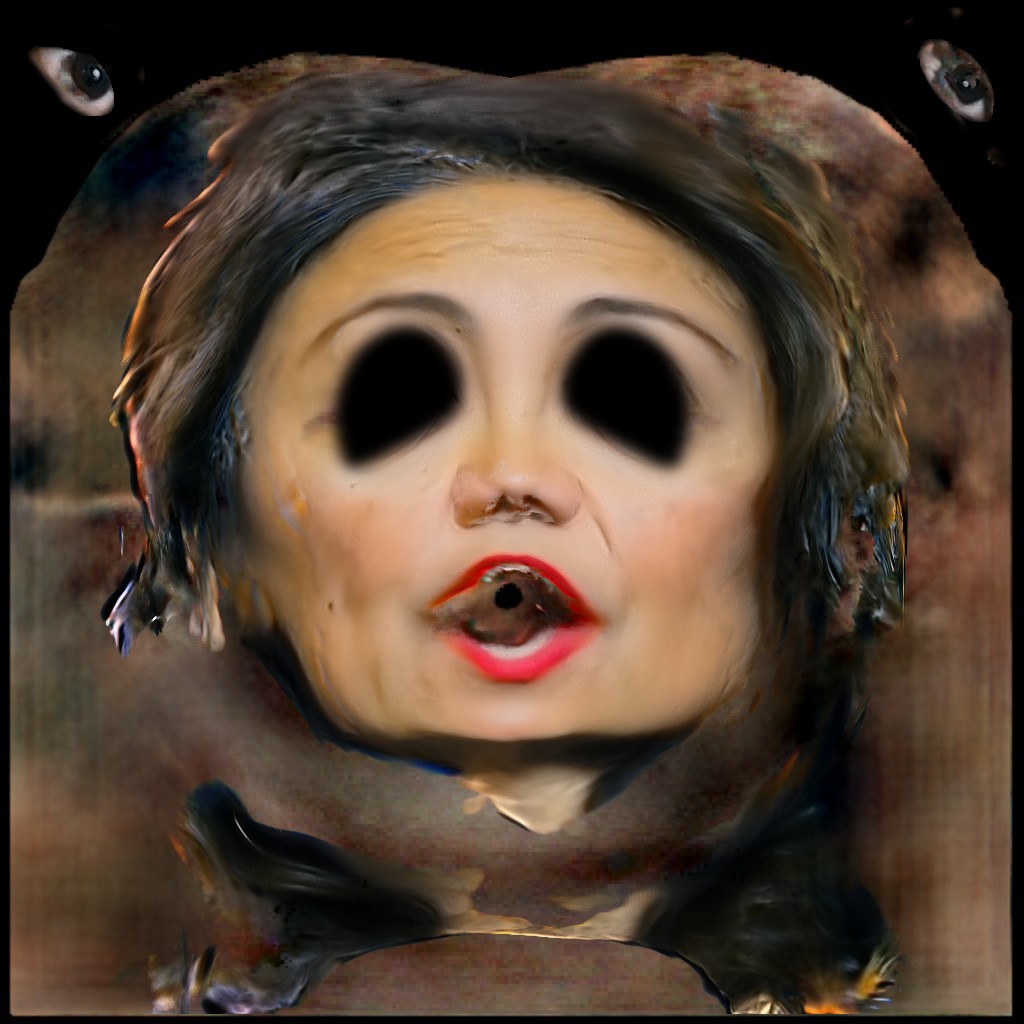}
    \end{tabular}
    }
    \centering
    \caption{Examples of simple albedo texture editing using our method; adding a star decal, changing hair, teeth or eye colors, removing skin imperfections or adding make-up. Changes remain consistent across poses and expressions.}
    \label{fig:exp-texture-editing}
\end{figure*}

The advantages of our novel representation are twofold:
\begin{itemize}
    \item \textbf{Expressive Gaussian primitives}. Spatially varying attributes and normals, in a Phong shading \cite{phong} fashion, as opposed to single values, yield higher reconstruction quality with a smaller total model size than state-of-the-art relightable avatar competition \cite{HRAvatar} (Figure~\ref{fig:lpips_params}). While similar expressiveness can also be achieved with small per-primitive textures (\eg \cite{gstex,BBSplat,texturedgaussiansenhanced3d,texturesplat,SuperGaussians,GaussianBillboards,HDGS}), our approach uses a compact global map, offering both efficiency and editability.
    \item \textbf{UV-based semantic control}. Defining primitive attributes continuously on a standard 3DMM template UV map enables a wide range of semantically grounded manipulations and enhancements, including:
    
    \begin{itemize}
        \item \textbf{Intrinsic material supervision}, such as our albedo regularization with the FLAME albedo model that guides physical decomposition mid-training and reduces artifacts (Figure~\ref{fig:exp-ablation-texturing}). We note that we also incorporate a diffusion-based albedo prior in screen space, similar to \cite{HRAvatar}, to further disambiguate the decomposition. %
        \item \textbf{Model compression} at test time by downsampling the converged textures with controllable loss in rendering quality (Figure~\ref{fig:exp-ablation-downscale-texture}).
        \item \textbf{Intuitive editing} of material and normals directly in the FLAME UV domain (\eg Figures \ref{fig:teaser}, \ref{fig:exp-texture-editing}, \ref{fig:exp-texture-editing-pbr} and supplementary video), without requiring any further optimization, while seamlessly integrating with the avatar and maintaining compatibility with the physical lighting representation.
    \end{itemize}
\end{itemize}

We demonstrate the benefits of our method compared to traditional splat-based reconstruction and rendering techniques, and illustrate its versatility through various texture editing examples. Our method produces highly realistic head reconstructions that not only match or surpass the quality of state-of-the-art monocular relightable avatar HRAvatar~\cite{HRAvatar} -- despite its lack of explicit texture representations -- but also unlocks new possibilities for artistic control.

\begin{figure*}[htbp]
    \centering
    \includegraphics[width=\linewidth]{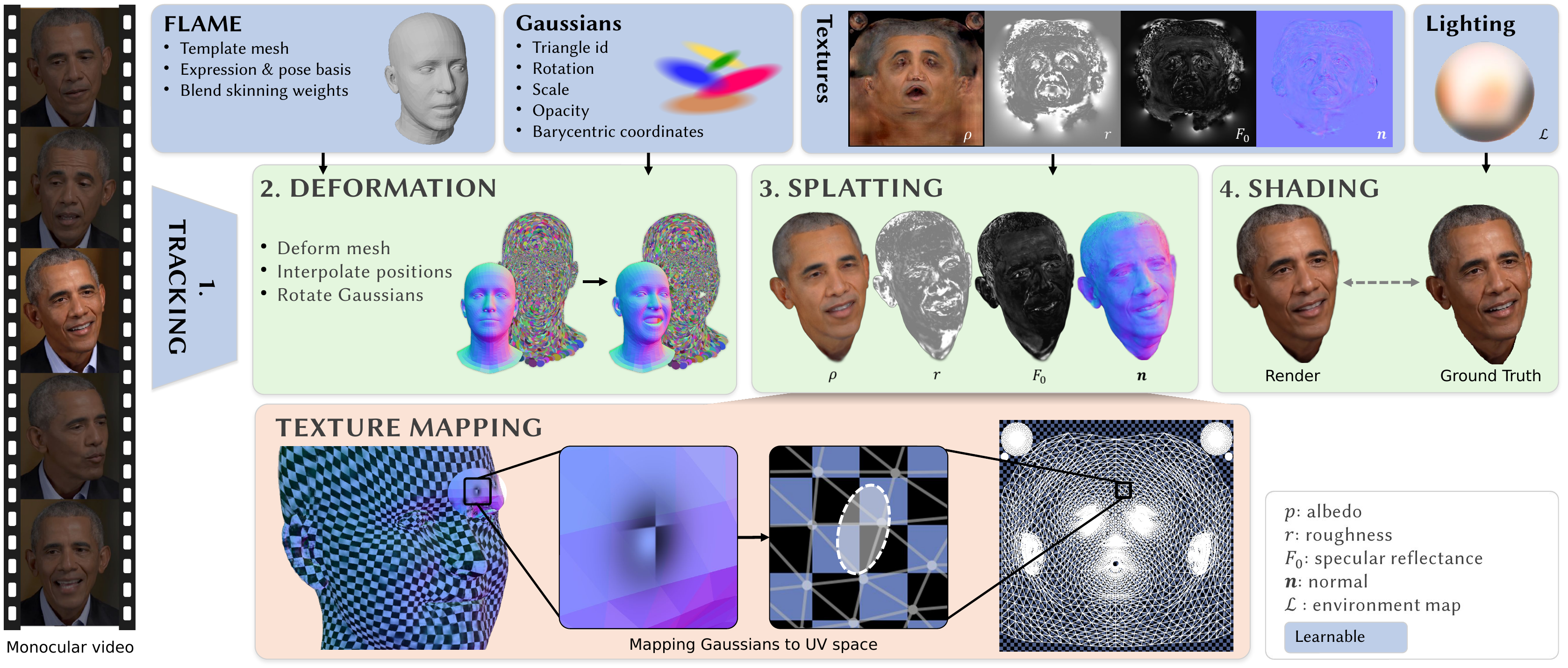}
    \caption{Method overview. GTAvatar takes as input a monocular video sequence and reconstructs it by optimizing parameters of 2D Gaussians, alongside physically-based rendering material and environment lighting. The resulting head avatar can be animated, relit under arbitrary lighting and edited in texture space. The key contribution lies in our efficient texture mapping technique that relates a splat's tangent space to a patch in canonical UV domain (see Section~\ref{sec:method-uvmapping}).}
    \label{fig:method}
\end{figure*}

\section{Related work}

\noindent \textbf{Avatar Representations.}
The prior success of NeRFs \cite{nerf,ingp} brought impressive levels of photorealism in neural avatars based on differentiable volume rendering \cite{headnerf,nerface,adnerf}. Rasterization approaches can alleviate the computational rendering burden of implicit neural ray casting. However, mesh-based rasterization \cite{nvdiffrec} can yield subpar photometric reconstruction performance (\eg FLARE~\cite{flare} in Table~\ref{tab:exp-reconstruction}) due to the limitations of meshes in modeling transparent and complex geometry. The recent advent of Gaussian Splatting \cite{3DGS} has enabled state-of-the-art reconstruction quality with real-time rendering \cite{gaussianheadavatar,gavatar,gaussianavatars}. Gaussian Splatting extends the EWA volume resampling framework~\cite{ewa,ewa2001} to learnable \cite{adam} inverse rendering, modeling 3D scenes with explicit anisotropic Gaussian kernel primitives that can be sorted and rasterized efficiently in tiles. In this respect, the 2DGS variant \cite{2DGS} leverages planar 2D primitives instead of volumetric ones, and performs precise 2D kernel evaluation in object space as opposed to approximate ones in screen space (3DGS), thus leading to superior geometry and multi-view consistency. The superior normals provided by 2DGS~\cite{surfhead,sheap,sparfels,GaussianSurfels} have shown to be beneficial in physics-based inverse rendering \cite{RefGaussian,texturesplat,SVG-IR,GS-2DGS,IRGS} where inferring precise reflection directions \cite{refnerf} is pivotal.
2DGS also defines a local splat coordinate frame that allows us to map ray-splat intersections to our canonical UV space continuously, which is key in our novel avatar representation.       
Generalizable models for 2DGS \cite{meshsplat,sparsplat} have been recently proposed similarly to 3DGS \cite{pixelSplat,mvsplat} before.

\noindent \textbf{Monocular \& Relightable Avatars.}
Learning avatars from casual consumer grade videos paves the road to democratizing volumetric capture, in contrast to the costly requirements of light stages or multi-view setups \cite{lightStage,relightables,travatar}. Recent advances enable avatar learning from monocular video input (\eg \cite{pointavatar,flashavatar,monogaussianavatar,splattingavatar,gaussianblendshapes}), thanks to built-in inductive biases and robust monocular facial tracking (\eg \cite{DECA,EMOCA,SMIRK,MICA}) and reconstruction (\eg \cite{Pixel3DMM, Sapiens, MonoNPHM, PRN, crossmodal}) methods.
Rigging avatars with underlying 3D morphable models (3DMMs) \cite{FLAME,BFM,decoupled,LSFM,CoMA,NonLinear3DMM,facetunegan} enables efficient and robust learning and parametric animation control. They define a canonical space for implicit and explicit representations and factor out expression and pose deformations. Several works leverage data captured under controlled lighting conditions to produce relightable head avatars \cite{rgca,vrmm} with dynamic radiance fields and physical reflectance models.
Relightable Gaussian Codec Avatar \cite{rgca} has shown good relighting capabilities for full multi-view OLAT captures through its material parameterization, and has also been extended to the phone scan setting \cite{URAvatar}.
Despite the increased ill-posedness of the problem, physics-based inverse rendering and relighting from mere monocular videos or sparse images is possible through further regularizations or appearance priors, such as modeling reflectance via simplified bidirectional reflectance distribution functions (BRDF) \cite{flare,spark}, or relying on large data corpora \cite{URAvatar}.
Closest to our context, HRAvatar~\cite{HRAvatar} employs deferred shading of FLAME-rigged 3D Gaussians augmented with material attributes. In contrast to the latter, we build on 2DGS deferred shading with the aim of extending controllability beyond relighting and reenactment, to additionally include intuitive local editing on conformal material maps, in the interest of more expressive and user-controllable avatar manipulation without sacrificing performance.

\noindent \textbf{UV Mapping \& Editing.} Recent research has sought to combine explicit surface parameterization with neural representations. Within the NeRF~\cite{nerf} framework, works such as NeuTex~\cite{xiang2021neutex}, Neural Gauge Field~\cite{zhan2023general} and Nuvo~\cite{srinivasan2024nuvo} learn neural UV mappings that establish bijective correspondences between surfaces and textures, enabling surface-aware rendering across general or category-specific domains. Simultaneously, Gaussian Splatting can support 3D domain editing through semantic grouping and manipulation \cite{gaussian_grouping}, dynamic 4D content~\cite{yu2024cogs}, text-driven scene modifications~\cite{chen2024gaussianeditor}, or brush painting \cite{painting}. Editing via screen-space gradients \cite{texttoon,mega,PortraitGen}, which relies mostly on extensive optimization with pretrained diffusion models \cite{realcompo,zhang2025itercomp} such as Instruct-Pix2Pix~\cite{InstructPix2Pix}, suffers from high computational costs and lacks precise controllability. UV-domain editing requires continuous texture maps. While some methods structure Gaussians in the UV domain \cite{gghead,GaussianShellMaps,splattingavatar,flashavatar,mega}, most still operate with discontinuous sparse maps. Texture-GS~\cite{texturegs} learns a neural UV mapping for 3DGS primitives for static objects, whilst we aim to use a predefined UV map for semantically grounded edits of a dynamic avatar. 
Contemporary work FATE~\cite{fate} attempts to alleviate the texture map continuity issue for non-relightable dynamic Gaussian head avatars. However, it requires a separate learning stage involving neural baking via additional networks to obtain smooth editable textures and fails to preserve sharp details when edited, as shown in Figure~\ref{fig:fate_comparison}.
Our method avoids these multiple stages, bypassing the need for any neural UV mapping or neural baking. Moreover, to our knowledge, our approach is the first Gaussian head avatar to conjointly enable relighting and UV editing of appearance and normals.

\noindent \textbf{2D Image-based relighting.} A significant area of research addresses portrait relighting from monocular inputs directly, bypassing the need for full 3D reconstruction. Many approaches decompose the input image into various intrinsic components such as albedo, normals and specular reflectance \cite{faceRelightArbitraryWang09, NeuralFace2017, sfsnetSengupta18, illuInvariantFaceRecWACV19, pandey2021total}, leveraging priors for accurate lighting decomposition. These priors often include parametric face models \cite{blanzvetter1999, FLAME, decoupled, LSFM, CoMA, NonLinear3DMM, facetunegan}, light-stage datasets \cite{debevec2000acquiring} or simplified physical models \cite{phong, cook1982reflectance}. In recent years, many works have used encoder-decoder architectures \cite{deepReflectanceFieldsMeka19, portraitRelightSun19, zhou2019deep, portraitRelightWang20, Nestmeyer_2020_CVPR, pandey2021total, lite2relight} or diffusion models \cite{ponglertnapakorn2023difareli, zeng2024dilightnet, chaturvedi2025synthlight} that can directly synthesize relit images. While these methods can capture complex illumination effects (e.g. hard shadows, inter-reflections and subsurface scattering), disentanglement of lighting from other factors remains difficult in the under-constrained monocular setting. Additionally, the lack of explicit 3D representations severely limits their ability to render novel viewpoints.

\section{Method}

Our method reconstructs a relightable and animatable 3D head avatar from a single monocular video by embedding 2D Gaussians onto the surface of a template mesh, where each Gaussian inherits physical material properties from texture maps dynamically. An overview is of the pipeline is shown in Figure \ref{fig:method}. The approach builds on the FLAME morphable model, which provides a parametric representation of head geometry and facial deformations. We first review the FLAME model and the principles of Gaussian splatting. We then describe how Gaussians are bound to mesh triangles, introduce our novel UV-space mapping that enables differentiable texturing, and present our real-time image-based lighting formulation for physically plausible shading.

\subsection{FLAME} \label{sec:method-flame}

The FLAME~\cite{FLAME} 3D morphable model provides a template head mesh with joints for eyes, jaw and neck, and a statistical model for identity and expressions. Identity and expression deformations are expressed as blendshapes while the joints transform vertices through Linear Blend Skinning \cite{LBS}.
For simplicity, we refer to FLAME as template vertex positions $V_t\in\mathbb{R}^{3V}$, triangle topology $[0..V-1]^{3F}$ and a deformation function $\mathcal{F}$:
$$V_d = \mathcal{F}(V_t, \Psi)$$
where the identity parameters are fixed and $\Psi$ is the concatenation of pose and expression parameters.
We refer the reader to \cite{FLAME} for exhaustive details on this 3DMM. 
Additionally, FLAME provides per-corner UV coordinates for texturing (\ie UV parameterization for the unwrapped template mesh).

\subsection{Gaussian Splatting} \label{sec:method-gs}

3D Gaussian Splatting \cite{kerbl3Dgaussians} (3DGS) uses 3D Gaussians primitives to reconstruct a 3D scene from a set of images. Each Gaussian is defined with a position $\mu$, a rotation matrix $R$ (parameterized as a quaternion), scales $s\in\mathbb{R}^3$, opacity $o$ and view-dependent color $c$ as spherical harmonics coefficients. For rendering, Gaussians are projected to screen space and alpha-blended front-to-back, yielding the final color: $$C=\sum_i c_i o_i G_i(x)\prod_{j=1}^{i-1}(1-o_jG_j(x))$$ where $G(x) = \text{exp}(-\frac{1}{2}(x-\mu)^T\Sigma^{-1}(x-\mu))$ is the Gaussian function, $\Sigma = RSS^TR^T$ the covariance matrix and $S$ the diagonal scaling matrix derived from $s$.

2D Gaussian Splatting \cite{2DGS} (2DGS) extends this by using flat 2D surfels better suited for reconstructing high-fidelity surfaces. Each gaussian defines a local tangent plane in world space, which is parameterized as: $$P(s,t) = p + s\  \mathbf{s} + t\ \mathbf{t}$$ where $\mathbf{s}\in\mathbb{R}^3$ and $\mathbf{t}\in\mathbb{R}^3$ are the scaled tangential vectors of the splat and $p\in\mathbb{R}^3$ its center position. Contrary to 3DGS, 2DGS enables computation of closed-form normal vectors $\mathbf{n} = \frac{\mathbf{s}\times\mathbf{t}}{||\mathbf{s}\times\mathbf{t}||}$ and precise depth at ray-splat intersections. Please note that we purposefully use $(s,t)$ coordinates to describe the Gaussian tangent planes in place of the more common $(u,v)$ to avoid confusion with $(u,v)$ coordinates used for texture sampling.

\subsection{Mesh Binding} \label{sec:method-meshbinding}

\noindent{\textbf{Representation.}} Similarly to existing Gaussian head avatars \cite{gaussianavatars, lee2025surfhead, HRAvatar, flashavatar}, we bind Gaussians to triangles of the FLAME mesh. Within each triangle, $n$ splats are initialized with random barycentric coordinates within $[0,1]$. Each Gaussian has a learnable rotation $r$ relative to the orientation of its triangle (represented as a quaternion), a displacement $d\in\mathbb{R}$ along the triangle normal, scales $s\in\mathbb{R}^2$, and opacity $o\in[0,1]$.

\noindent{\textbf{Deformation.}} Given pose and expression parameters for FLAME, we first compute the deformed vertex positions $V_d\in\mathbb{R}^{3V}$ and face normals $N_{d}\in\mathbb{R}^{3F}$.
The position of each Gaussian is obtained by interpolating the deformed vertex positions of its triangle with barycentric coordinates. The rotation is the Gaussian's relative rotation $r$ multiplied by the triangle's orientation matrix, whose columns are a tangent, bi-tangent and normal vector of the triangle plane.

\subsection{UV-Mapping for Textured Gaussians} \label{sec:method-uvmapping}

\begin{figure}[h]
    \centering
    \includegraphics[width=0.7\columnwidth]{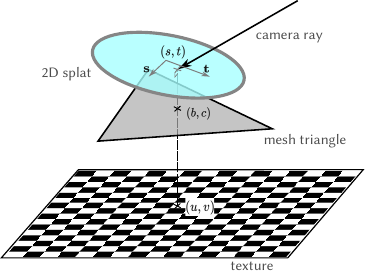}
    \caption{From ray-splat intersection to $(u,v)$ texture coordinates. The local 2D splat coordinate $(s,t)$ is orthogonally projected onto the splat's reference triangle, expressed as triangle barycentric coordinates and subsequently mapped to a $(u,v)$ position.}
    \label{fig:raysplat}
\end{figure}

Our method extends Gaussian Splatting by replacing the color of Gaussians with texture patches. Contrary to existing works that enhance the representation power of Gaussians with small per-primitive textures \cite{gstex,BBSplat,texturedgaussiansenhanced3d}, we seek to embed Gaussians into a single texture space with a semantic mapping predefined on the template mesh. To achieve this, we map the tangent plane of each splat to a continuous patch in the mesh's UV space. We modify the Gaussian Splatting rasterizer to retrieve colors by sampling a texture at the mapped UV coordinates efficiently with bilinear filtering. This texture is learned alongside the Gaussian parameters.

Given a ray-splat intersection position $(s,t)$ in the tangent space of the splat,
an intuitive and practical solution for the mapping to $(u,v)$ coordinates is to use the nearest mesh point. For ease of computation, we approximate this point as the nearest point on the primitive's bound triangle plane, \ie  the orthogonal projection of the ray-splat intersection point on its triangle plane as illustrated in Figure~\ref{fig:raysplat}. The final UV coordinates are thus obtained by linearly interpolating the UV coordinates of the triangle vertices at this projection location. Irregularities resulting from the projection falling outside the triangle are mitigated by the splat scales remaining generally smaller than triangles, and by our UV regularization loss introduced in Section \ref{sec:uv_reg}. 

Minimizing the number of operations per ray-splat intersection is key to maintaining high rendering speed.
To this end, we devise a method that only requires a single matrix multiplication per intersection to compute the final UV coordinates, as naive alternatives do not scale well with the large number of ray-splat intersections involved in rendering.
For a splat attached to a mesh triangle with UV coordinates $\text{UV}_X\in[0,1]^2$\ and FLAME-deformed vertices $\text{V}_X\in\mathbb{R}^3$ ($X\in\{A,B,C\}$), we adopt triangle barycentric coordinates with origin at $V_A$ and define two functions that map from barycentric coordinates $(b,c)$ to world space and UV respectively:
$$
\text{v}(b, c) = \text{V}_A + \Jtri \begin{bmatrix}b\\ c\end{bmatrix} \quad
\text{uv}(b, c) = \text{UV}_A + \Juv \begin{bmatrix}b\\ c\end{bmatrix}
$$
where \Jtri{} and \Juv{} denote the Jacobians of the two transformations:
\begin{align}
\Jtri = & \begin{bmatrix}\begin{array}{@{}c|c@{}}\text{V}_B-\text{V}_A & \text{V}_C-\text{V}_A\end{array}\end{bmatrix}\in\mathbb{R}^{3\times2} \nonumber \\
\Juv = & \begin{bmatrix}\begin{array}{@{}c|c@{}}\text{UV}_B-\text{UV}_A & \text{UV}_C-\text{UV}_A\end{array}\end{bmatrix}\in\mathbb{R}^{2\times2} \nonumber
\end{align}
Similarly, let $f$ denote the transformation from the splat's tangential space to world space:
$$
f(s,t) = p + \Jst \begin{bmatrix}s\\ t\end{bmatrix}, \quad
\Jst = \begin{bmatrix}\begin{array}{@{}c|c@{}}\mathbf{s} & \mathbf{t}\end{array}\end{bmatrix}\in\mathbb{R}^{3\times2}
$$
Assuming a non-degenerate triangle, $\text{v}$ is invertible and its inverse is defined on the triangle plane. Thus, if the Gaussian is aligned with the triangle plane, we can define the mapping ${(u,v) = \mathcal{M}(s,t) = (\text{uv} \circ \text{v}^{-1} \circ f) (s,t)}$.
We exploit the linearity of $\text{uv}$, $\text{v}$, $f$ and $\mathcal{M}$ to rewrite it as follows:
\begin{gather}
\uvzero = \mathcal{M}(0,0) \quad \Jstuv = \frac{\partial \mathcal{M}}{\partial st} = \Juv \Jtriinv \Jst \\
\mathcal{M}(s,t) = \uvzero + \Jstuv \begin{bmatrix} s \\ t \end{bmatrix}
\end{gather}
where \Jtriinv{} is the pseudo-inverse of \Jtri{} and \uvzero{} the UV coordinates at the center of the splat, computed using the Gaussian's barycentric coordinates directly. 
We extend this to Gaussians that are not aligned with their triangle by noting that $\Jtri\Jtriinv$ is the orthogonal projector onto the column space of \Jtri{} (the triangle plane) and $x\mapsto\Jtriinv (x-\text{V}_A)$ yields the barycentric coordinates of the projection. Thus, $\mathcal{M}$ maps $(s,t)$ to the UV coordinates of this projection (the $\Jtriinv \text{V}_A$ term is accounted for in \uvzero{}). The pseudo-inversion is implemented by solving a least-squares problem, which does not back-propagate gradients to the template vertices.

This lightweight mapping only requires a single matrix multiplication at each ray-splat intersection during rendering. The largest computational costs are the pseudo-inversion of \Jtriinv{} for each triangle and the bilinear texture filtering; the latter is a well-understood cost of graphics pipelines that can be mitigated with hardware acceleration.
Notice that \Jtri{} -- and its pseudo-inverse \Jtriinv{} -- are mesh and expression-dependent but do not need to be updated for viewpoint changes. See Section~\ref{sec:exp-rendering-time} for more details on rendering speed and a comparison with a naive projection baseline.

\subsection{Appearance Modeling} \label{sec:method-appearance}

\newcommand{\wo}{\ensuremath{\mathbf{\omega_0}}}
\newcommand{\wi}{\ensuremath{\mathbf{\omega_i}}}
\newcommand{\wrefl}{\ensuremath{\mathbf{\omega_r}}}
\newcommand{\n}{\ensuremath{\mathbf{n}}}
\newcommand{\Li}{\ensuremath{\mathcal{L}}}

The final per-pixel color is derived using deferred shading on the G-buffers obtained through Gaussian Splatting. For a given pixel, let $\n\in\mathbb{R}^3$ denote the splatted normal vector (normalized) and $\wo\in\mathbb{R}^3$ the view direction. We replace typical per-Gaussian spherical harmonics with a 5-channel material texture: albedo $\rho\in[0,1]^3$, roughness $r\in[0,1]$ and specular reflectance $f_0\in[0,1]$. After rasterization, we retrieve those values per-pixel in the G-buffers. 

\subsubsection{Real-time Physically-Based Rendering} \label{sec:pbr}

We adopt the real-time shading model of \cite{karis2013real}, which we briefly summarize in this section. The rendering equation expresses the outgoing light $L_0$ leaving surface point $x$ in the camera direction \wo{} by integrating the light reaching it from all directions:
$$L_0(x, \omega_0) = \int_{\Omega}f_r(x,\wi, \wo) L_i(x,\omega_i)(\wi\cdot\n) d\wi$$
where $f_r$ is the bidirectional reflectance distribution function of the material (BRDF). We decompose the BRDF as a diffuse Lambertian term independent of incoming light direction:
\begin{equation}
    L_{0,\text{diffuse}} = \rho \int_{\Omega} L_i(x,\omega_i)(\wi\cdot\n) d\wi
    \label{eq:L0diff}
\end{equation}
and a view-dependent specular term, calculated using the split-sum approximation \cite{karis2013real}:
{
    \small %
    \begin{equation}
        L_{0,\text{spec}} =
        \left(\int_{\Omega}f_{r,\text{spec}}(x,\wi, \wo) d\wi\right)
        \left(\int_{\Omega} L_i(x,\omega_i)(\wi\cdot\n) d\wi\right)
        \label{eq:L0spec}
    \end{equation}
}
Similarly to previous relightable head avatars \cite{nvdiffrec, flare, HRAvatar}, we adopt the Cook-Torrance microfacet reflectance model \cite{cook1982reflectance}, which parameterizes the specular BDRF with surface roughness $r$ and specular reflectance at normal incidence $f_0$.
In practice, the second integral incorporates parts of $f_{r,\text{spec}}$ that are not dependent on \wo{}, and is stored as a pre-filtered environment map $\Li(\wrefl, r)$ with mip levels sampled by spatially-varying roughness $r$, where \wrefl{} is the reflection of \wo{} onto the normal. We learn this function as a RGB cubemap that represents the lighting of the scene.
The first integral is only dependent on material properties and viewing angle and independent of lighting. It is pre-computed in a look-up table; for conciseness, we refer the reader to \cite{karis2013real} for details and denote this reflectance term as $R(f_0, r, \wo\cdot\n)$.
Note that due to properties of the BRDF omitted here, the integral in Eq. \ref{eq:L0diff} can be computed as $\Li(\n, 0)$. Hence the final color is:
\begin{align}
L_0(x, \omega_0) =& L_{0,\text{diffuse}} + L_{0,\text{spec}} \nonumber\\
=& \rho \cdot \Li(\n, 0) + R(f_0, r, \wo\cdot\n) \cdot \Li(\wrefl, r)
\end{align}

\subsubsection{Normal Mapping} \label{sec:method-normalmapping}

The use of Texture Mapping for appearance properties increases the representation power of each Gaussian. However, the shading model described in Section~\ref{sec:pbr} requires precise surface normals to achieve photo-realistic results. Relying solely on increasing the number of Gaussians to model this high-frequency geometry is inefficient and can lead to a prohibitively large memory footprint. Instead, inspired by classical normal mapping in real-time graphics, we augment our material texture with two additional channels that define a local perturbation relative to the splat's own geometric normal. This allows a single, relatively large Gaussian to represent a surface patch with intricate geometric detail.

\noindent \textbf{From 2D Texture to 3D Tangent-Space Normal.}
The two normal map channels, $(n_x, n_y)$, parameterize a 3D unit normal vector, $\mathbf{n}_t$, within the local tangent space of the splat. The splat's local coordinate system is defined by its orthonormal basis: the two principal tangential vectors $(\mathbf{s}, \mathbf{t})$ and its geometric normal $\mathbf{n}_g = \frac{\mathbf{s} \times \mathbf{t}}{||\mathbf{s} \times \mathbf{t}||}$. To reconstruct the full 3D vector from the 2-channel texture, the third component $n_z$ is derived assuming the normal points away from the surface ($z > 0$):
\begin{equation}
    \mathbf{n}_t = 
    \begin{bmatrix}
        n_x & n_y & \sqrt{1 - n_x^2 - n_y^2}
    \end{bmatrix}^T
\end{equation}
The resulting unit vector $\mathbf{n}_t$ represents the high-frequency surface normal in the splat's local space. A value of $(0,0)$ in the texture corresponds to the tangent-space normal $[0, 0, 1]^T$, representing no perturbation from the splat's geometric normal $\mathbf{n}_g$.

\noindent \textbf{Transformation to World Space.}
During rendering, we sample the tangent-space normal $\mathbf{n}_t$ from our texture. This local normal is then transformed into world space by multiplying it with the splat's rotation matrix $R$:
\begin{equation}
    \mathbf{n}_w = R \mathbf{n}_t, \quad \text{where} \quad R = 
    \begin{bmatrix}
        \mathbf{s} & \mathbf{t} & \mathbf{n}_g
    \end{bmatrix}
\end{equation}
This final perturbed world-space normal, $\mathbf{n}_w$, is then used for all shading calculations as described in Section~\ref{sec:pbr}. This strategy enables higher-quality shading and more accurate reconstruction of fine details with a significantly reduced number of primitives. Figure~\ref{fig:exp-ablation-normal-map} shows examples of the detailed normal maps recovered by our method and demonstrates the resulting improvement in rendering quality compared to using only the base splat normals.

\begin{figure*}
    \newcommand{\w}{0.14\linewidth}
    \setlength{\tabcolsep}{0.0em} %
    \def\arraystretch{0.8}{ %
    \begin{tabular}{cccccccc}
         Ground truth & Render & Normal & Albedo & \multicolumn{3}{c}{Relight}
        \\
        \includegraphics[width=\w,valign=c]{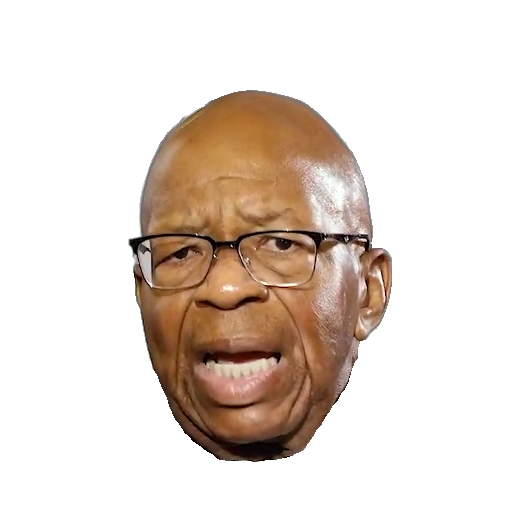} &
        \includegraphics[width=\w,valign=c]{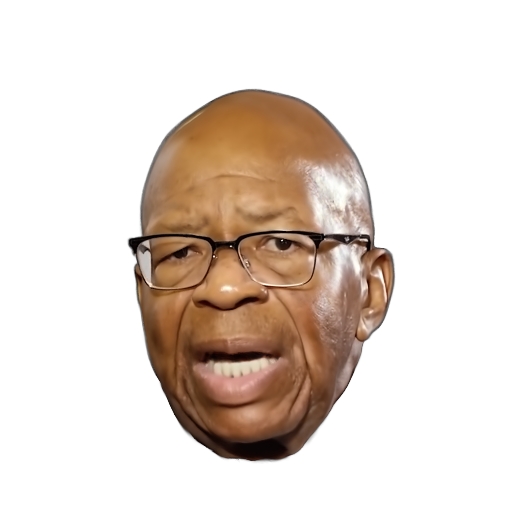} &
        \includegraphics[width=\w,valign=c]{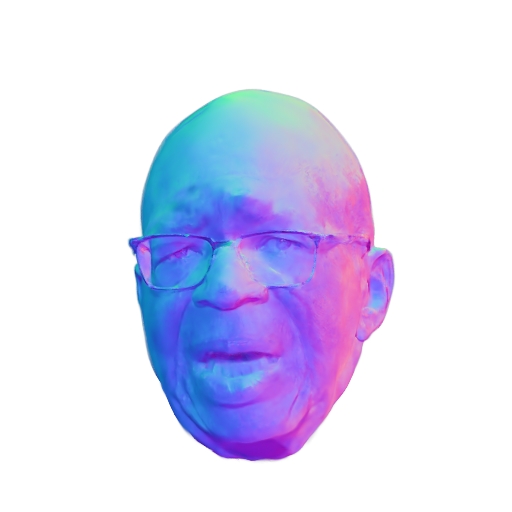} &
        \includegraphics[width=\w,valign=c]{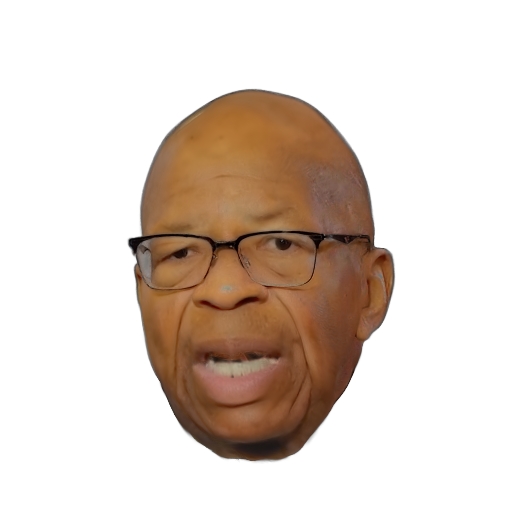} &
        \includegraphics[width=\w,valign=c]{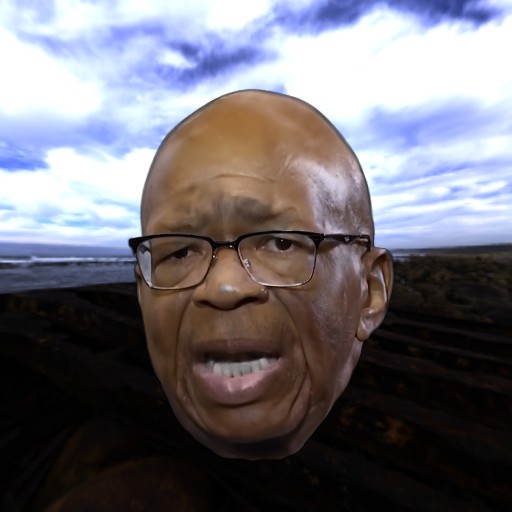} &
        \includegraphics[width=\w,valign=c]{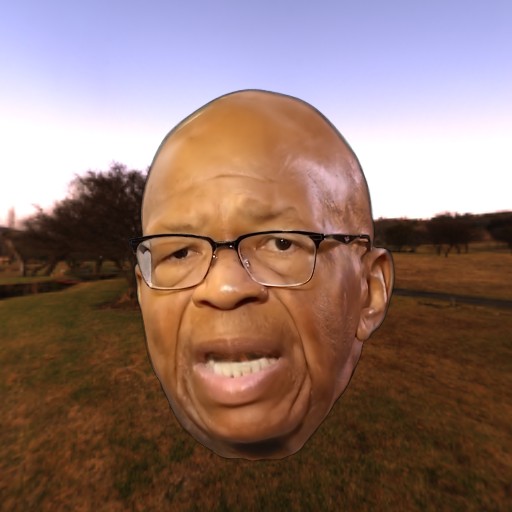} &
        \includegraphics[width=\w,valign=c]{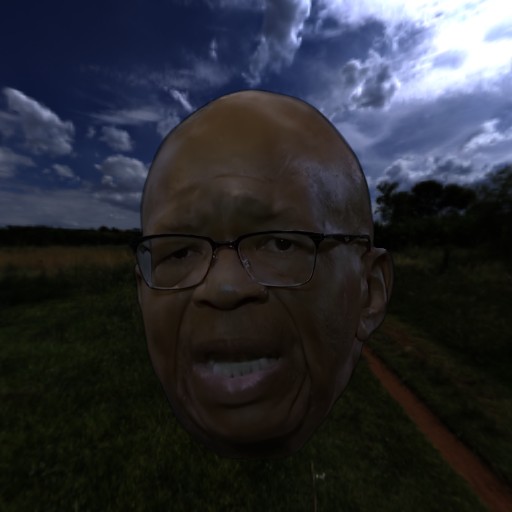}
        \\
        \includegraphics[width=\w,valign=c]{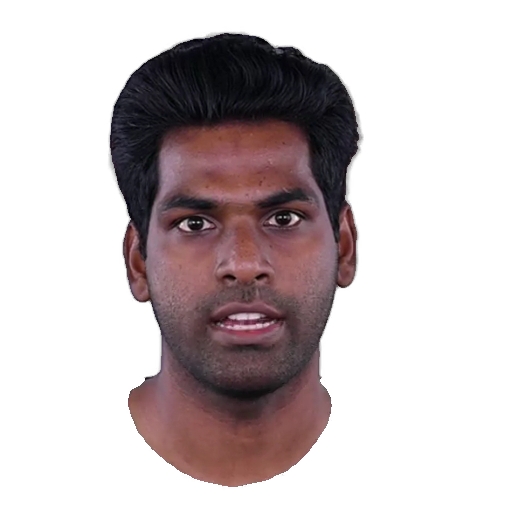} &
        \includegraphics[width=\w,valign=c]{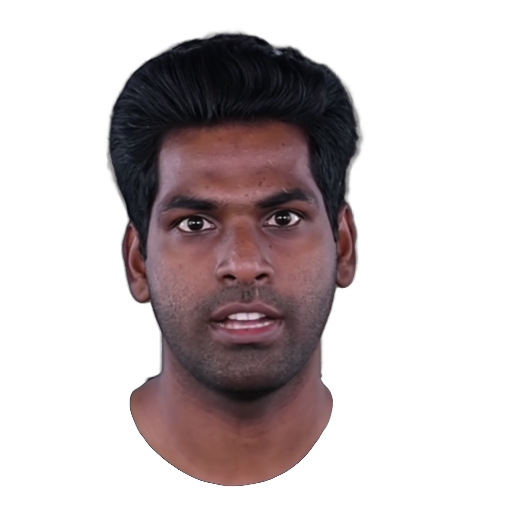} &
        \includegraphics[width=\w,valign=c]{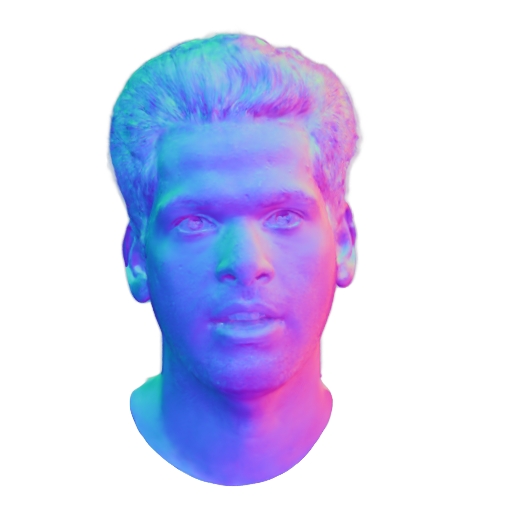} &
        \includegraphics[width=\w,valign=c]{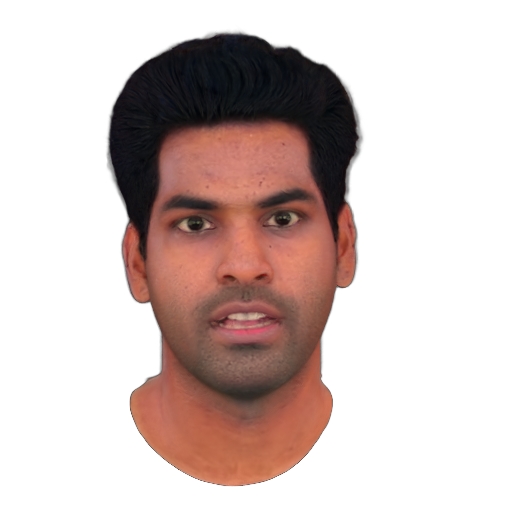} &
        \includegraphics[width=\w,valign=c]{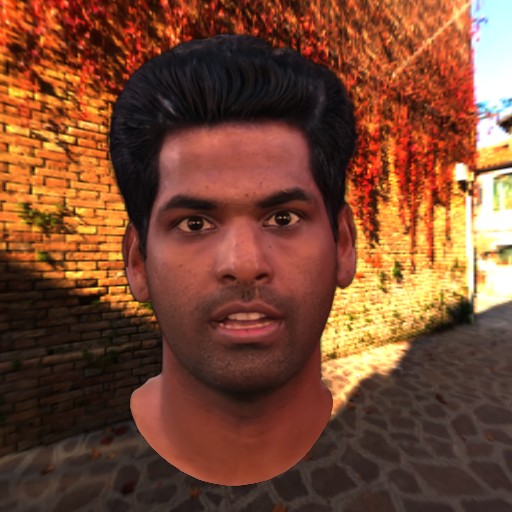} &
        \includegraphics[width=\w,valign=c]{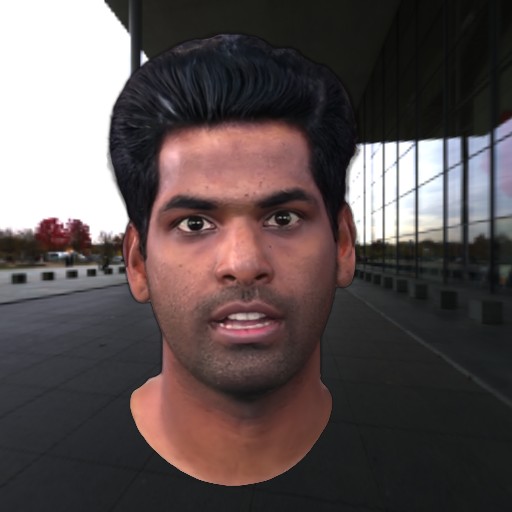} &
        \includegraphics[width=\w,valign=c]{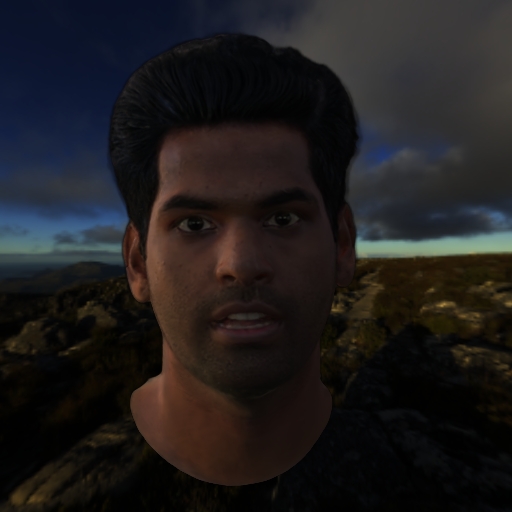}
        \\
        \includegraphics[width=\w,valign=c]{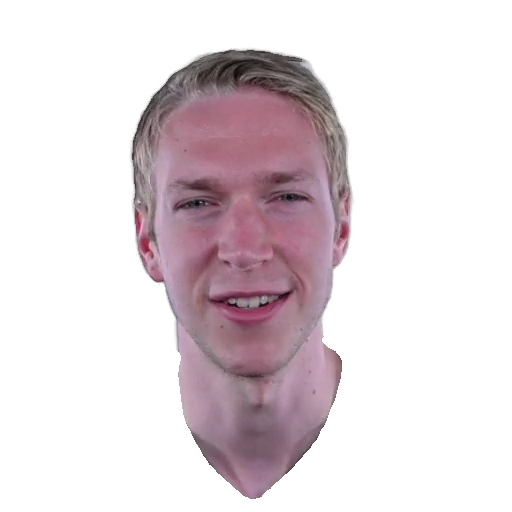} &
        \includegraphics[width=\w,valign=c]{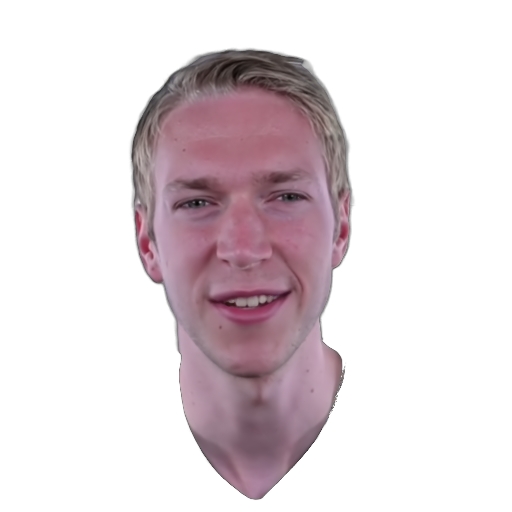} &
        \includegraphics[width=\w,valign=c]{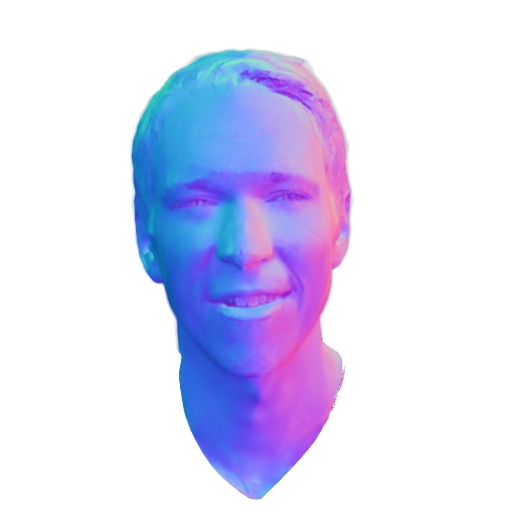} &
        \includegraphics[width=\w,valign=c]{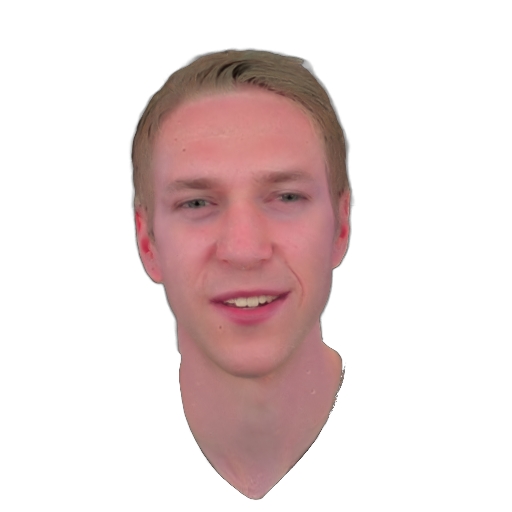} &
        \includegraphics[width=\w,valign=c]{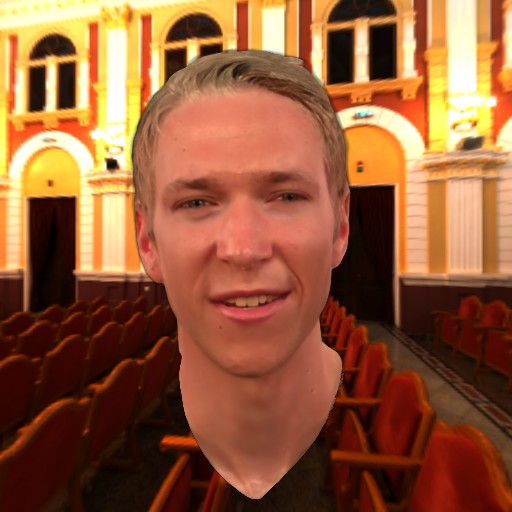} &
        \includegraphics[width=\w,valign=c]{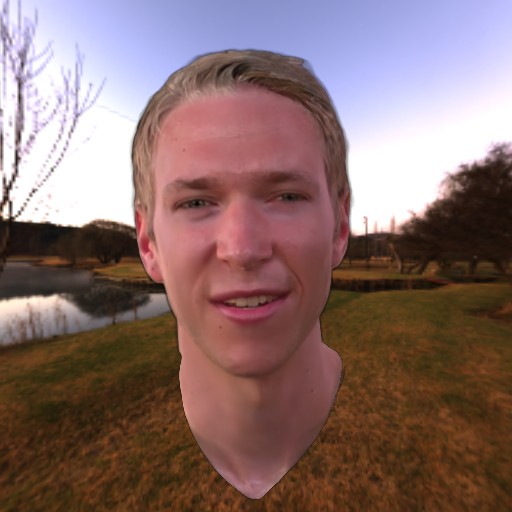} &
        \includegraphics[width=\w,valign=c]{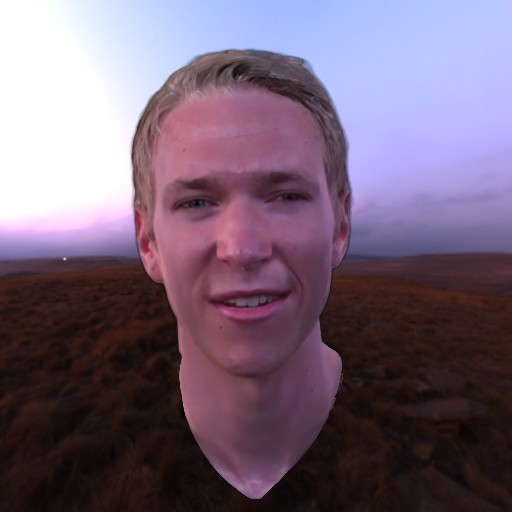}
        \\
        \includegraphics[width=\w,valign=c]{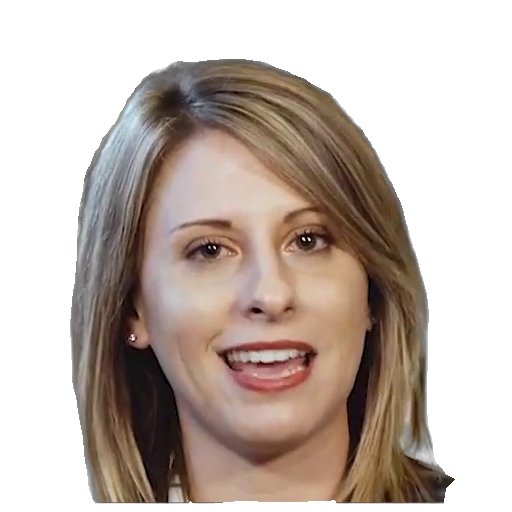} &
        \includegraphics[width=\w,valign=c]{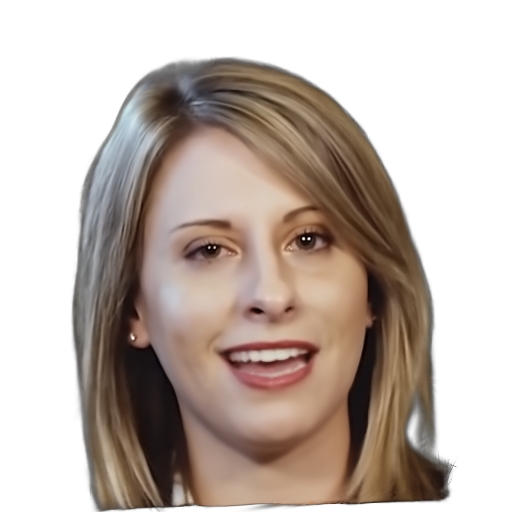} &
        \includegraphics[width=\w,valign=c]{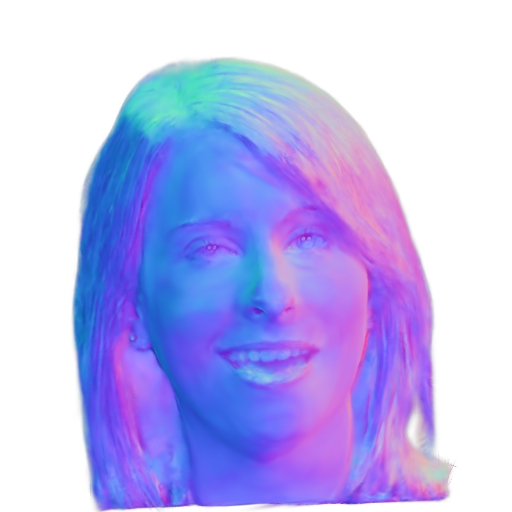} &
        \includegraphics[width=\w,valign=c]{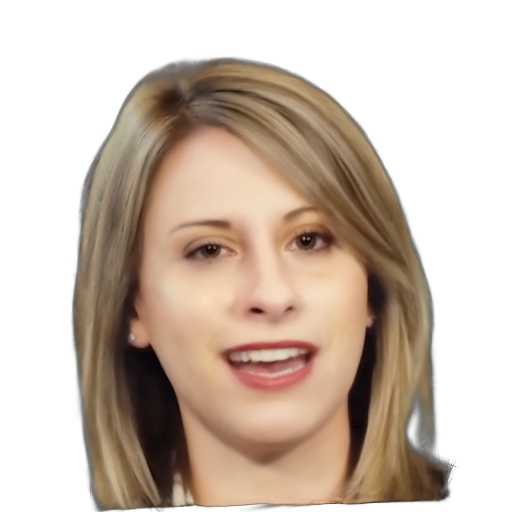} &
        \includegraphics[width=\w,valign=c]{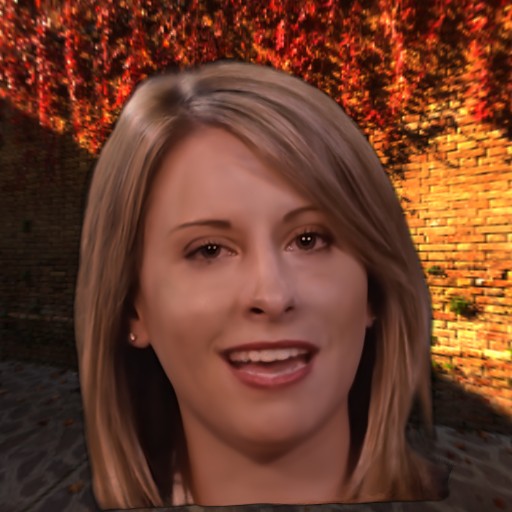} &
        \includegraphics[width=\w,valign=c]{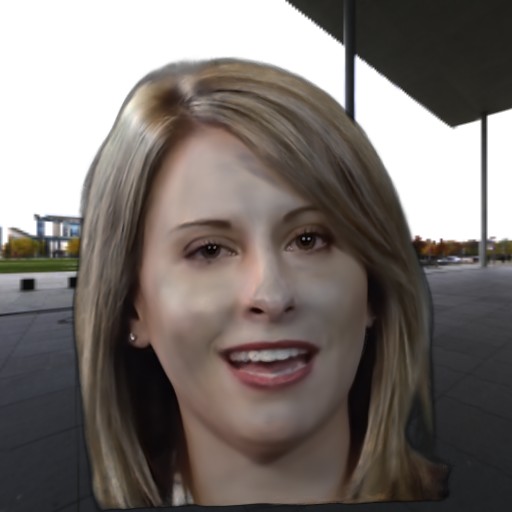} &
        \includegraphics[width=\w,valign=c]{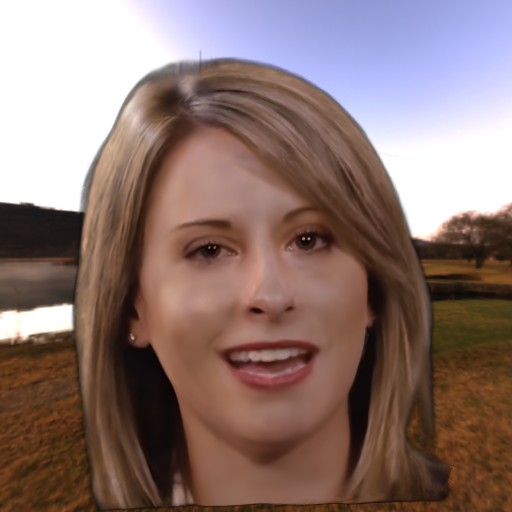}
        \\
        \includegraphics[width=\w,valign=c]{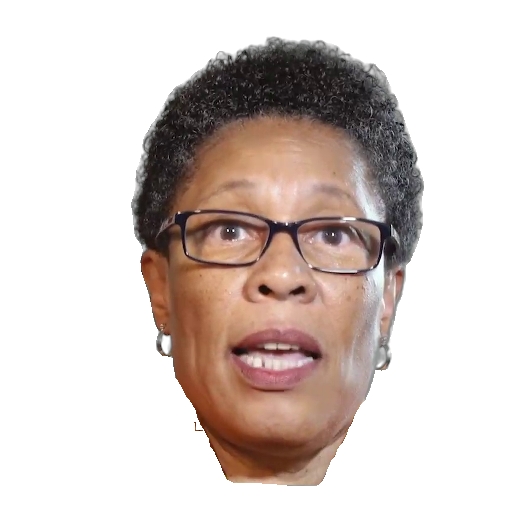} &
        \includegraphics[width=\w,valign=c]{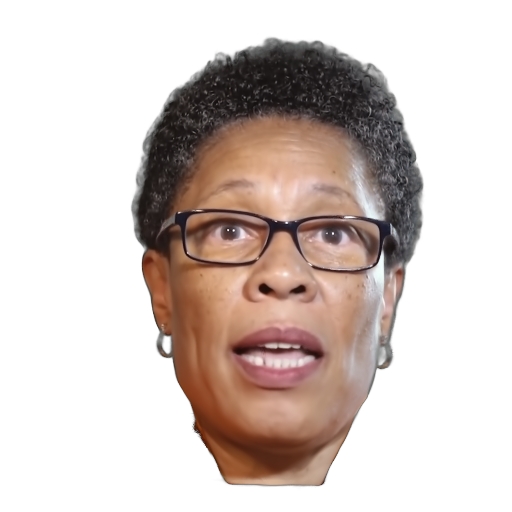} &
        \includegraphics[width=\w,valign=c]{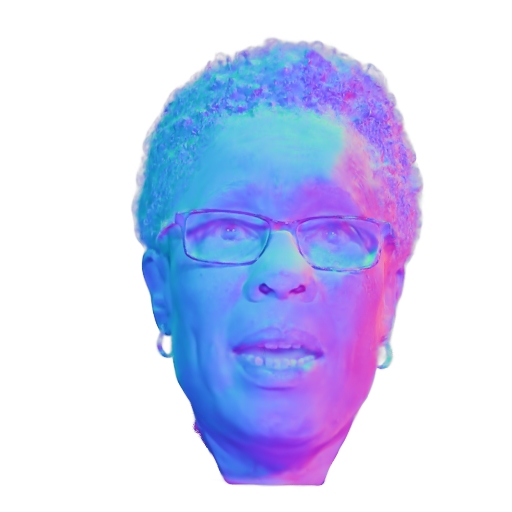} &
        \includegraphics[width=\w,valign=c]{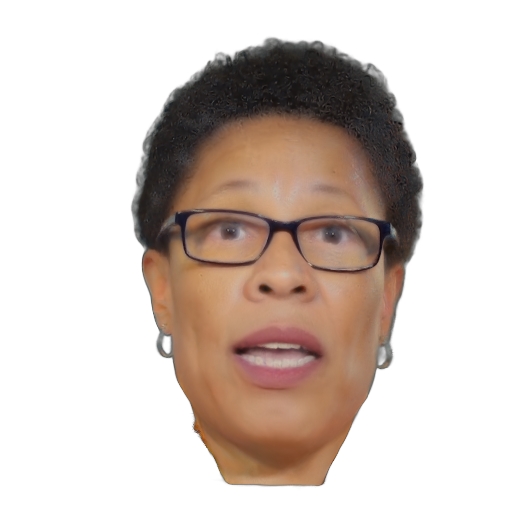} &
        \includegraphics[width=\w,valign=c]{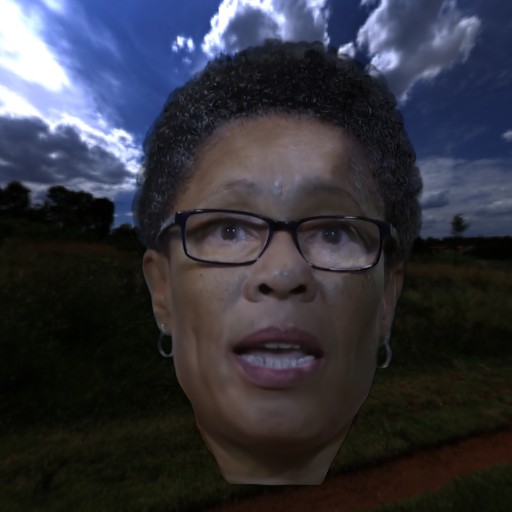} &
        \includegraphics[width=\w,valign=c]{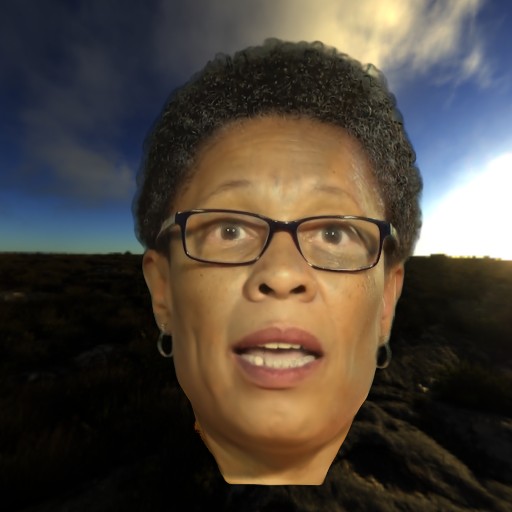} &
        \includegraphics[width=\w,valign=c]{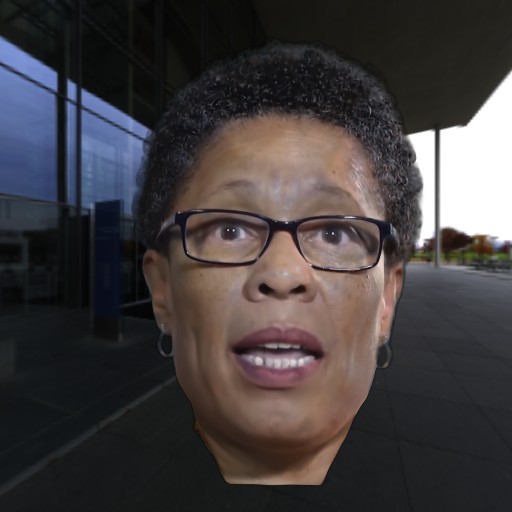}
        \\
        \includegraphics[width=\w,valign=c]{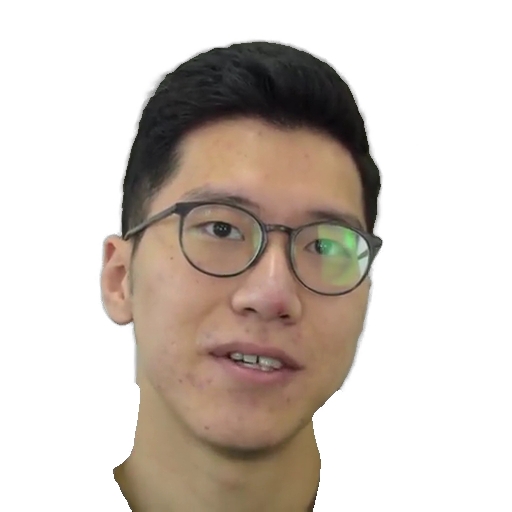} &
        \includegraphics[width=\w,valign=c]{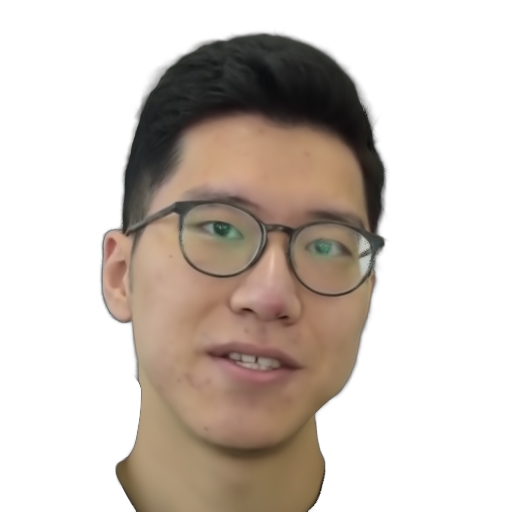} &
        \includegraphics[width=\w,valign=c]{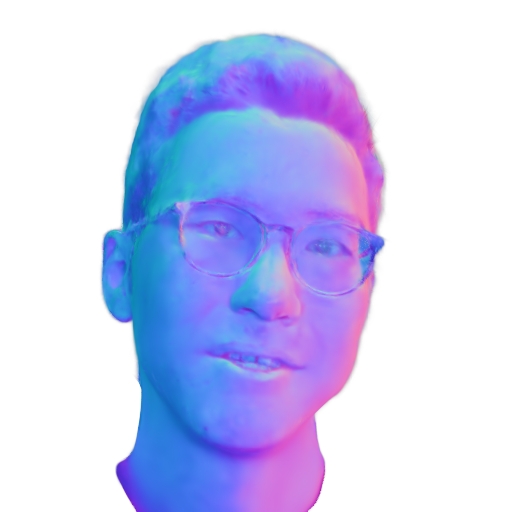} &
        \includegraphics[width=\w,valign=c]{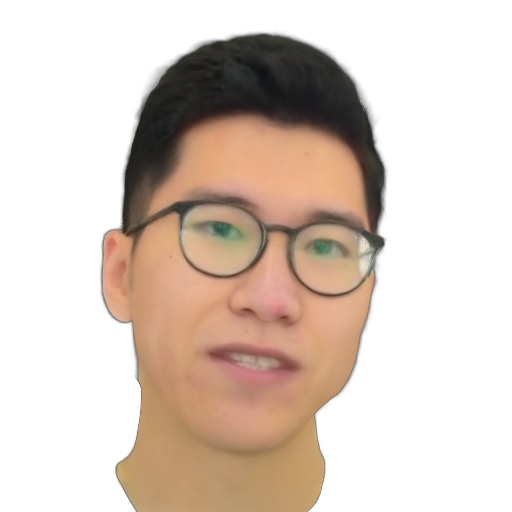} &
        \includegraphics[width=\w,valign=c]{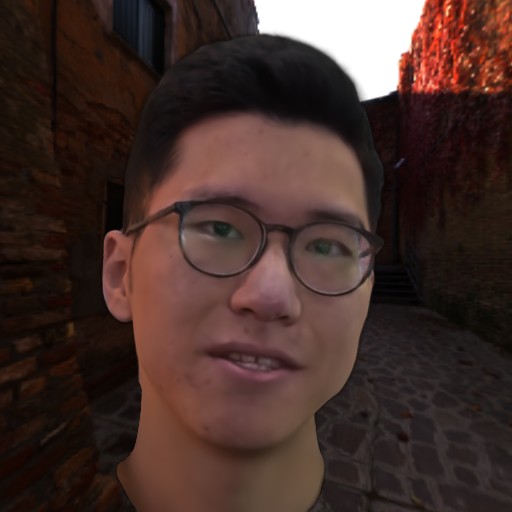} &
        \includegraphics[width=\w,valign=c]{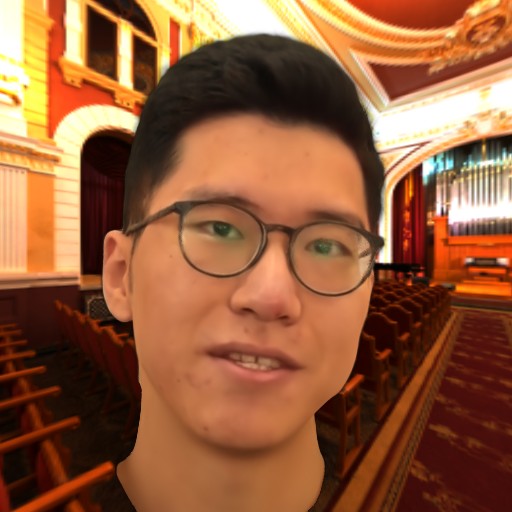} &
        \includegraphics[width=\w,valign=c]{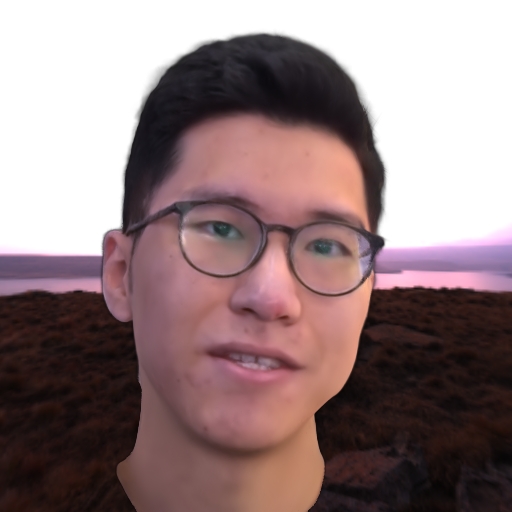}
    \end{tabular}
    }
    \centering
    \caption{Reconstruction and relighting examples obtained with our method. From left to right: original frame, reconstruction, rendered normals  and albedo, relighting under various environment maps.}
    \label{fig:result_ours}
\end{figure*}

\begin{table*}[h]
    \centering
    \newcommand{\coloryes}[1]{\color{ForestGreen}#1}
    \newcommand{\colorno}[1]{\color{gray}#1}
    \begin{tabular}{rlcccccccc}
        \toprule
        
        \multicolumn{2}{c}{\multirow{2}{*}[-3pt]{Method}} &
        
        \multirow{2}{*}[-3pt]{Relighting} & 
        \multirow{2}{*}[-3pt]{\shortstack{Texture\\ editing}} & 
        \multicolumn{3}{c}{INSTA dataset} & 
        \multicolumn{3}{c}{HDTF dataset} \\
        \cmidrule{5-10}
        & & & & PSNR $\uparrow$ & SSIM $\uparrow$ & LPIPS $\downarrow$ & PSNR $\uparrow$ & SSIM $\uparrow$ & LPIPS $\downarrow$ \\ 
        \midrule
        
        \cite{zielonka2022insta} & INSTA & \colorno{$\times$}   & \colorno{$\times$}   & 27.85 & 0.9110 & 0.1047 & 25.03 & 0.8475 & 0.1614 \\
        
        \cite{pointavatar} & Point-avatar & \colorno{$\times$}   & \colorno{$\times$}   & 26.84 & 0.8970 & 0.0926 & 25.14 & 0.8385 & 0.1278 \\
        
        \cite{splattingavatar} & Splatting-avatar & \colorno{$\times$}   & \colorno{$\times$}   & 28.71 & 0.9271 & 0.0862 & 26.66 & 0.8611 & 0.1351 \\
        
        \cite{flashavatar} & Flash-avatar & \colorno{$\times$}   & \colorno{$\times$}   & 29.13 & 0.9255 & 0.0719 & 27.58 & 0.8664 & 0.1095 \\
        
        \cite{gaussianblendshapes} & GBS & \colorno{$\times$}   & \colorno{$\times$}   & 29.64 & 0.9394 & 0.0823 & 27.81 & 0.8915 & 0.1297 \\
        
        \cite{flare} & FLARE & \coloryes{\checkmark} & \colorno{$\times$}     & 26.80 & 0.9063 & 0.0816 & 25.55 & 0.8479 & 0.1183 \\
        
        \cite{HRAvatar} & HRAvatar & \coloryes{\checkmark} & \colorno{$\times$}   & \ranktwo{30.36} & \ranktwo{0.9482} & \ranktwo{0.0569} & \ranktwo{28.55} & \ranktwo{0.9089} & \ranktwo{0.0825} \\
        
        & GTAvatar (ours)                         & \coloryes{\checkmark} & \coloryes{\checkmark} & \rankone{30.52} & \rankone{0.9537} & \rankone{0.0552} & \rankone{28.83} & \rankone{0.9130} & \rankone{0.0794} \\
        
        \bottomrule
    \end{tabular}
    \caption{Results of various methods for the self-reenactment task on the INSTA and HDTF datasets. Our method outperforms all others in PSNR, SSIM and LPIPS.}
    \label{tab:exp-reconstruction}
\end{table*}

\begin{figure*}[h]
    \newcommand{\w}{0.132\linewidth}
    \setlength{\tabcolsep}{0em} %
    \begin{tabular}{c@{\hskip -0.25in}cc@{\hskip -0.25in}cccc@{\hskip 0.1in}c}
        \multicolumn{2}{c}{\textbf{Reconstruction}} & \multicolumn{2}{c}{\textbf{Edit}} & \multicolumn{4}{c}{\textbf{Relight}}\\
        Normals & Render & Normals & Render & \multicolumn{3}{c}{Full model} & w/o normal map\\
        \includegraphics[width=\w,valign=c]{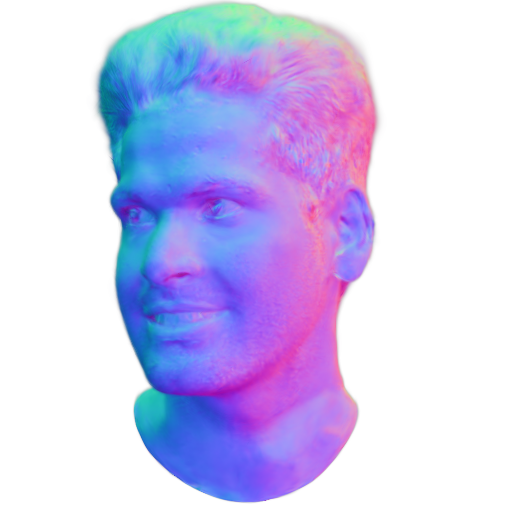} &
        \includegraphics[width=\w,valign=c]{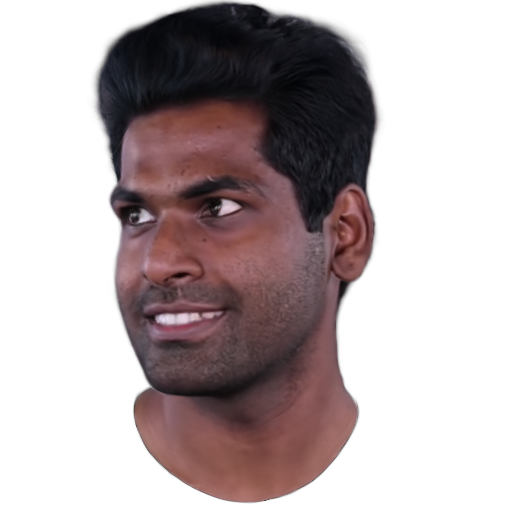} &
        \includegraphics[width=\w,valign=c]{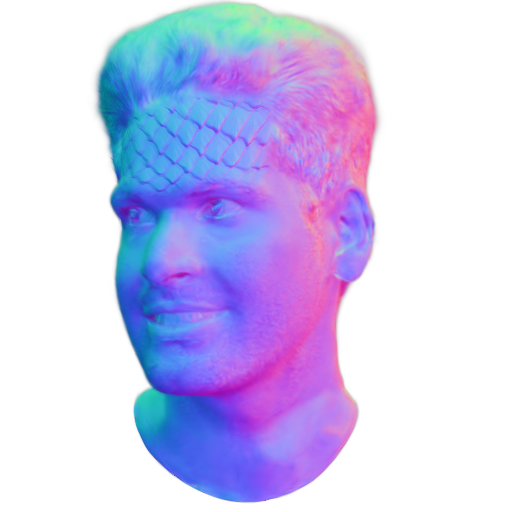} &
        \includegraphics[width=\w,valign=c]{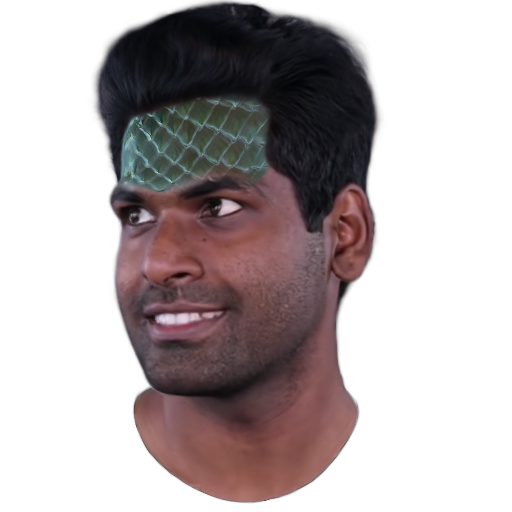} &
        \includegraphics[width=\w,valign=c]{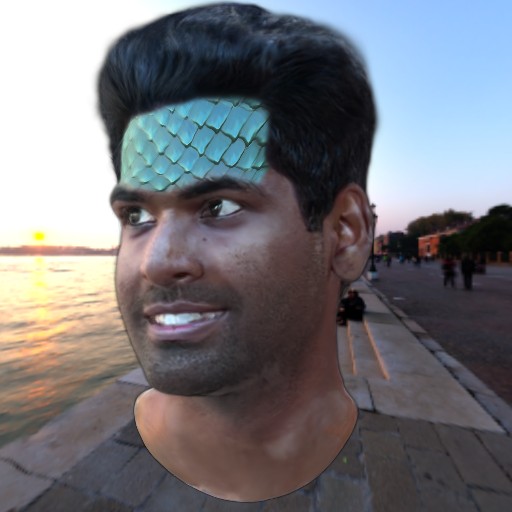} &
        \includegraphics[width=\w,valign=c]{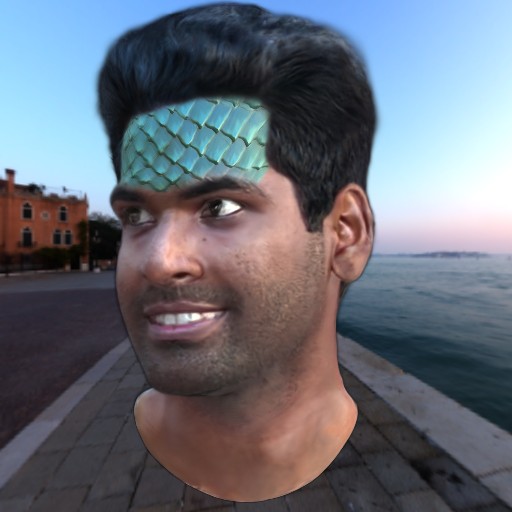} &
        \includegraphics[width=\w,valign=c]{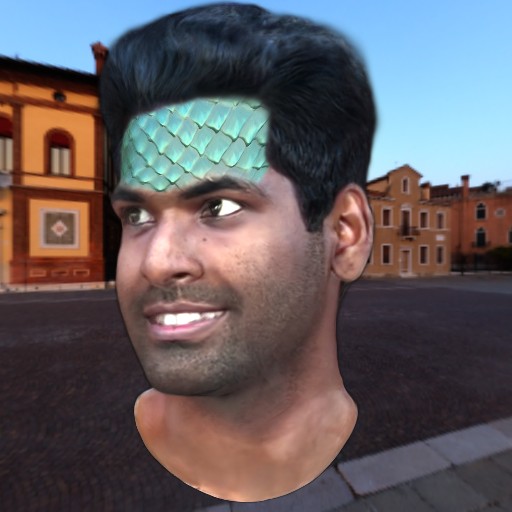} &
        \includegraphics[width=\w,valign=c]{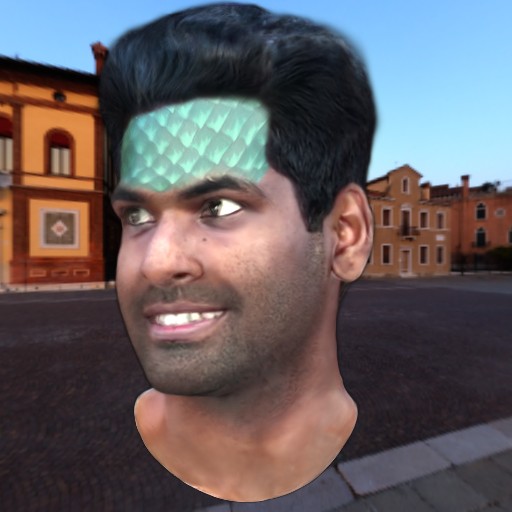}
        \\
        \includegraphics[width=\w,valign=c]{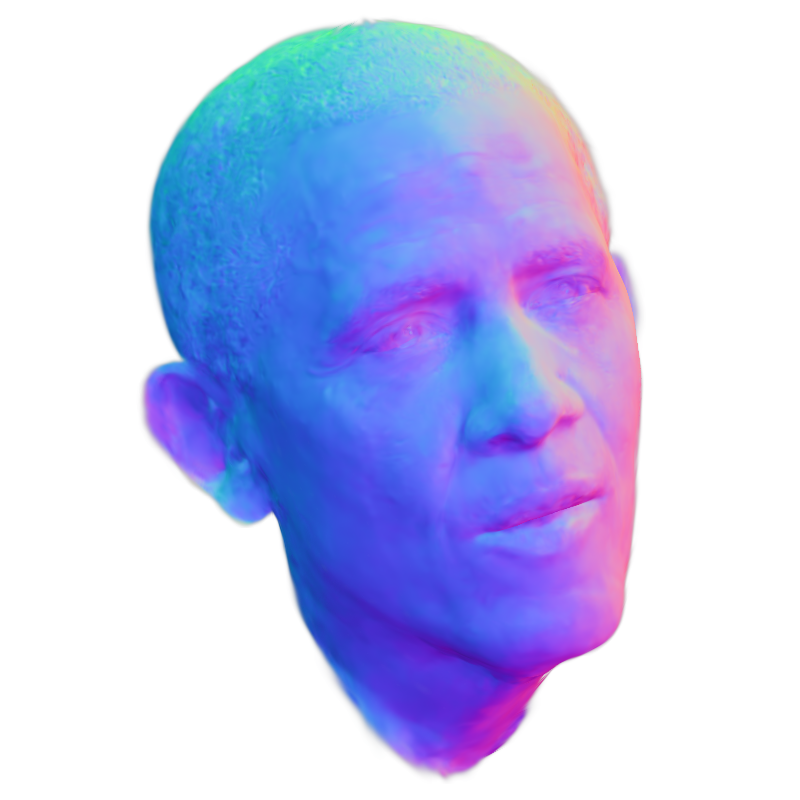} &
        \includegraphics[width=\w,valign=c]{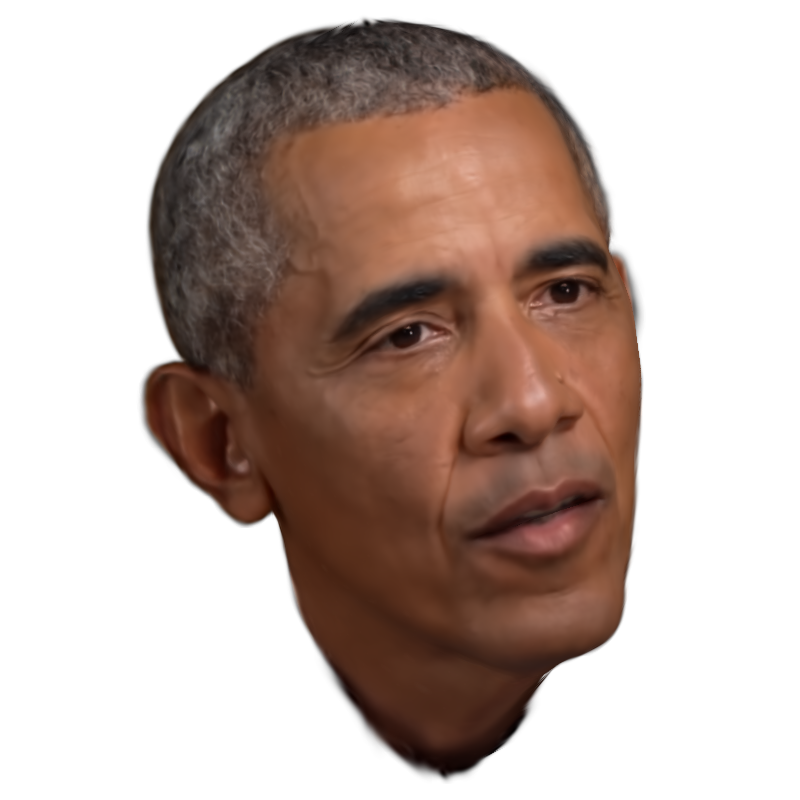} &
        \includegraphics[width=\w,valign=c]{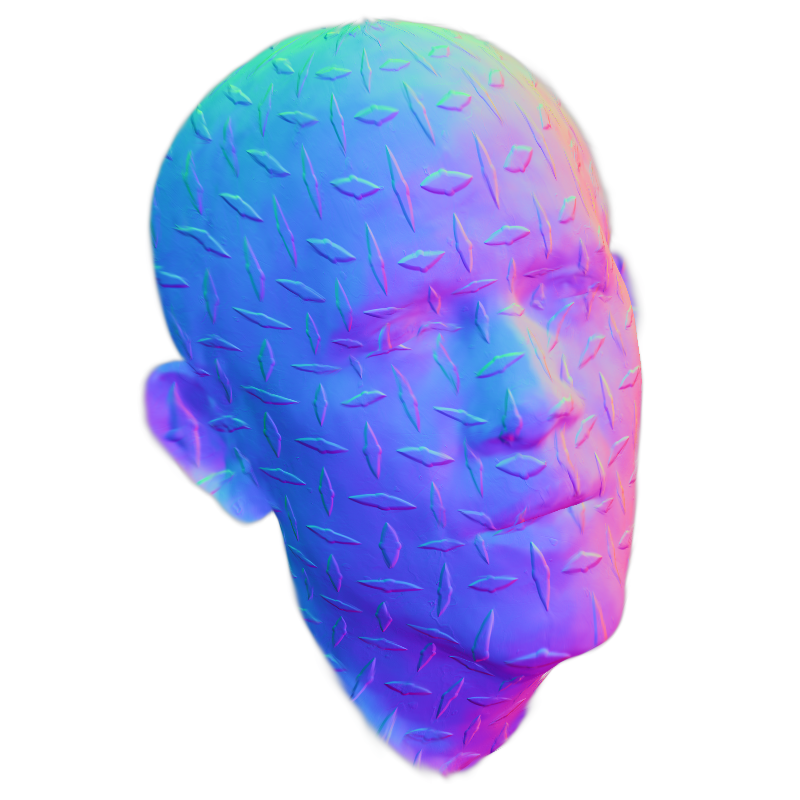} &
        \includegraphics[width=\w,valign=c]{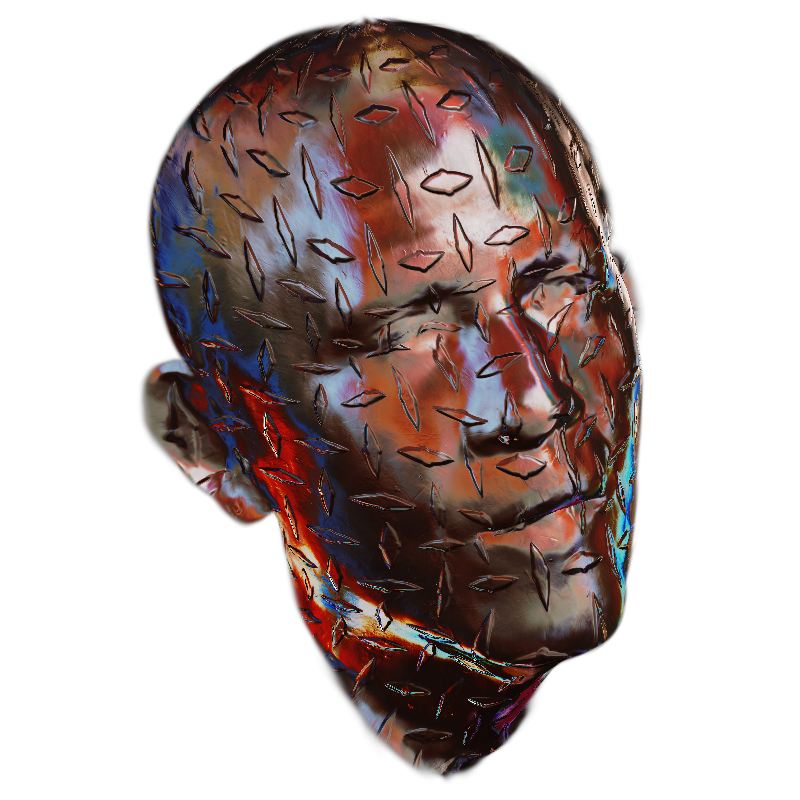} &
        \includegraphics[width=\w,valign=c]{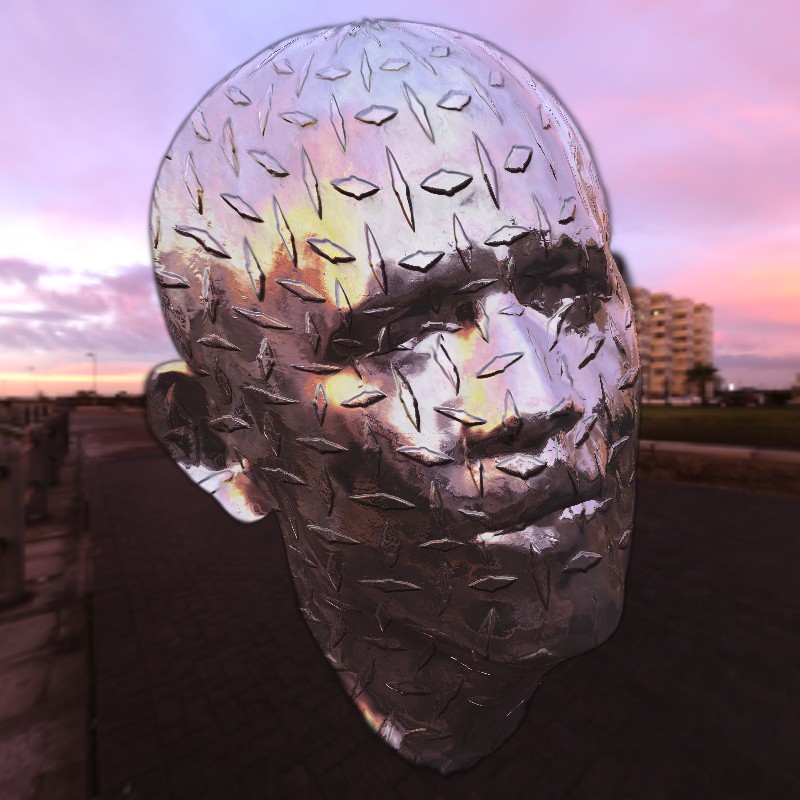} &
        \includegraphics[width=\w,valign=c]{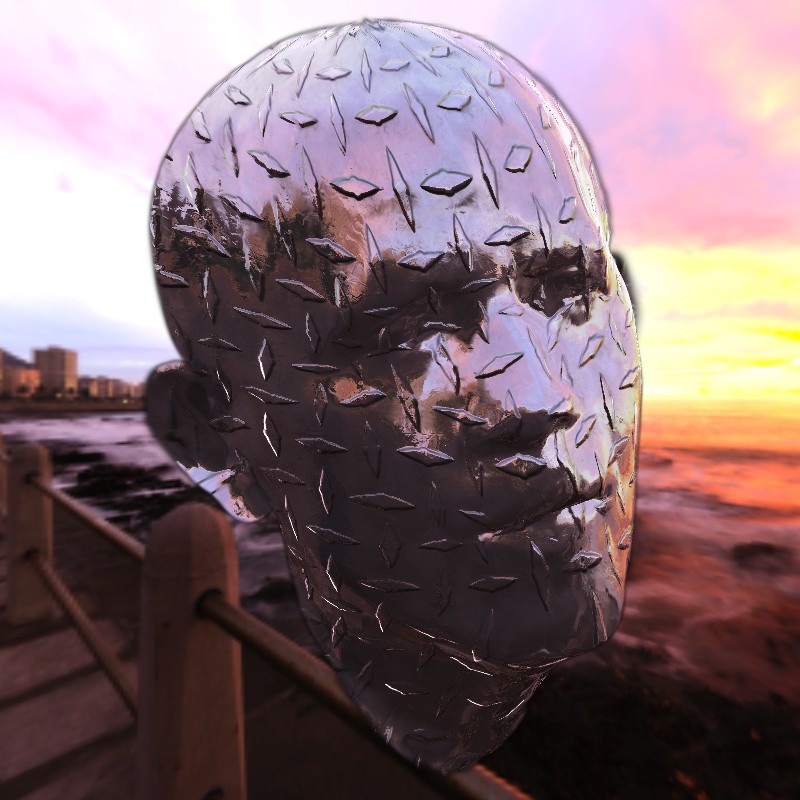} &
        \includegraphics[width=\w,valign=c]{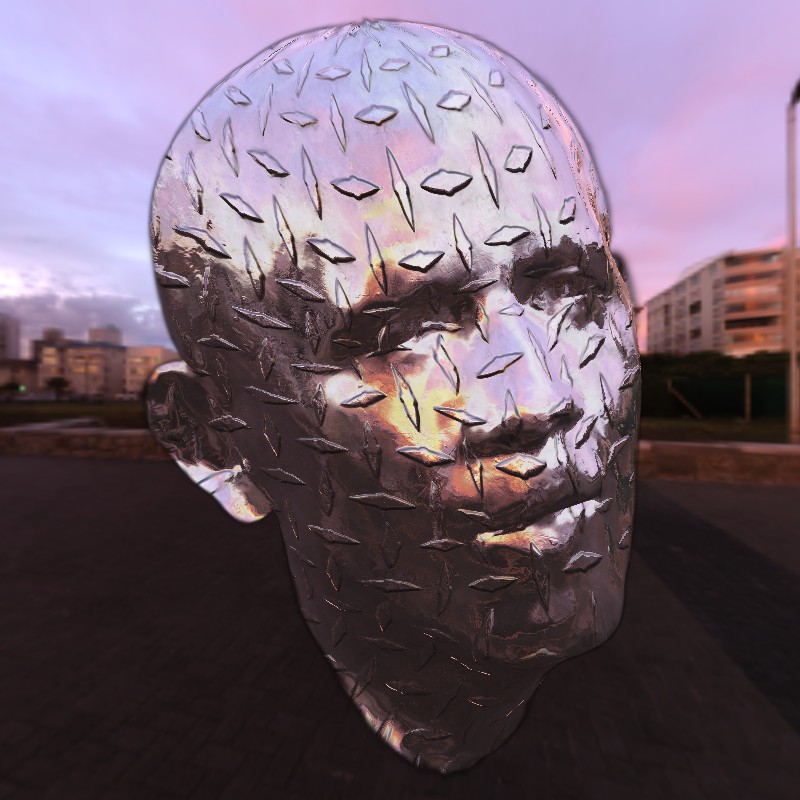} &
        \includegraphics[width=\w,valign=c]{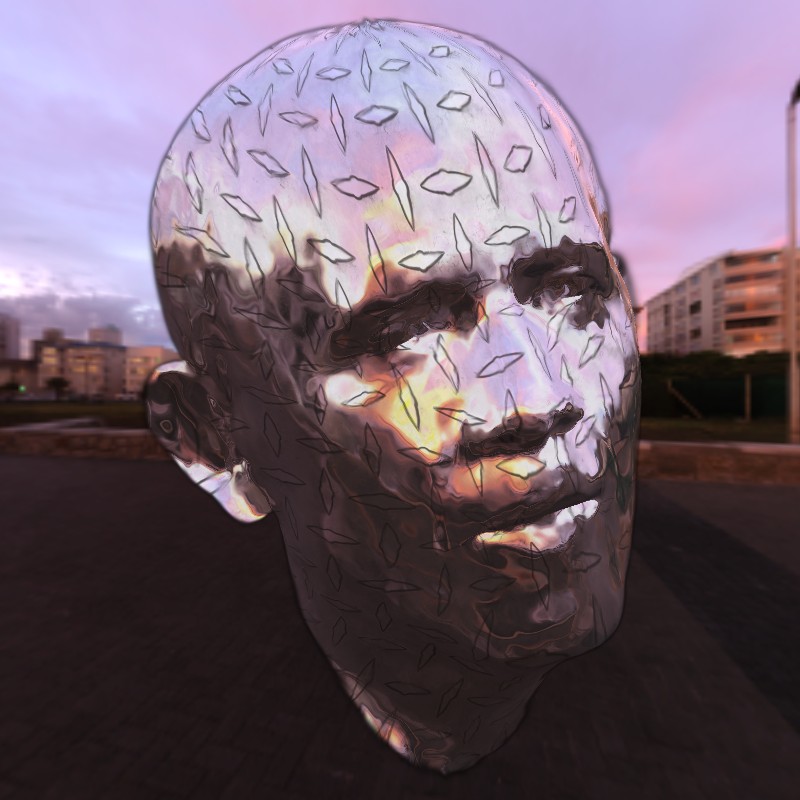}
    \end{tabular}
    \centering
    \caption{Texture editing with off-the-shelf PBR material maps. Our method enables consistent rendering of edited avatars under varying illumination with conventional material definitions. The last column underlines the importance of normal mapping for creating edits that interact with lighting convincingly.}
    \label{fig:exp-texture-editing-pbr}
\end{figure*}

\begin{figure}
    \centering
    \subfloat[LPIPS for varying number of Gaussians and texture resolutions.]{\includegraphics[width=\linewidth]{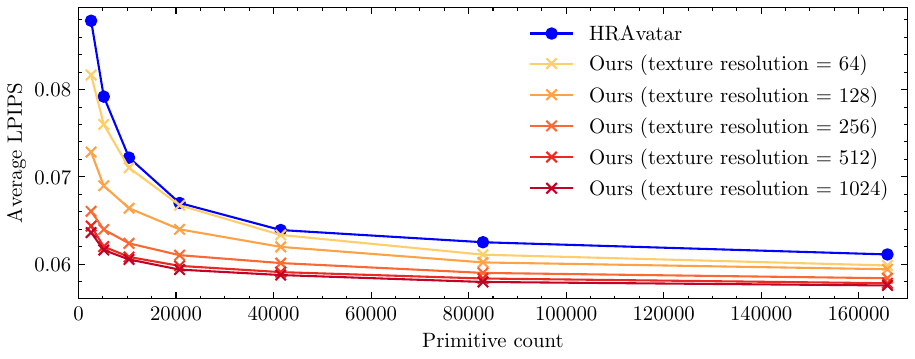}\label{fig:lpips_texres_gaussians}}
    \qquad
    \subfloat[LPIPS for varying number of learned parameters, accounting for Gaussians and texture resolution. For our method, we vary both simultaneously to find a trade-off between quality and model size. As indicated by the dotted line, our method achieves the same quality with significantly lower size.]{\includegraphics[width=\linewidth]{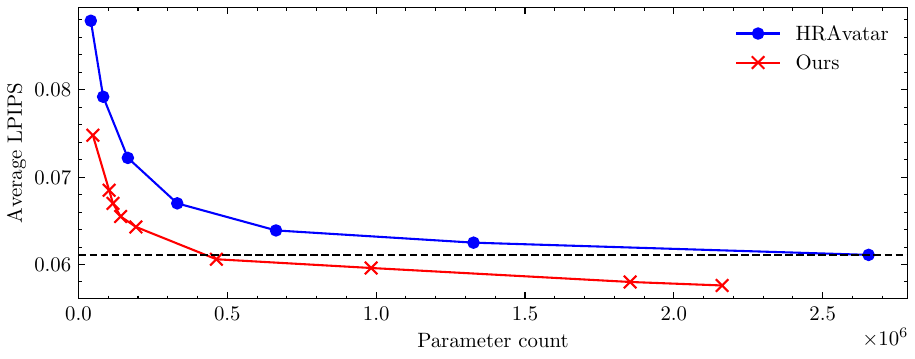}\label{fig:lpips_params}}
    \qquad
    \caption{Reconstruction quality on unseen frames averaged over 10 videos of the INSTA dataset, compared with HRAvatar~\cite{HRAvatar} (LPIPS, lower is better). Densification and pruning of Gaussians are disabled in these experiments for manual control of splat count.}
    \label{fig:lpips_texres_gaussians_params}
\end{figure}

\begin{figure}[h]
    \newcommand{\w}{0.26\linewidth}
    \setlength{\tabcolsep}{0.0em} %
    \begin{tabular}{c@{\hskip -0.1in}c@{\hskip -0.2in}c@{\hskip -0.2in}c@{\hskip -0.2in}c}
         & 16 & 64 & 256 & 1024
        \\
        \rotatebox[origin=c]{90}{normal} &
        \includegraphics[width=\w,valign=c]{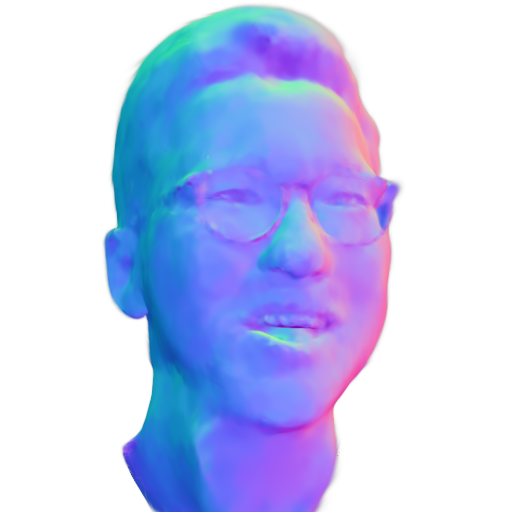} &
        \includegraphics[width=\w,valign=c]{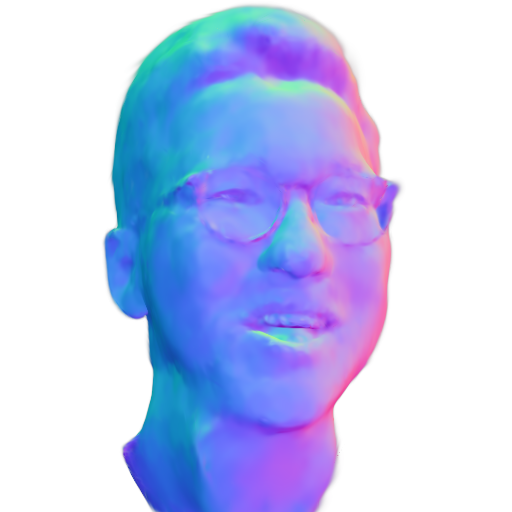} &
        \includegraphics[width=\w,valign=c]{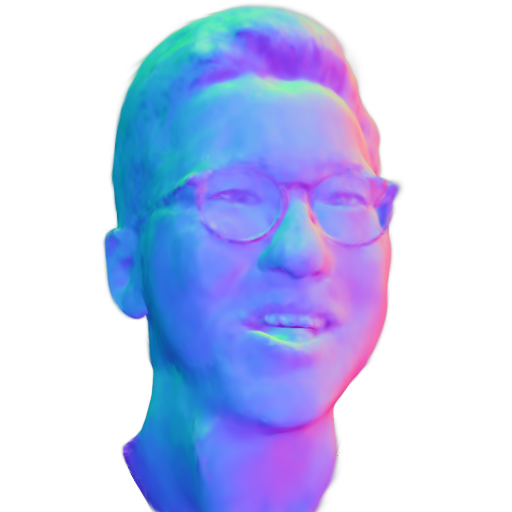} &
        \includegraphics[width=\w,valign=c]{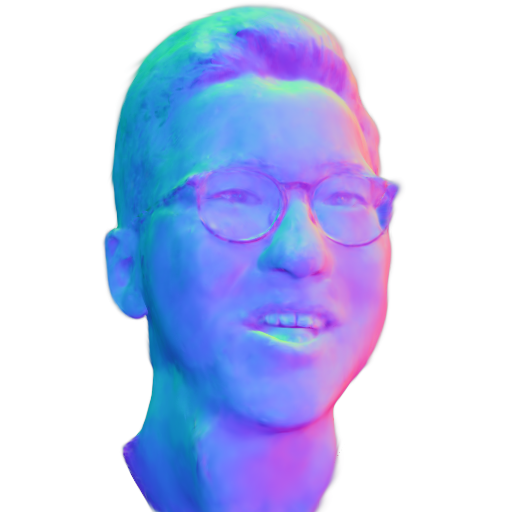}
        \\
        \rotatebox[origin=c]{90}{render} &
        \includegraphics[width=\w,valign=c]{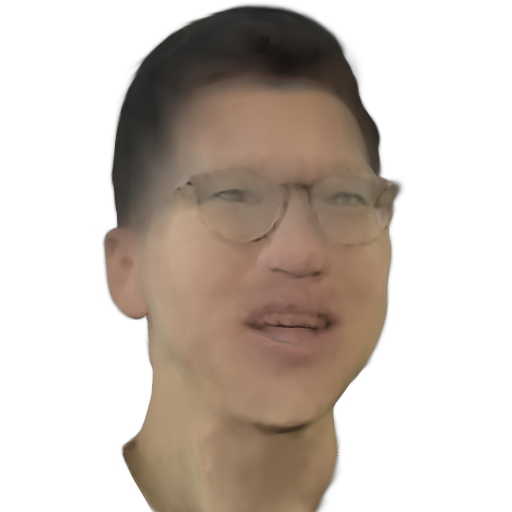} &
        \includegraphics[width=\w,valign=c]{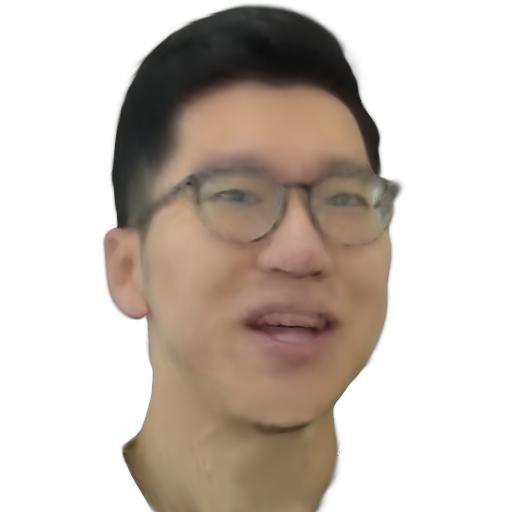} &
        \includegraphics[width=\w,valign=c]{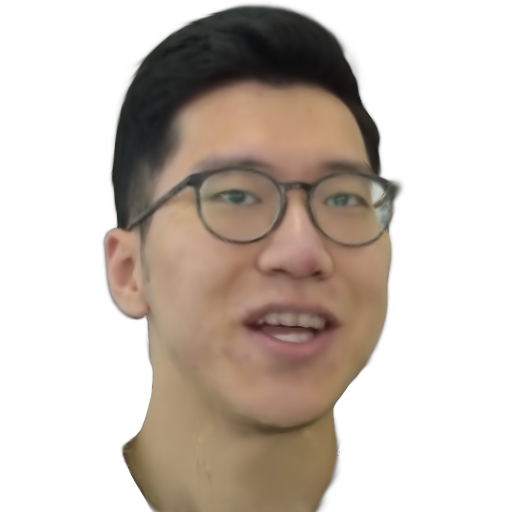} &
        \includegraphics[width=\w,valign=c]{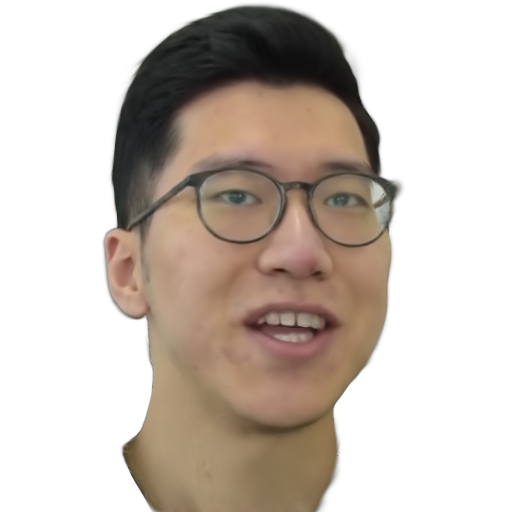}
    \end{tabular}
    \centering
    \caption{Normals and render for different levels of texture scaling after training with a $1024 \times 1024$ texture. As the texture is downscaled, the reconstruction quality decreases gradually.}
    \label{fig:exp-ablation-downscale-texture}
\end{figure}

\section{Training}
\label{sec:training}

Our model is trained end-to-end from a monocular RGB video. Following previous work \cite{pointavatar, flare, HRAvatar}, the input video is preprocessed prior to training to extract alpha masks using a matting method \cite{robustvideomatting} and initial FLAME parameters using an off-the-shelf facial tracker \cite{SMIRK}.
During training, we simultaneously optimize the parameters of our 2D Gaussians (barycentric coordinates, relative rotation, scale, displacement and opacity), the textures (albedo, roughness, specular reflectance and normals) and the scene's lighting environment map (a $6\times32^2\times3$ RGB cubemap). Gaussians undergo densification and pruning as in other Gaussian-based methods \cite{3DGS,2DGS,gaussianavatars,HRAvatar}. We also optimize the FLAME model (template mesh and deformation basis) for higher fidelity and fine-tune the expression encoder from the FLAME tracker~\cite{SMIRK} while continuously predicting updated expression parameters during training, following recent work~\cite{HRAvatar}.
We train for 15 epochs using Adam optimizer \cite{adam}, requiring one to two hours on an NVIDIA RTX A5000 GPU. Learning rates are given in Table~\ref{tab:learning-rates}.

Our total loss function is a weighted sum of terms designed to supervise the image reconstruction, guide the physical property disentanglement, regularize our novel UV mapping, and ensure geometric stability. The overall objective is:
\begin{equation}
\mathcal{L}_\text{total} = \mathcal{L}_\text{photo} + \mathcal{L}_\text{PBR} + \mathcal{L}_\text{uv} + \mathcal{L}_\text{geom}
\end{equation}
Values used for the weights referenced in this section are provided in Table~\ref{tab:loss-weights}.

\subsection{Image Reconstruction Losses}
The primary supervisory signal comes from comparing the rendered image with the ground truth. Let $I$ and $M$ be the ground-truth image and foreground mask for a given frame. Let $\hat{I}$ and $\hat{M}$ be our corresponding rendered image and alpha mask. The photometric loss $\mathcal{L}_\text{photo}$ is a combination of an L1 loss, a structural similarity (SSIM) loss and a mask loss.
\begin{equation}\begin{split}
    \mathcal{L}_\text{photo} =\; &
    \lambda_\text{L1}||I - \hat{I}||_1 +
    \lambda_\text{SSIM}(1 - \text{SSIM}(I, \hat{I})) \\
    & + \lambda_\text{mask}||M - \hat{M}||_1 
\end{split}\end{equation}

\subsection{Priors for Physical Disentanglement} \label{sec:pbr_priors}

Decomposing appearance into intrinsic physical properties from a monocular video is a highly ill-posed problem. To guide this decomposition, we introduce a set of priors, $\mathcal{L}_\text{PBR}$.

First, we supervise the rendered albedo $\hat{I}_\rho$ with a diffusion-based prior $I_{\rho}$ \cite{chen2024intrinsicanything} precomputed for every third frame of the video, as recent work has proven this effective for disentangling albedo and lighting \cite{HRAvatar}. 
$$
    \mathcal{L}_{\text{diff\_albedo}} = M ||I_{\rho} - \hat{I}_\rho||_1
$$

Second, we enforce smoothness on the learned material maps, which is a natural prior for skin appearance. We apply a total variation (TV) loss to the roughness ($\mathcal{T}_r$) and specular reflectance ($\mathcal{T}_{f_0}$) textures:
$$
    \mathcal{L}_{\text{smooth}} = \text{TV}(\mathcal{T}_r) + \text{TV}(\mathcal{T}_{f_0})
$$

Finally, we regularize the detailed normal map ($\mathcal{T}_n$) toward the default up-vector $[0,0,1]^T$, encouraging it to capture only necessary deviations from the base splat normal:
$$
    \mathcal{L}_{\text{normal\_reg}} = || \mathcal{T}_n - [0,0,1]^T ||_1
$$
The total PBR loss writes:
\begin{equation}\begin{split}
    \mathcal{L}_\text{PBR} =\; &
    \lambda_{\text{diff\_albedo}}\mathcal{L}_{\text{diff\_albedo}} +
    \lambda_{\text{smooth}}\mathcal{L}_{\text{smooth}} + \\
    & \lambda_{\text{normal\_reg}}\mathcal{L}_{\text{normal\_reg}}
\end{split}\end{equation}

\subsection{UV-Mapping Regularization} \label{sec:uv_reg}
A key goal of our method is to produce continuous, artifact-free texture maps suitable for editing. Since multiple semi-transparent Gaussian splats can contribute to a single pixel, it is crucial that they all map to a consistent location in the UV texture for a given ray. To enforce this, we introduce $\mathcal{L}_\text{uv}$, a set of novel regularization terms.

\noindent\textbf{UV Distortion Loss.}
Inspired by the depth distortion loss in 2DGS~\cite{2DGS}, we propose a UV distortion loss to concentrate the texture coordinates of all ray-splat intersections contributing to a single ray. For a given pixel ray, we minimize the weighted pairwise distance between the UV coordinates of all ray-splat intersections. Let $uv_i$ be the $(u,v)$ coordinate derived from the $i$-th splat intersection along a ray. The loss then writes:
$$
    \mathcal{L}_{\text{uv\_dist}} = \sum_{\text{rays}} \sum_{i,j} \omega_i \omega_j || uv_i - uv_j ||_2
$$
where $\omega_i$ is the volumetric blending weight of the $i$-th splat intersection: $\omega_i = \alpha_i \mathcal{G}_i \prod_{j=1}^{i-1}(1-\alpha_j\mathcal{G}_j)$, where $\alpha_i$ and $\mathcal{G}_i$ are the opacity and Gaussian value at the intersection respectively. This loss penalizes rays where multiple highly-weighted splats contribute to the same pixel but map to distant UV coordinates. Minimizing this term forces the UV projections of overlapping Gaussians to converge to a single, sharp point in the texture map, preventing blurry or "ghosting" artifacts and creating an editable texture.

\noindent\textbf{UV Mask Loss.} To prevent Gaussians from sampling invalid areas of the texture atlas (i.e. the gaps between UV islands), we introduce a boundary loss. We pre-compute a binary mask $M_{uv}$ that is 0 inside valid triangle regions of the UV map and 1 elsewhere. The loss regularizes the intensity of texels outside of valid areas:
$$
    \mathcal{L}_{\text{uv\_boundary}} = M_{uv}||\mathcal{T} - \mathcal{T}_{\text{init}}||_1
$$
where $\mathcal{T}_{\text{init}}$ is the initial value for that texture -- $[0,0,0]$ for albedo, $0.5$ for roughness, $0.05$ for specular reflectance and $[0,0,1]$ for normals.

\noindent\textbf{Statistical Albedo} The albedo texture ($\mathcal{T}_\rho$) is further regularized using the texture-space PCA albedo prior provided by FLAME. PCA coefficients $\omega_\rho$ are optimized during training, yielding pseudo ground-truth texture $\tilde{\mathcal{T}}_\rho = \tilde{\mathcal{T}}_{\rho,\text{mean}} + \omega_\rho\tilde{\mathcal{T}}_{\rho, \text{basis}}$, used in:
$$
    \mathcal{L}_{\text{stat\_albedo}} = ||\mathcal{T}_\rho - \tilde{\mathcal{T}}_\rho||_1
$$ 

\begin{figure}[h]
    \newcommand{\w}{0.23\linewidth}
    \newcommand{\spa}{\vspace{0.04in}}
    \setlength{\tabcolsep}{0.0em} %
    \def\arraystretch{0.8}{ %
    \begin{tabular}{cc@{\hskip 0.05in}c}
         Reconstruction
         & Texture edit
         & Rec. texture
        \spa{}\\
        
        \includegraphics[width=\w,valign=c]{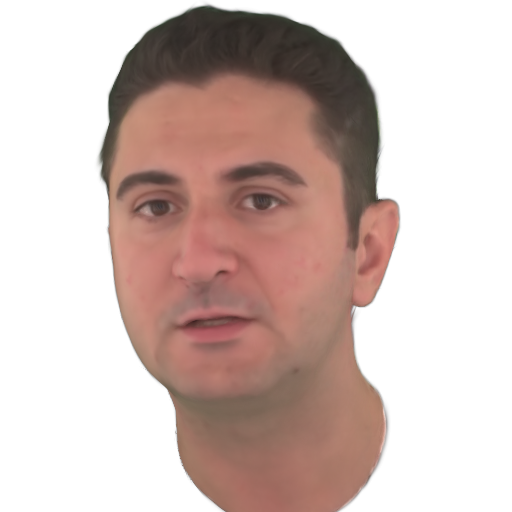} &
        \includegraphics[width=\w,valign=c]{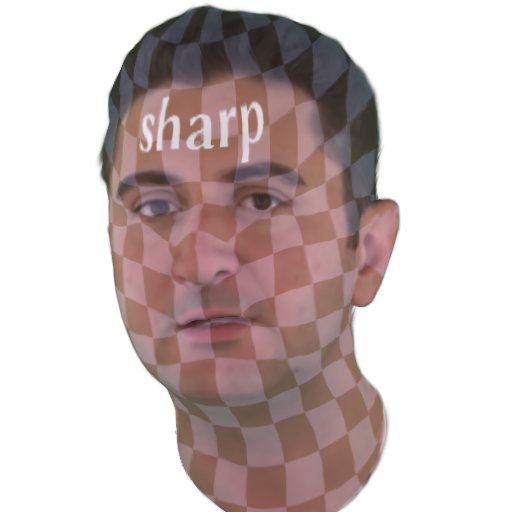} &
        \includegraphics[width=\w,valign=c]{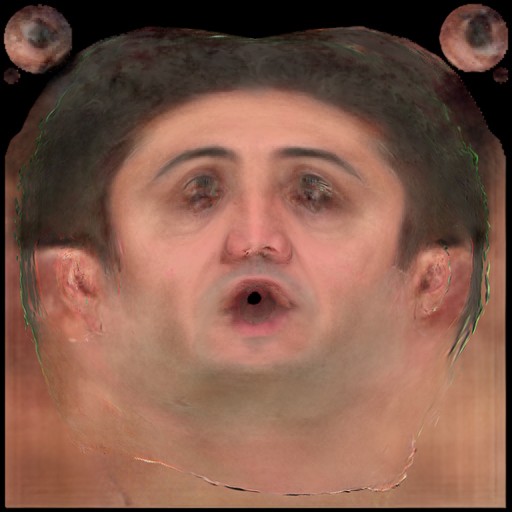}
        
        \spa{}\\
        \multicolumn{3}{c}{\small{Full method}}
        \spa{}\\
     
        \includegraphics[width=\w,valign=c]{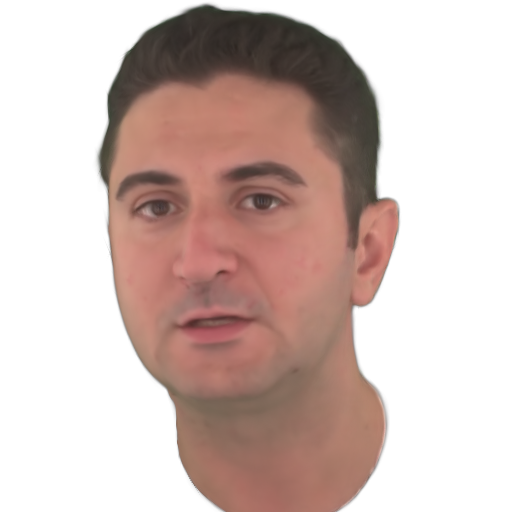} &
        \includegraphics[width=\w,valign=c]{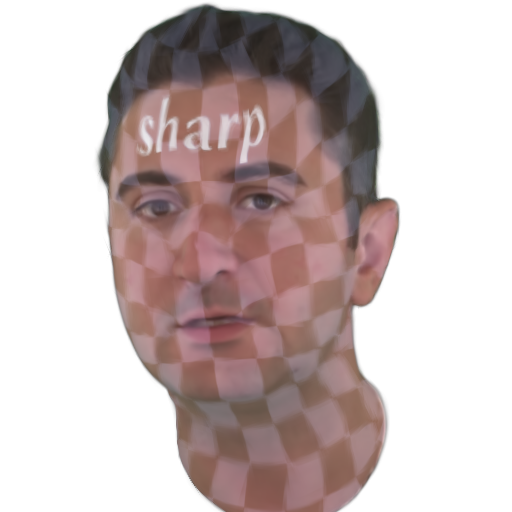} &
        \includegraphics[width=\w,valign=c]{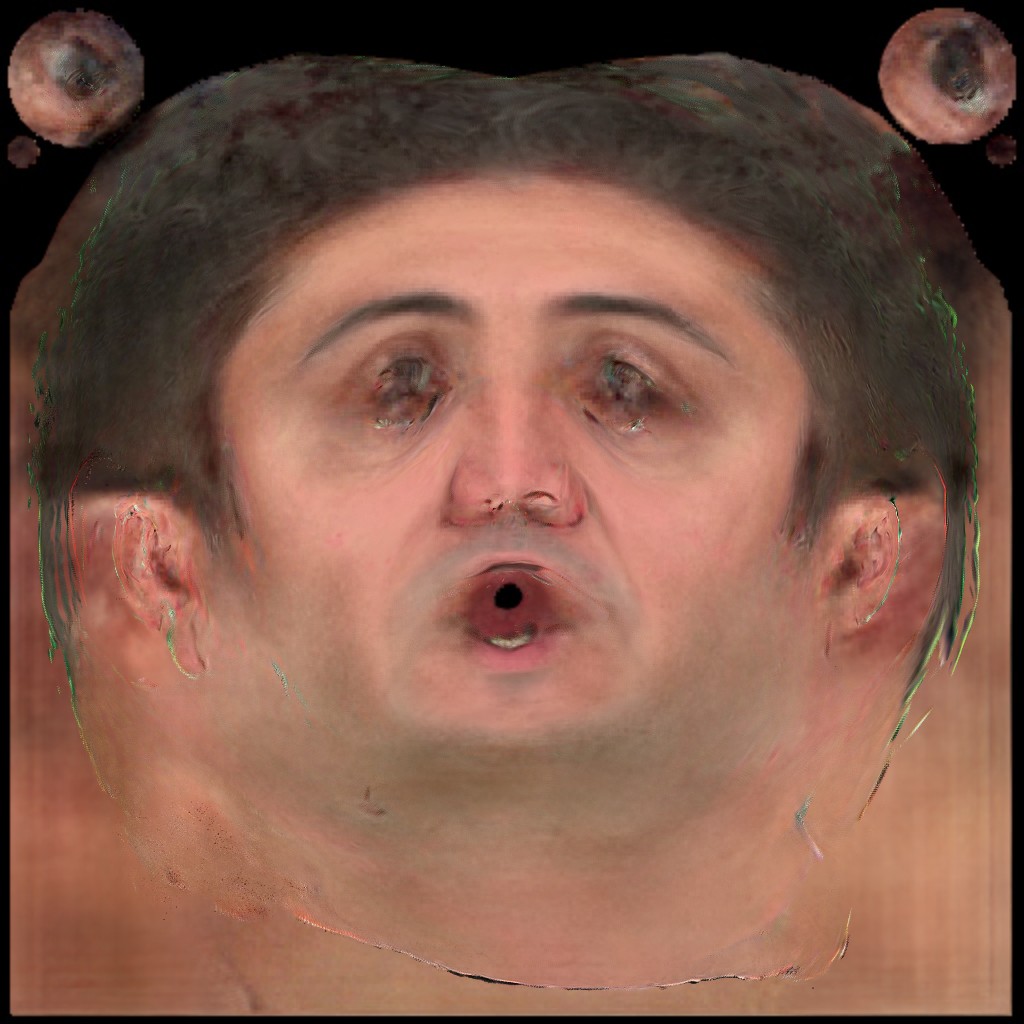}
        
        \spa{}\\
        \multicolumn{3}{c}{\small{w/o UV distortion loss}}
        \spa{}\\
        
        \includegraphics[width=\w,valign=c]{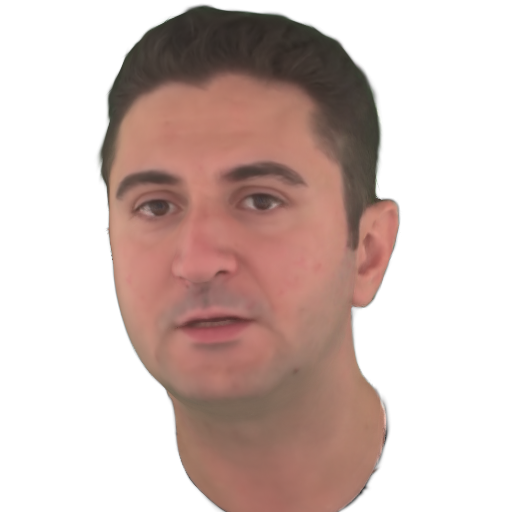} &
        \includegraphics[width=\w,valign=c]{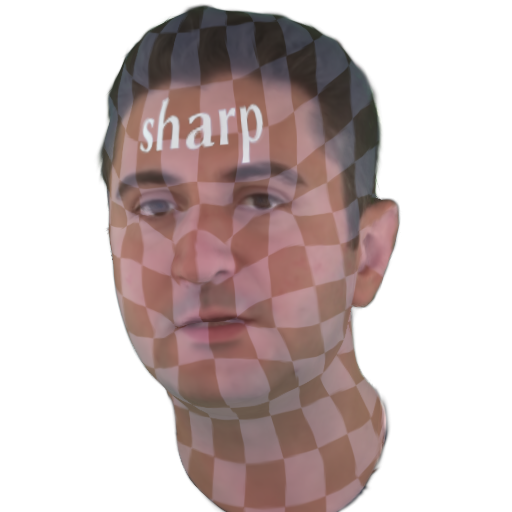} &
        \includegraphics[width=\w,valign=c]{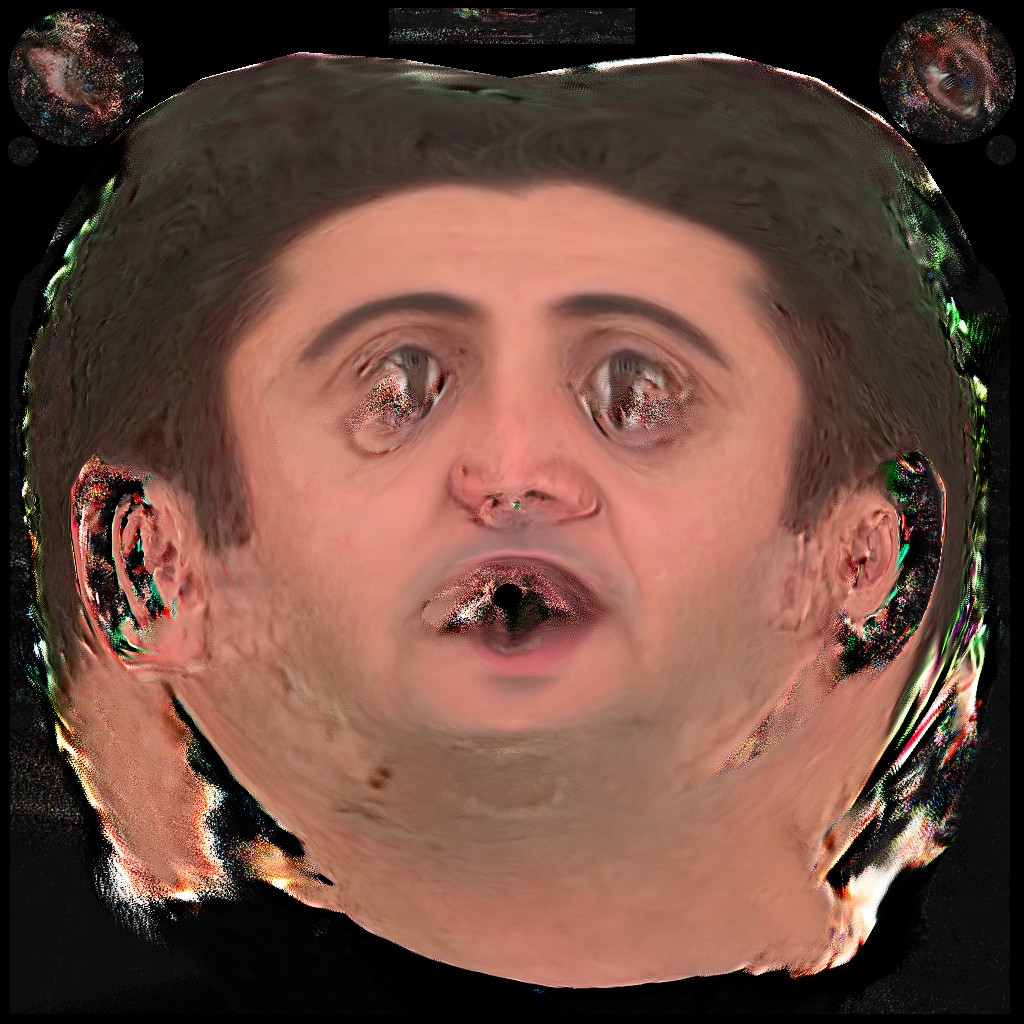}

        \spa{}\\
        \multicolumn{3}{c}{\small{w/o FLAME statistical albedo}}
        \spa{}\\
        
        \includegraphics[width=\w,valign=c]{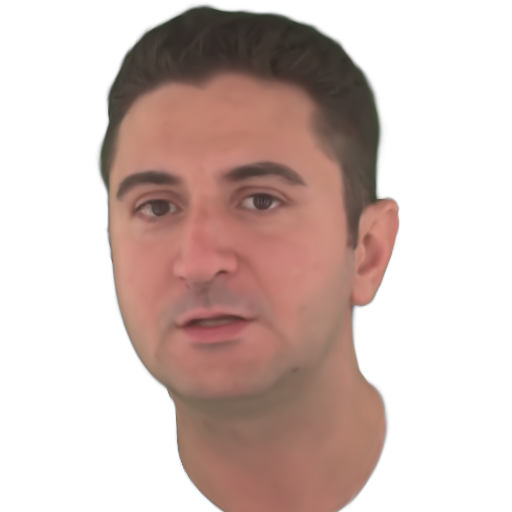} &
        \includegraphics[width=\w,valign=c]{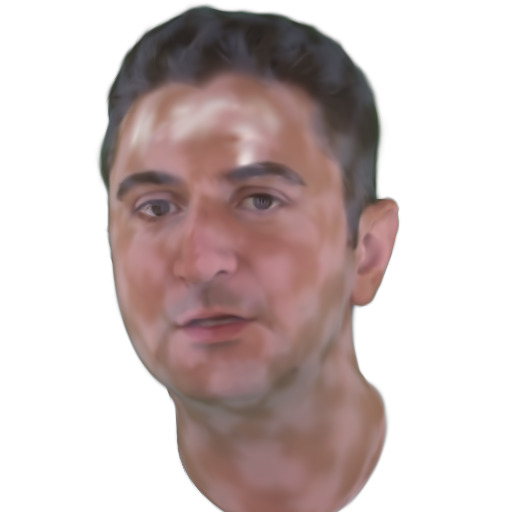} &
        \includegraphics[width=\w,valign=c]{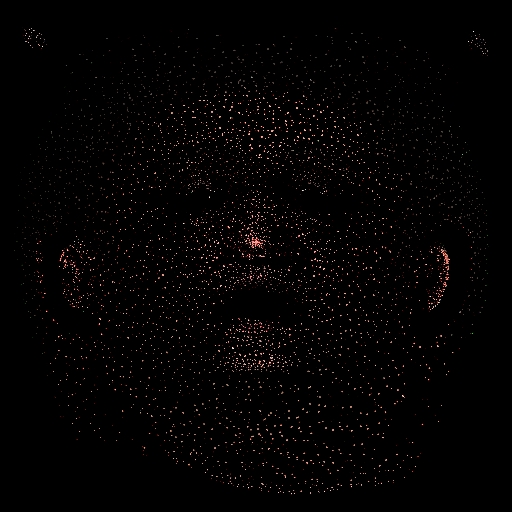}
        
        \spa{}\\
        \multicolumn{3}{c}{\small{w/o FLAME statistical albedo, $\Jstuv=\mathbf{0}$}}
    \end{tabular}
    }
    \centering
    \caption{Ablation of key aspects of our texturing approach. While the reconstruction quality remains similar, the UV distortion regularization yields a sharper UV mapping and helps maintain high-frequency details from the texture. The statistical albedo smooths the texture and reduces artifacts. In the last row, we set \Jstuv{} to zero (see Section \ref{sec:method-uvmapping}), resulting in a discontinuous texture that only uses a sparse set of texels. Please zoom in for details.}
    \label{fig:exp-ablation-texturing}
\end{figure}

Together, $\mathcal{L}_\text{uv} = \lambda_{\text{uv\_dist}}\mathcal{L}_{\text{uv\_dist}} + \lambda_{\text{boundary}}\mathcal{L}_{\text{uv\_boundary}} + \lambda_{\text{stat\_albedo}}\mathcal{L}_{\text{stat\_albedo}}$. These terms are critical for producing high-quality editable textures as shown in Figure~\ref{fig:exp-ablation-texturing}.

\subsection{Geometric and Gaussian Regularization}
Finally, to maintain a stable and plausible underlying structure, we use a set of geometric regularizers, $\mathcal{L}_\text{geom}$.

\noindent\textbf{Surface Normal Consistency.} 
To ensure that our 2D splats align locally with the overall reconstructed surface, we adopt the normal consistency loss from 2DGS~\cite{2DGS}. This loss aligns the normal of each individual splat $\mathbf{n}_i$ with the macro-scale surface normal $\mathbf{N}$, derived from the screen-space gradients of the rendered depth map. The loss for each ray is expressed as:
$$
    \mathcal{L}_{\text{normal\_consist}} = \sum_i \omega_i (1 - \mathbf{n}_i^\top \mathbf{N})
$$
This encourages the formation of a smooth, coherent geometric surface, which is an important foundation for a stable UV mapping.

\noindent\textbf{FLAME mesh regularization.} 
Following prior work on drivable avatars \cite{flare}, we maintain the structure of the underlying mesh through a Laplacian smoothing loss that regularizes the offsets between the original and the fine-tuned FLAME mesh: $\mathcal{L}_\text{lap} = || L(\mathcal{F}(V_t , \Psi) - \tilde{\mathcal{F}}(\tilde{V_t}, \tilde{\Psi})) ||_{2}^2 $
where $\tilde{\mathcal{F}}$ denotes FLAME with optimized basis, $\tilde{V_t}$ the tuned template vertices, $\tilde{\Psi}$ the FLAME parameters with expression computed using the tuned encoder and $L$ the graph Laplacian of the mesh \cite{sorkine2005laplacian}. We additionally regularize the difference in all attributes of FLAME with a L2 loss $\mathcal{L}_\text{FLAME}$.
Finally, we bias the predicted expression parameters towards the initial values to prevent excessive divergence from the initial tracker: $\mathcal{L}_{\text{expr}} = ||\tilde\Psi_\text{expression} - \Psi_\text{expression}||_2^2$. 
Together, these losses ensure our underlying mesh maintains geometric integrity and the surface UV mapping is preserved.

Lastly, we encourage Gaussian primitives to remain near the center of their parent triangle by regularizing their barycentric coordinates $\mathbf{b}$: $\mathcal{L}_{\text{bary}} = || \mathbf{b} - [\frac{1}{3}, \frac{1}{3}, \frac{1}{3}]^T ||_2^2$.

In summary, the total geometric loss is $\mathcal{L}_\text{geom} = \lambda_{\text{normal}}\mathcal{L}_{\text{normal\_consist}} + \lambda_{\text{lap}}\mathcal{L}_{\text{lap}} + \lambda_{\text{FLAME}}\mathcal{L}_{\text{FLAME}} + \lambda_{\text{expr}}\mathcal{L}_{\text{expr}} + \lambda_{\text{bary}}\mathcal{L}_{\text{bary}}$.

\begin{figure}[h]
    \newcommand{\w}{0.26\columnwidth}
    \setlength{\tabcolsep}{0.2em} %
    \def\arraystretch{0.8}{ %
    \begin{tabular}{c@{\hskip -0.5em}c@{\hskip -0.5em}cc}
        & Normal & Close-up & Relight
        \\
        \rotatebox[origin=c]{90}{without map} &
        \includegraphics[width=\w,valign=c]{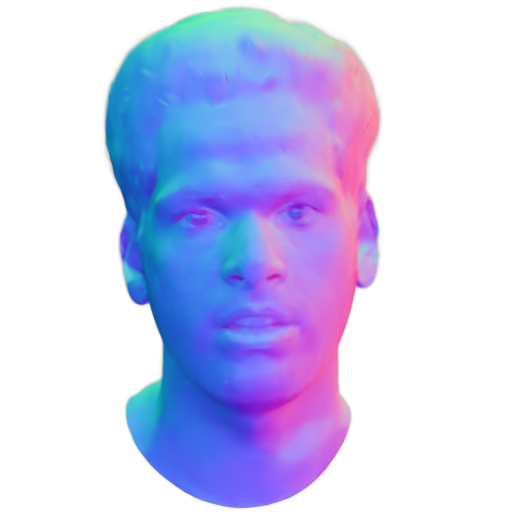} &
        \includegraphics[width=\w,valign=c]{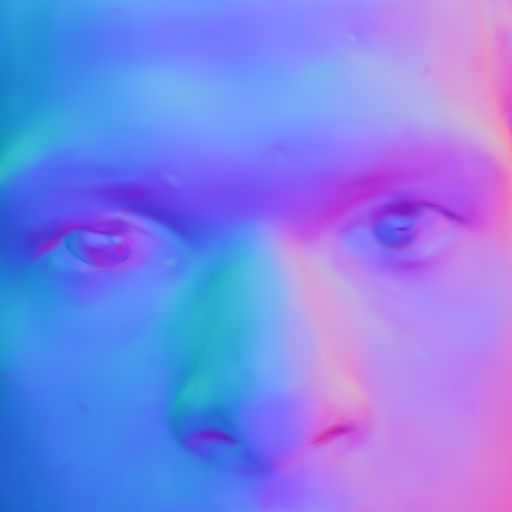} &
        \includegraphics[width=\w,valign=c]{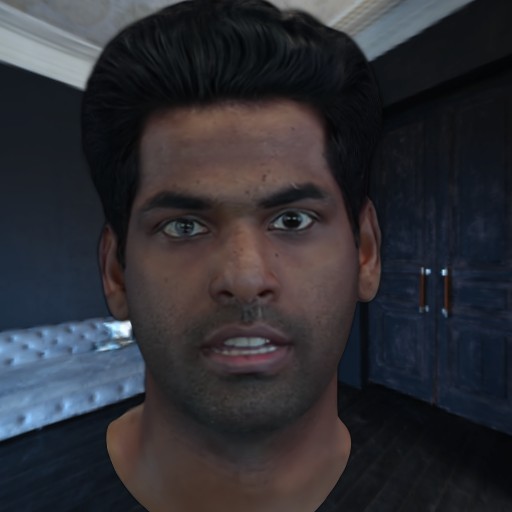} 
        \\
        \rotatebox[origin=c]{90}{with map} &
        \includegraphics[width=\w,valign=c]{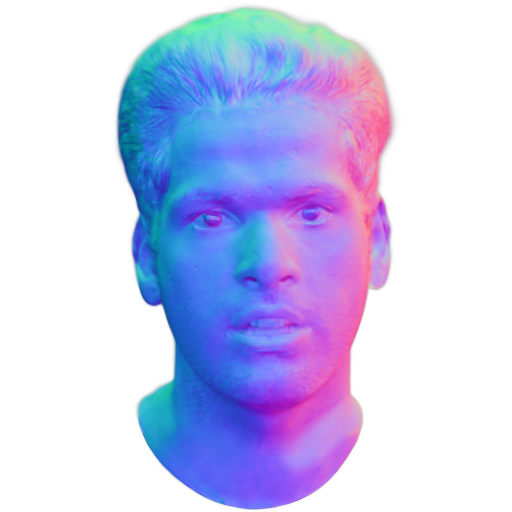} &
        \includegraphics[width=\w,valign=c]{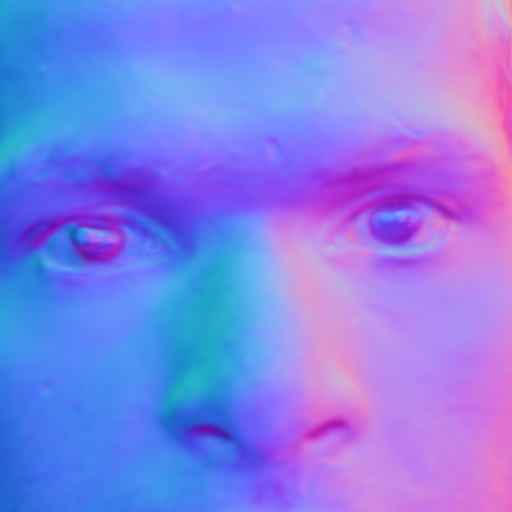} &
        \includegraphics[width=\w,valign=c]{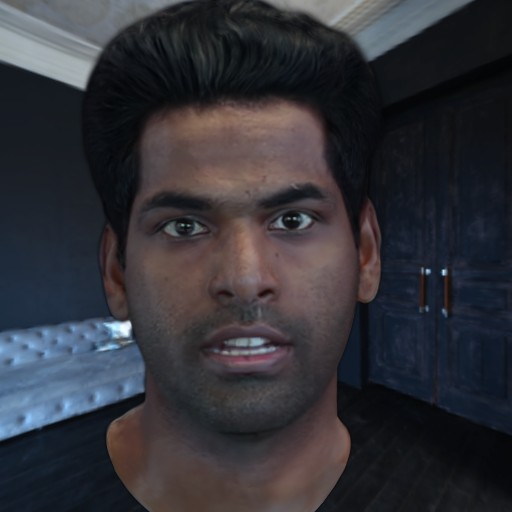} 
    \end{tabular}
    }
    \centering
    \caption{Ablation test on the normal map. 3D reconstruction is performed without relying on the normal map, \emph{i.e.} normals are only computed from the Gaussian splats and fine surface details are baked in the other channels (top row), compared to reconstruction with a normal map (bottom row).}
    \label{fig:exp-ablation-normal-map}
\end{figure}

\section{Results}

\noindent \textbf{Dataset.}  We evaluate various aspects of our method on two common datasets for monocular avatar reconstruction: INSTA~\cite{zielonka2022insta} and HDTF~\cite{zhang2021flow}. Both provide 2-3 minute $512 \times 512$ resolution talking head videos. Our experimental setting aligns with that of our main baseline, HRAvatar~\cite{HRAvatar}: we use 10 subjects from the INSTA dataset, and 8 from HDTF. For self-reenactment evaluation, the last 350 frames are left out of training for the former, and the last 500 frames for the latter.

\subsection{Self-reenactment} \label{sec:exp-reconstruction}

\begin{figure}[h]
    \newcommand{\w}{0.23\linewidth}
    \newcommand{\sk}{\hskip -0.06\linewidth}
    \newcommand{\skh}{\hskip -0.04\linewidth}
    \setlength{\tabcolsep}{0.0em} %
    \begin{tabular}{@{\sk}c@{\skh}c@{\skh}c@{\sk}c@{\sk}c@{\sk}c@{\sk}c@{\sk}c}
        Ground truth & Rec. & \multicolumn{4}{c}{Novel views}
        \\
        \includegraphics[width=\w,valign=c]{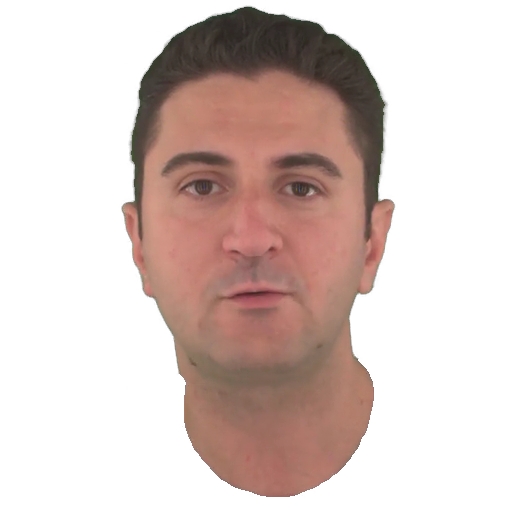} &
        \includegraphics[width=\w,valign=c]{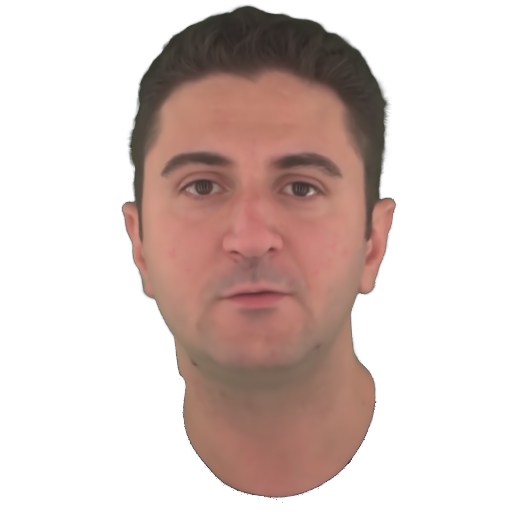} &
        \includegraphics[width=\w,valign=c]{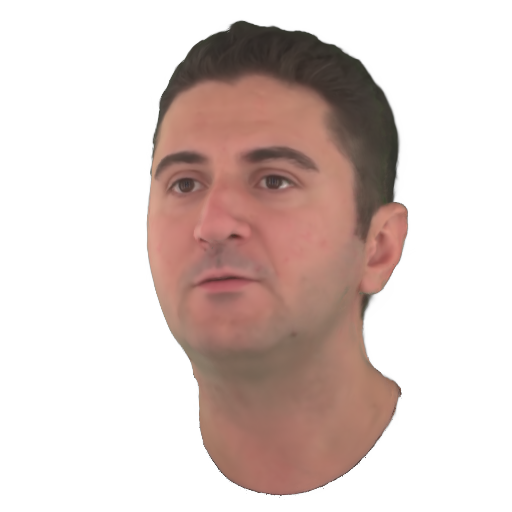} &
        \includegraphics[width=\w,valign=c]{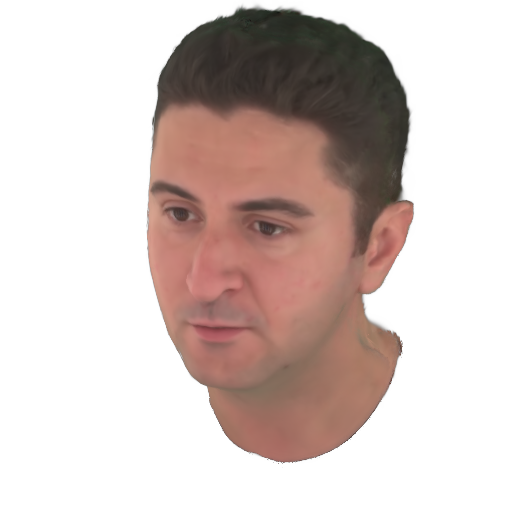} &
        \includegraphics[width=\w,valign=c]{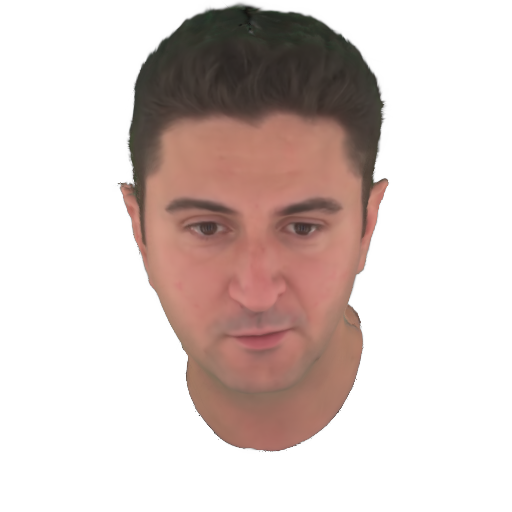} &
        \includegraphics[width=\w,valign=c]{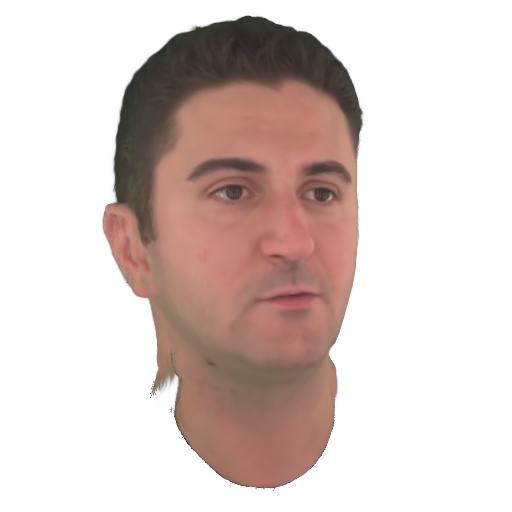}
        \\
        \includegraphics[width=\w,valign=c]{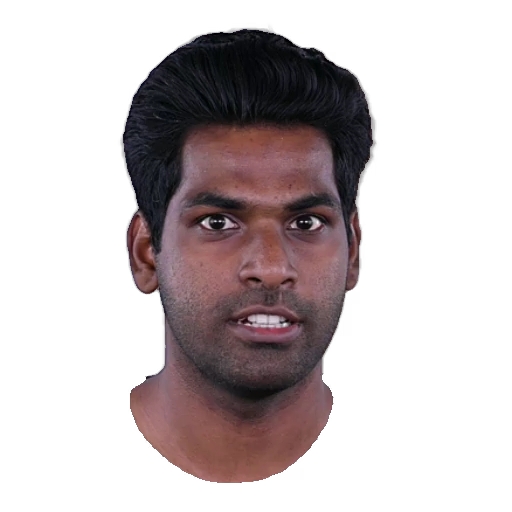} &
        \includegraphics[width=\w,valign=c]{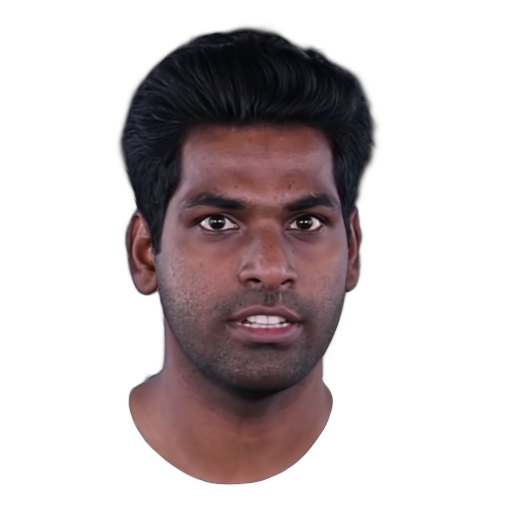} &
        \includegraphics[width=\w,valign=c]{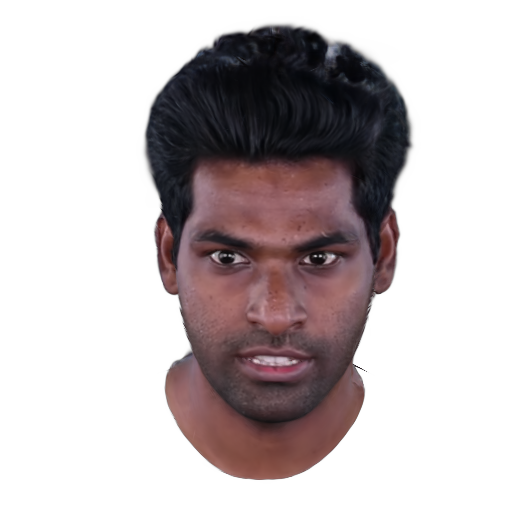} &
        \includegraphics[width=\w,valign=c]{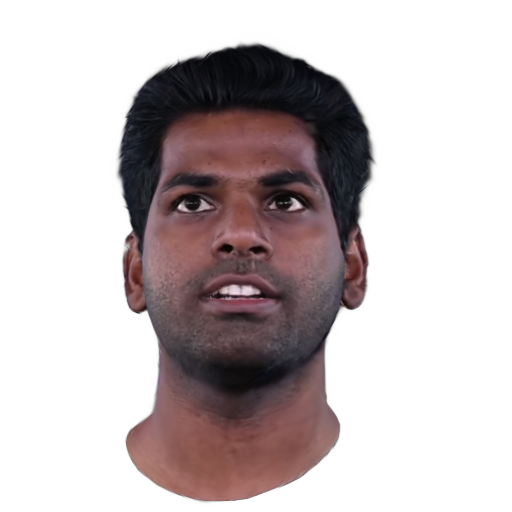} &
        \includegraphics[width=\w,valign=c]{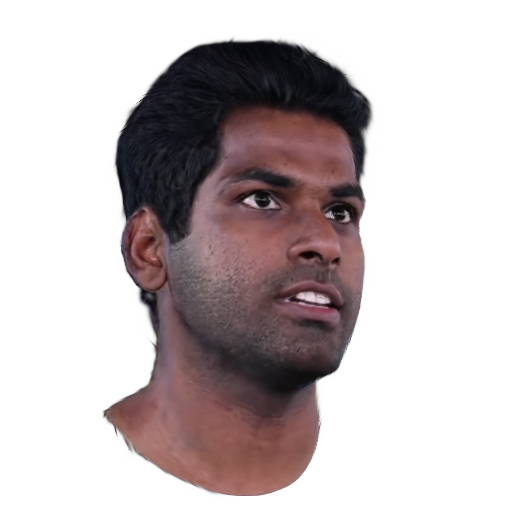} &
        \includegraphics[width=\w,valign=c]{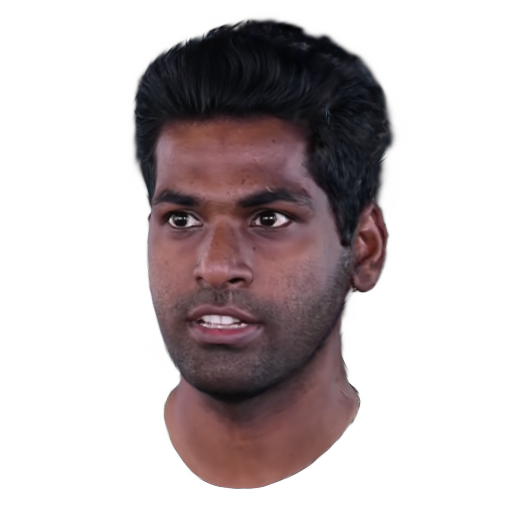}
        \\
        \includegraphics[width=\w,valign=c]{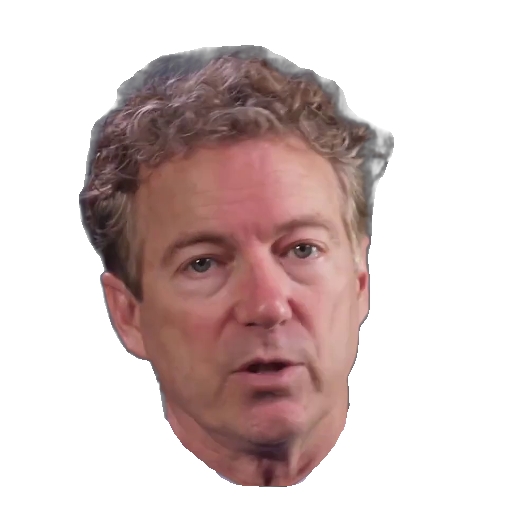} &
        \includegraphics[width=\w,valign=c]{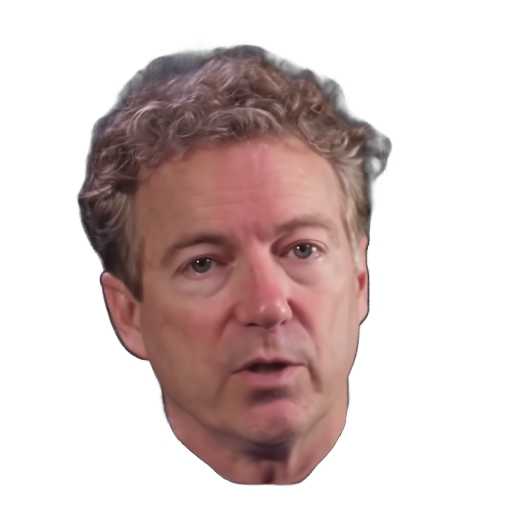} &
        \includegraphics[width=\w,valign=c]{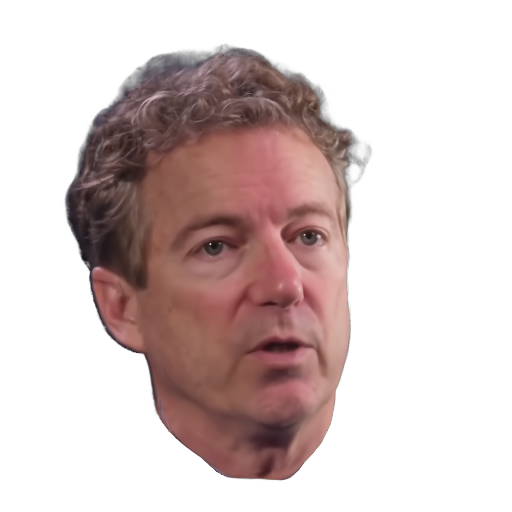} &
        \includegraphics[width=\w,valign=c]{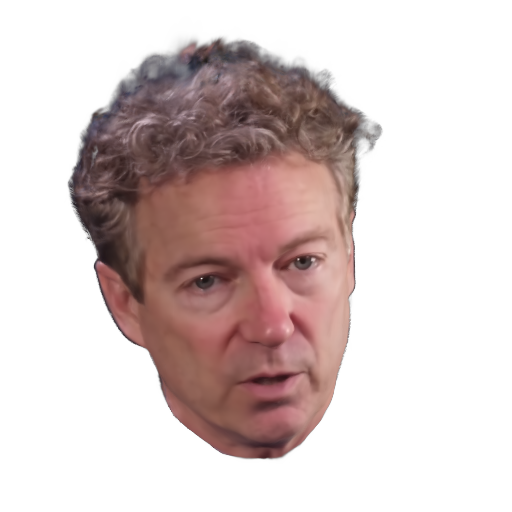} &
        \includegraphics[width=\w,valign=c]{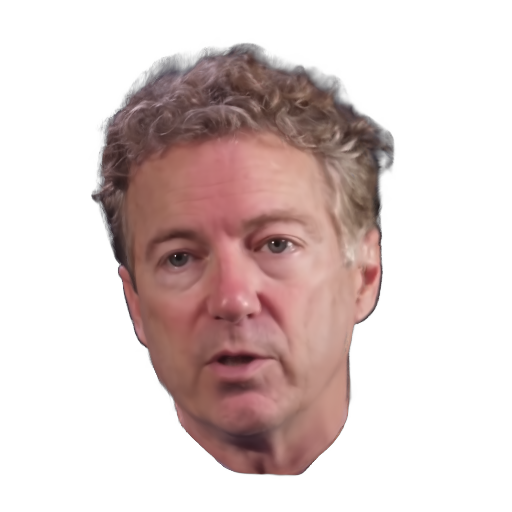} &
        \includegraphics[width=\w,valign=c]{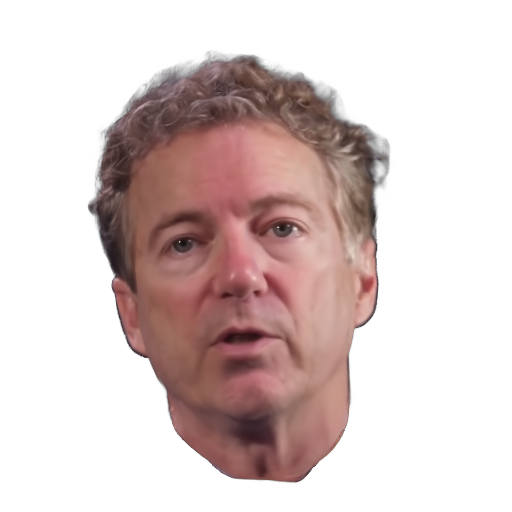}
        \\
    \end{tabular}
    \centering
    \caption{Examples of reconstructions and novel viewpoints.}
    \label{fig:novel_view}
\end{figure}

\begin{figure}[h]
    \newcommand{\w}{0.3\linewidth}
    \setlength{\tabcolsep}{0em} %
    \begin{tabular}{ccc}
        FATE (55k) & FATE (93k) & Ours (10k) \\
        \includegraphics[width=\w,valign=c]{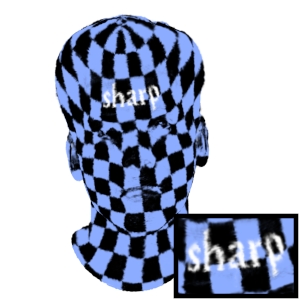} &
        \includegraphics[width=\w,valign=c]{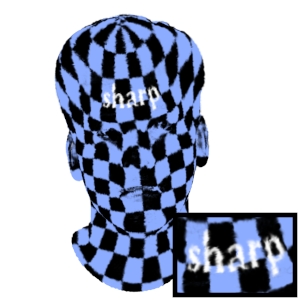} &
        \includegraphics[width=\w,valign=c]{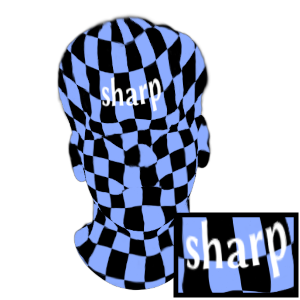}
    \end{tabular}
    \centering
    \caption{Visual comparison of our results with FATE~\cite{fate}. While both approaches provide texture editing features, our UV mapping technique yields sharper renders with high-frequency details. Numbers indicate the respective number of splats.}
    \label{fig:fate_comparison}
\end{figure}

\begin{figure*}[!hbtp]
    \centering
    \newcommand{\w}{0.098\linewidth}
    \setlength{\tabcolsep}{0.0em} %
    \begin{tabular}{cc@{\hskip -0.1in}c@{\hskip -0.1in}cccc@{\hskip 0.05in}c}
        Reference & Render & Normal & Albedo & \multicolumn{3}{c}{Environment map relighting}

        \\
        &
        \includegraphics[width=\w,valign=c]{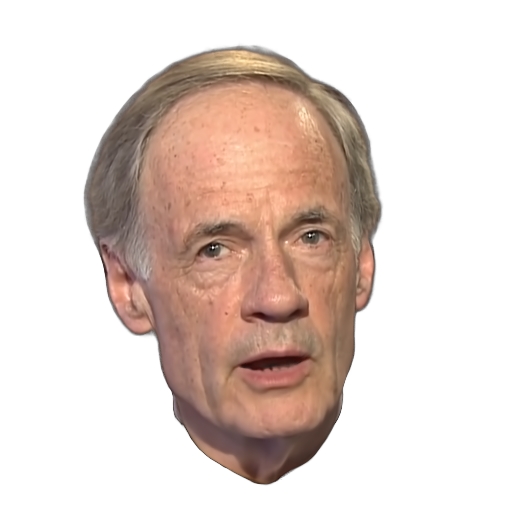} &
        \includegraphics[width=\w,valign=c]{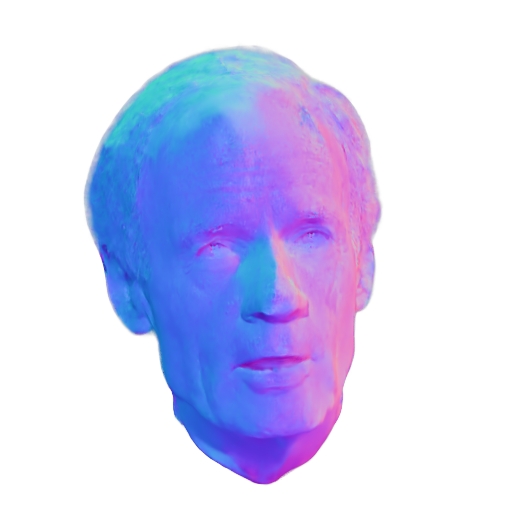} &
        \includegraphics[width=\w,valign=c]{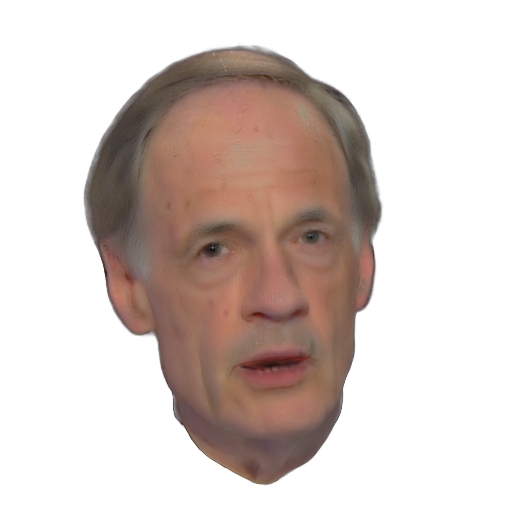} &
        \includegraphics[width=\w,valign=c]{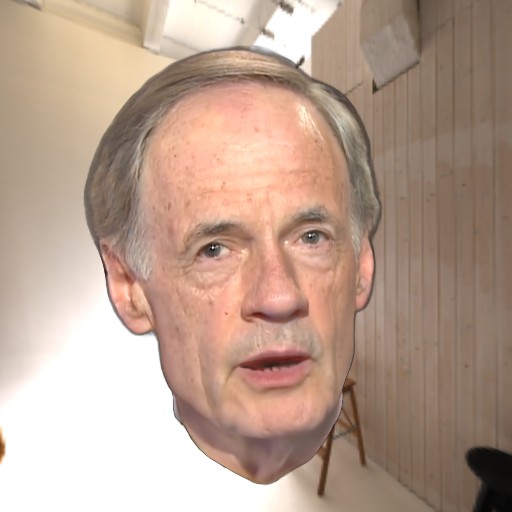} &
        \includegraphics[width=\w,valign=c]{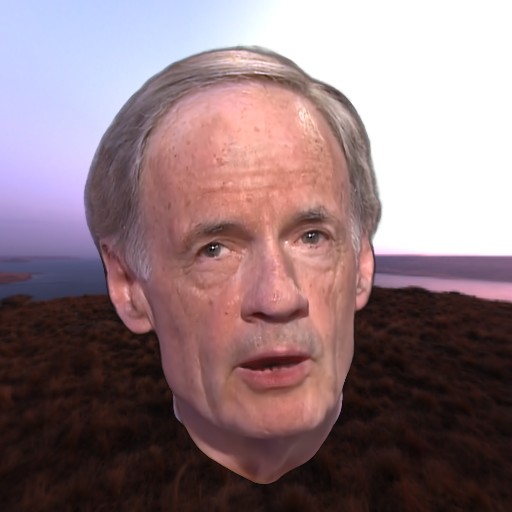} &
        \includegraphics[width=\w,valign=c]{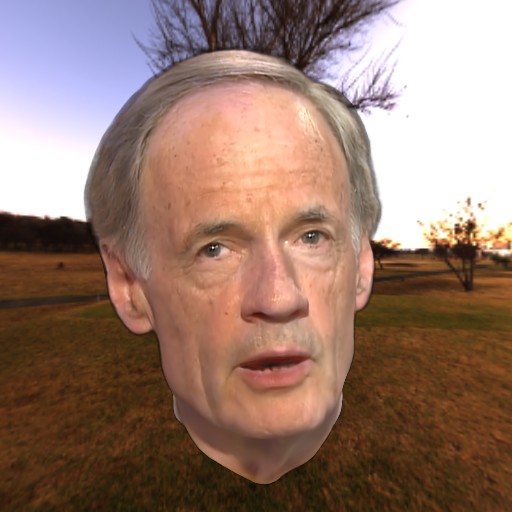} &
        \rotatebox[origin=c]{-90}{Ours}
        \\
        \includegraphics[width=\w,valign=c]{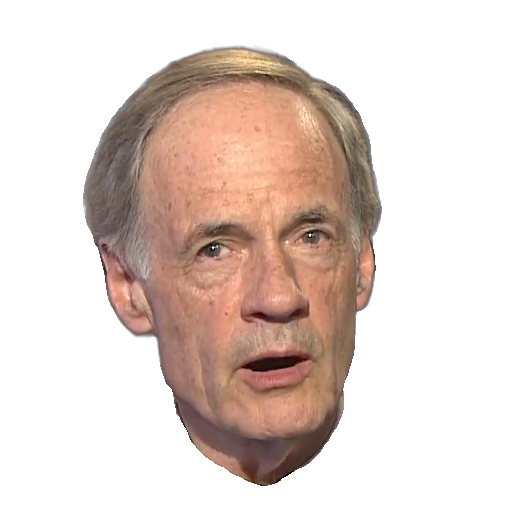} &
        \includegraphics[width=\w,valign=c]{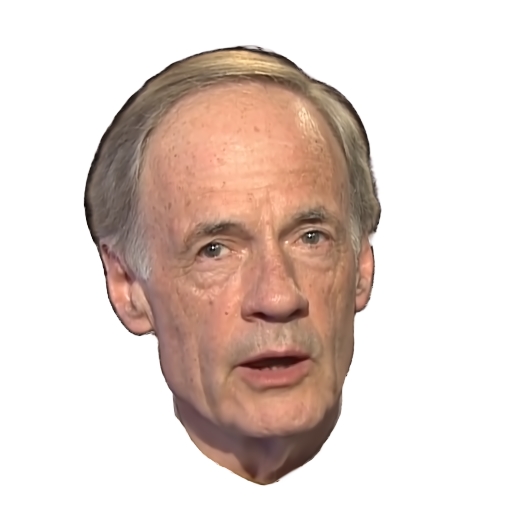} &
        \includegraphics[width=\w,valign=c]{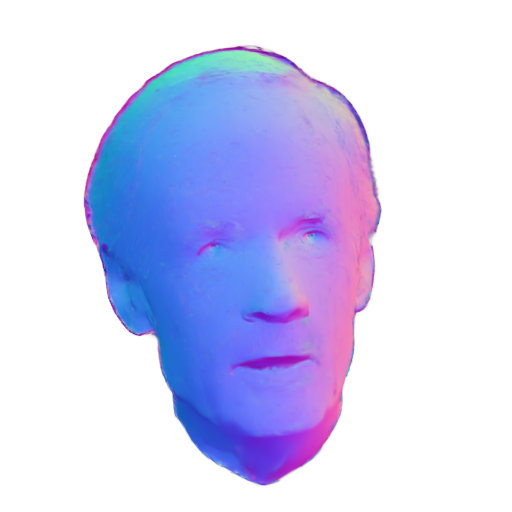} &
        \includegraphics[width=\w,valign=c]{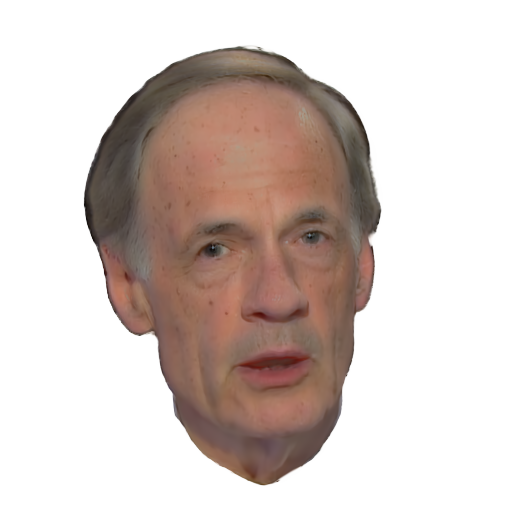} &
        \includegraphics[width=\w,valign=c]{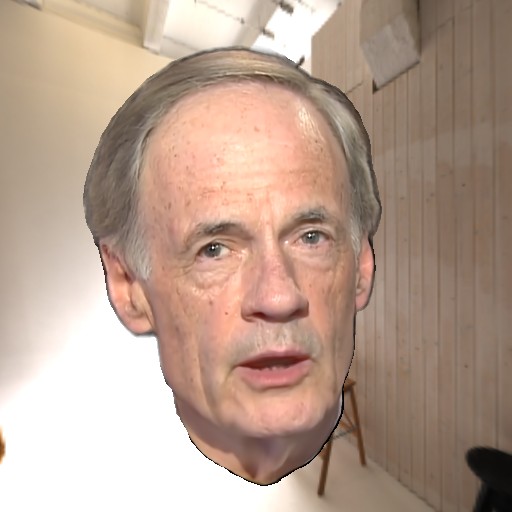} &
        \includegraphics[width=\w,valign=c]{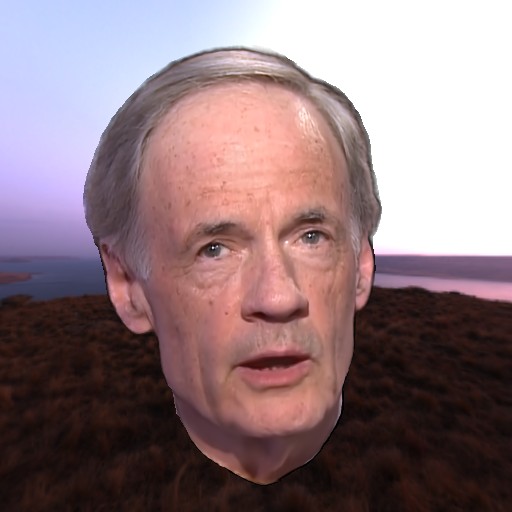} &
        \includegraphics[width=\w,valign=c]{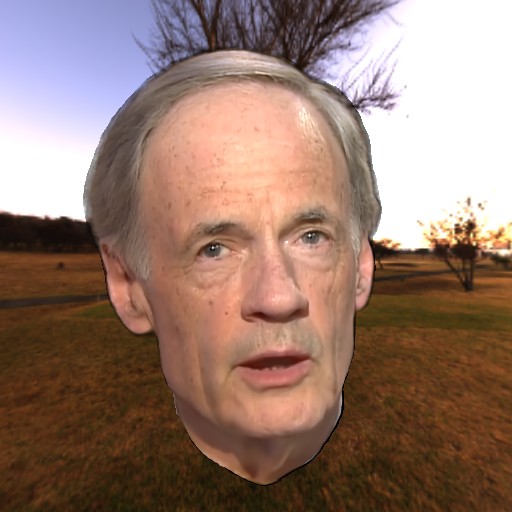} &
        \rotatebox[origin=c]{-90}{HRAvatar}
        \\
        &
        \includegraphics[width=\w,valign=c]{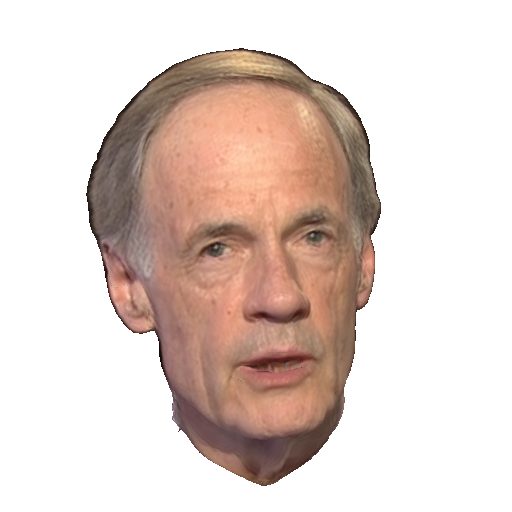} &
        \includegraphics[width=\w,valign=c]{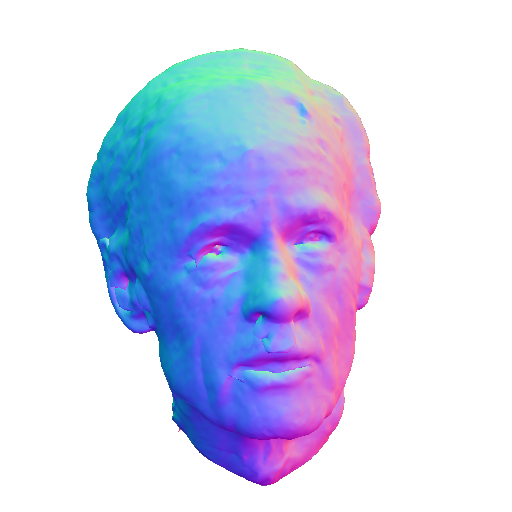} &
        \includegraphics[width=\w,valign=c]{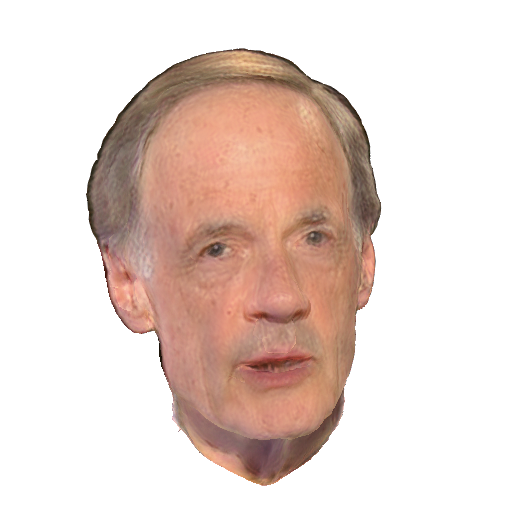} &
        \includegraphics[width=\w,valign=c]{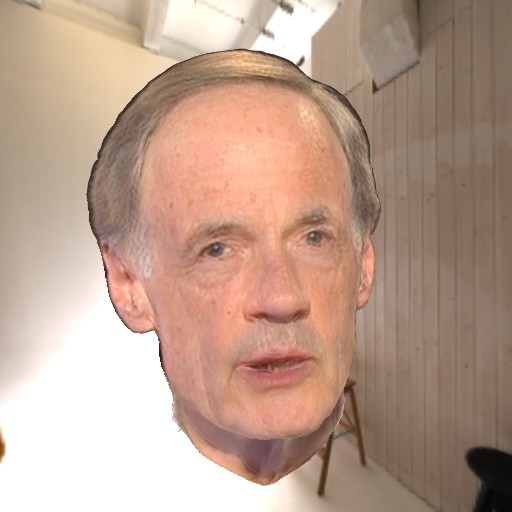} &
        \includegraphics[width=\w,valign=c]{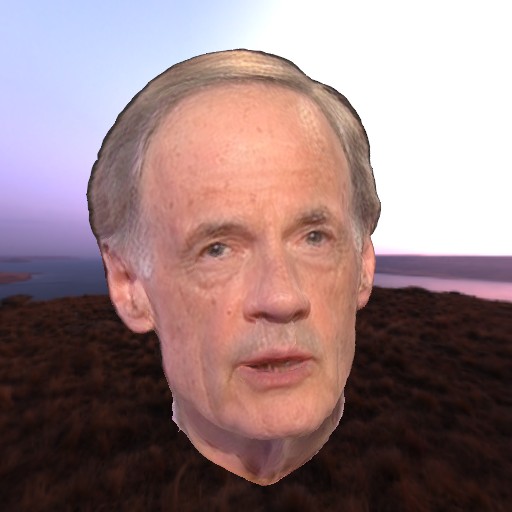} &
        \includegraphics[width=\w,valign=c]{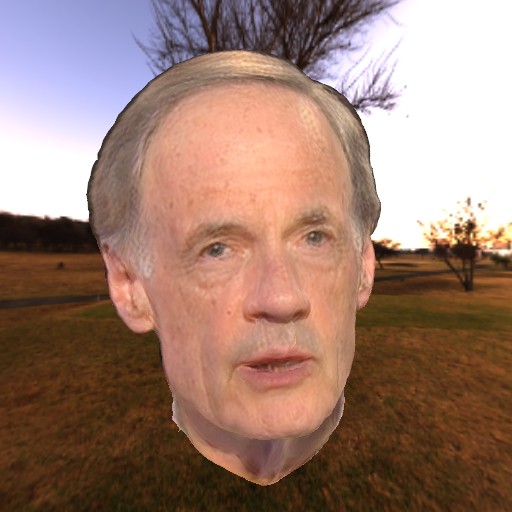} &
        \rotatebox[origin=c]{-90}{FLARE}

        \\
        &
        \includegraphics[width=\w,valign=c]{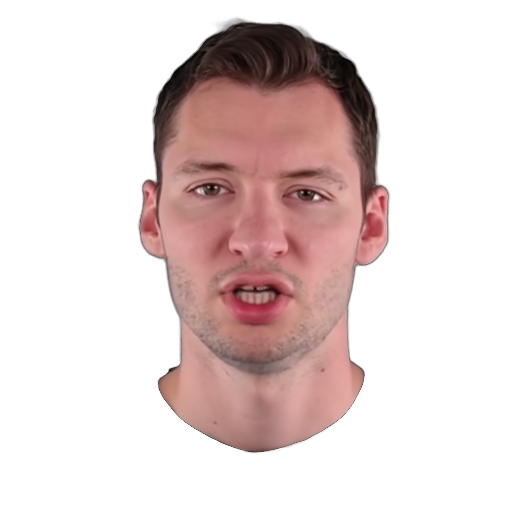} &
        \includegraphics[width=\w,valign=c]{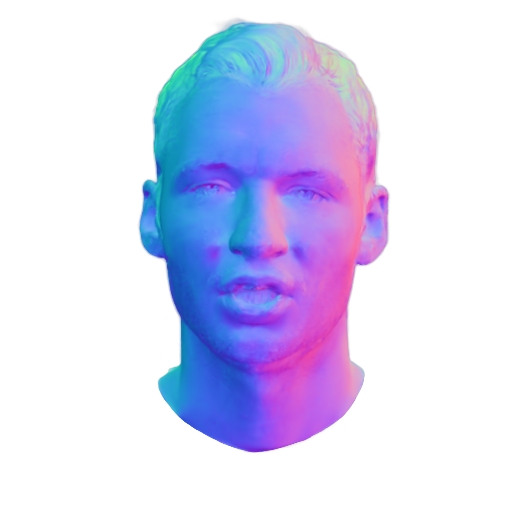} &
        \includegraphics[width=\w,valign=c]{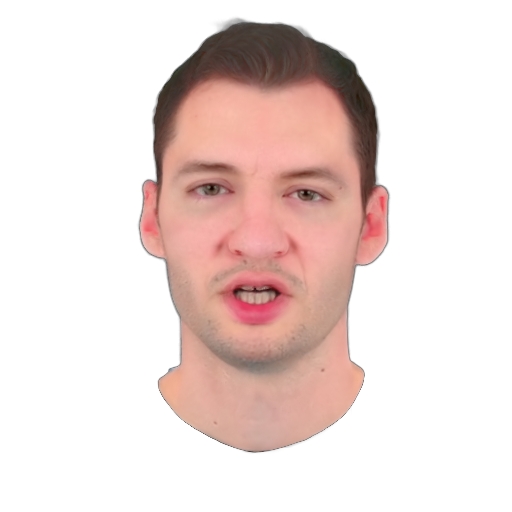} &
        \includegraphics[width=\w,valign=c]{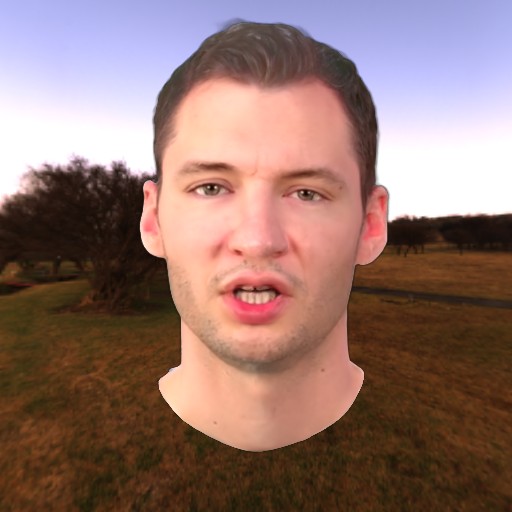} &
        \includegraphics[width=\w,valign=c]{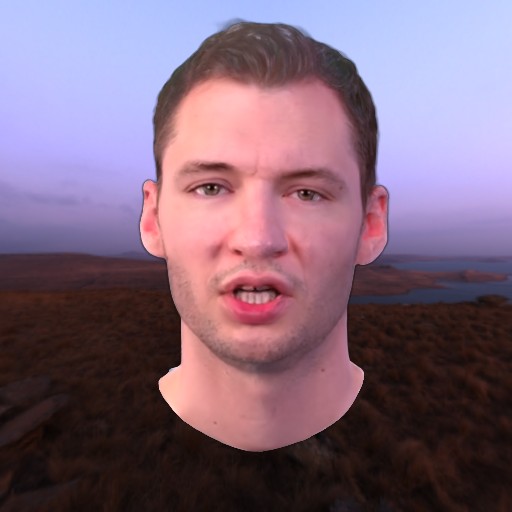} &
        \includegraphics[width=\w,valign=c]{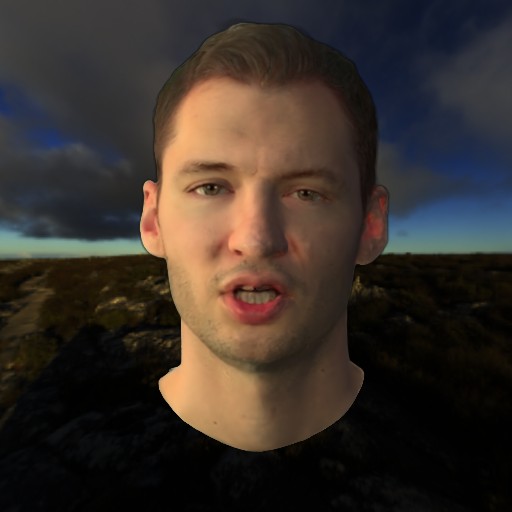} &
        \rotatebox[origin=c]{-90}{Ours}
        \\
        \includegraphics[width=\w,valign=c]{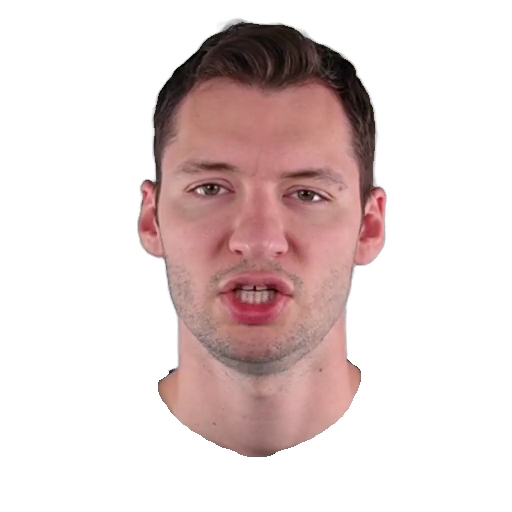} &
        \includegraphics[width=\w,valign=c]{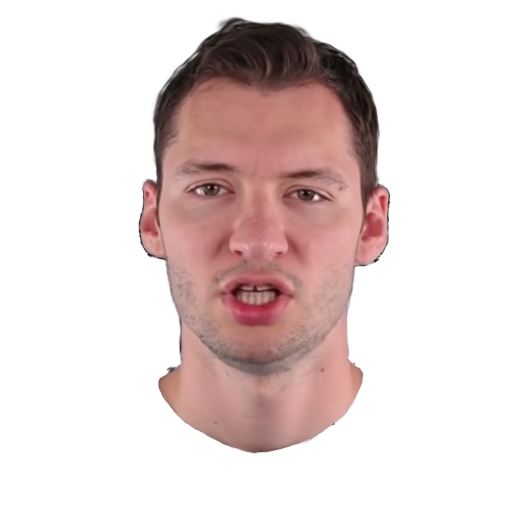} &
        \includegraphics[width=\w,valign=c]{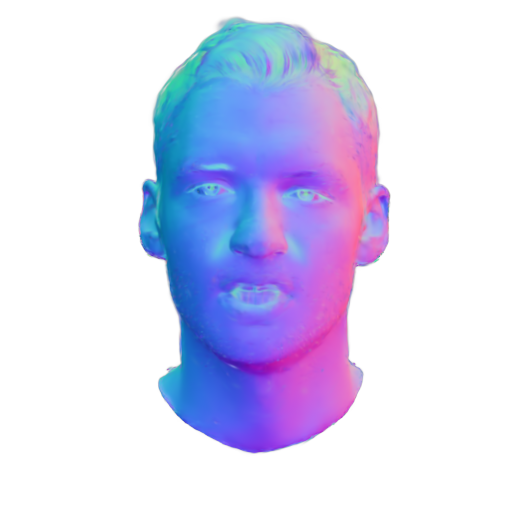} &
        \includegraphics[width=\w,valign=c]{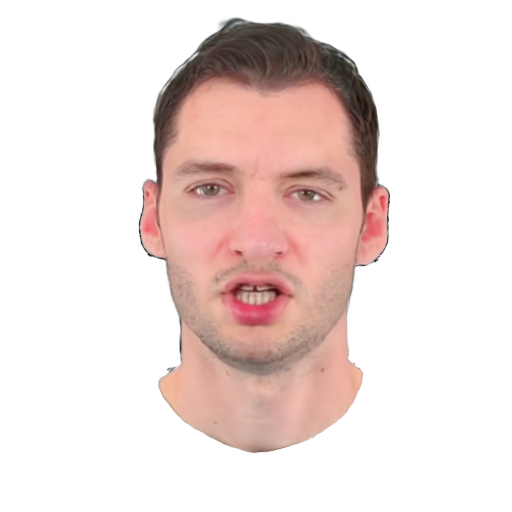} &
        \includegraphics[width=\w,valign=c]{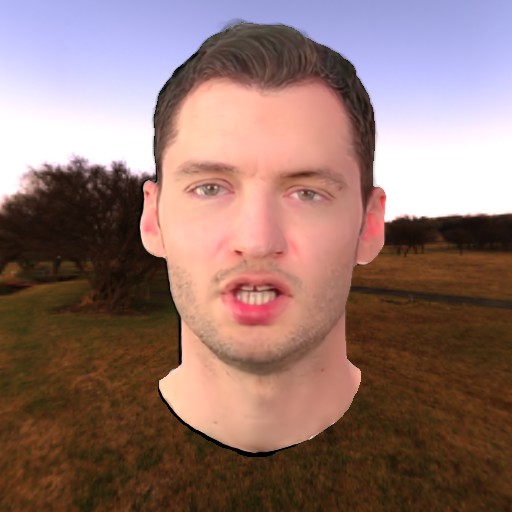} &
        \includegraphics[width=\w,valign=c]{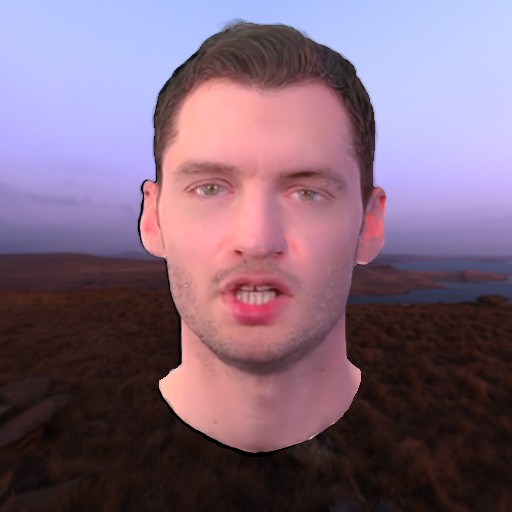} &
        \includegraphics[width=\w,valign=c]{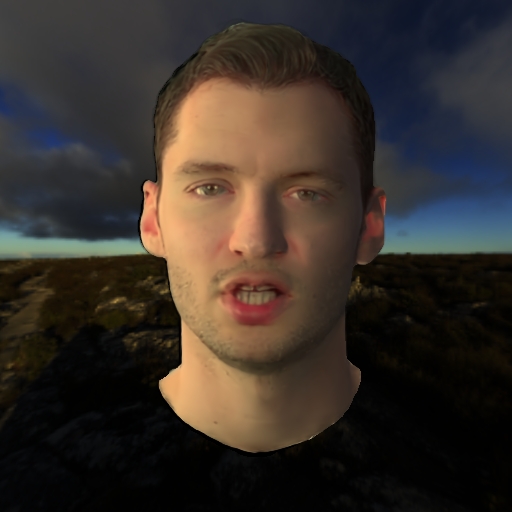} &
        \rotatebox[origin=c]{-90}{HRAvatar}
        \\
        &
        \includegraphics[width=\w,valign=c]{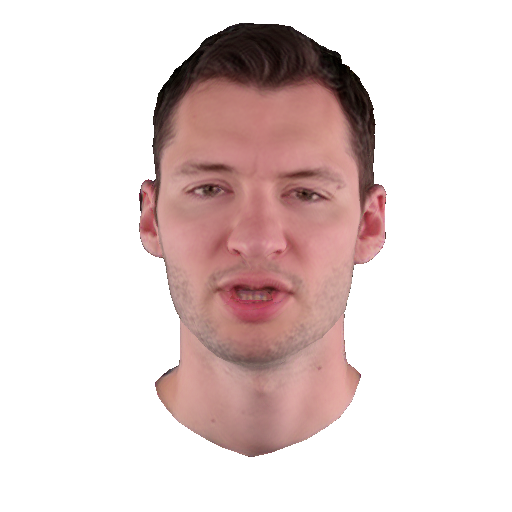} &
        \includegraphics[width=\w,valign=c]{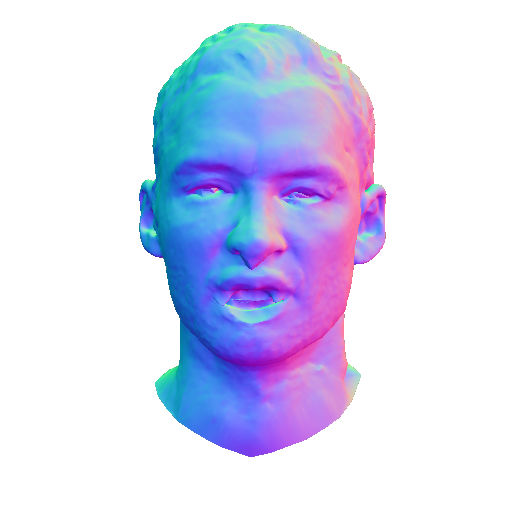} &
        \includegraphics[width=\w,valign=c]{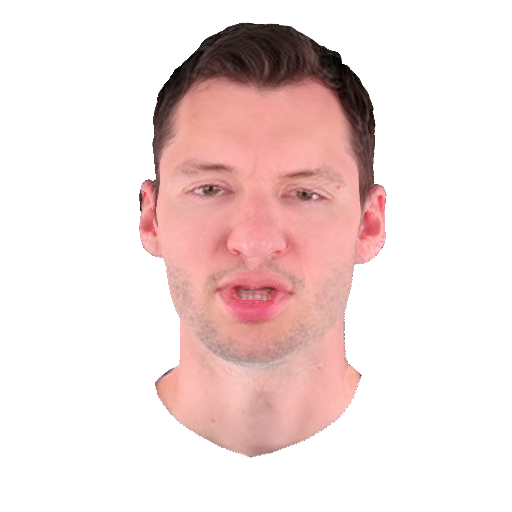} &
        \includegraphics[width=\w,valign=c]{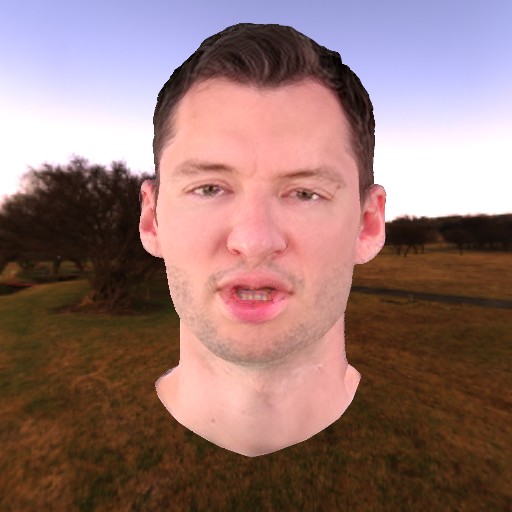} &
        \includegraphics[width=\w,valign=c]{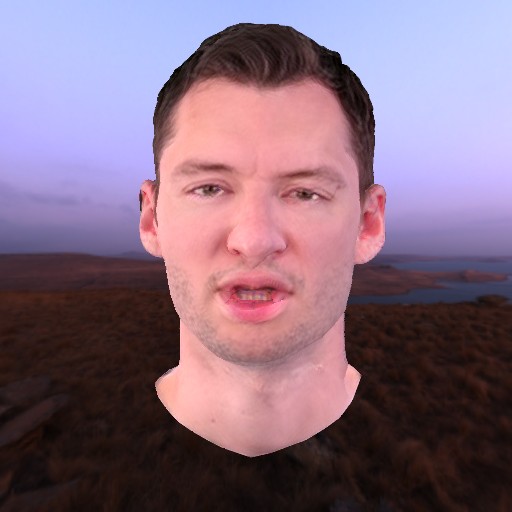} &
        \includegraphics[width=\w,valign=c]{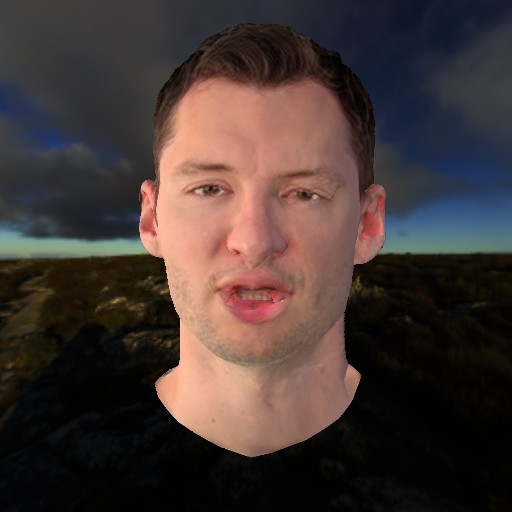} &
        \rotatebox[origin=c]{-90}{FLARE}

        \\
        &
        \includegraphics[width=\w,valign=c]{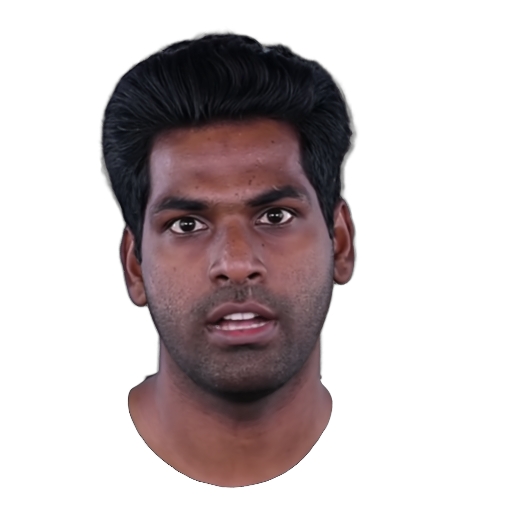} &
        \includegraphics[width=\w,valign=c]{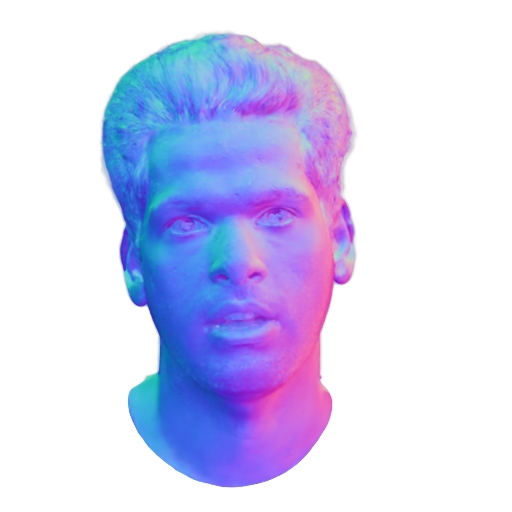} &
        \includegraphics[width=\w,valign=c]{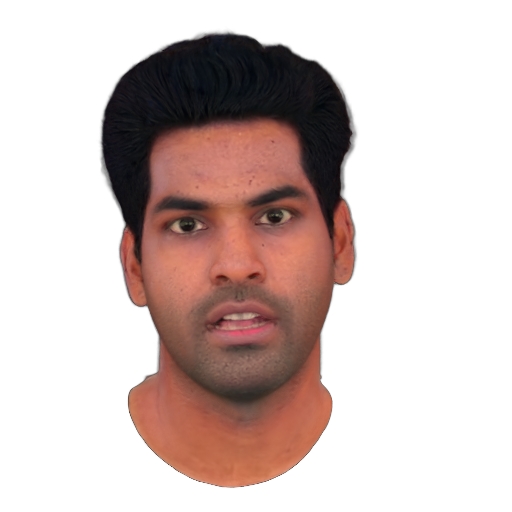} &
        \includegraphics[width=\w,valign=c]{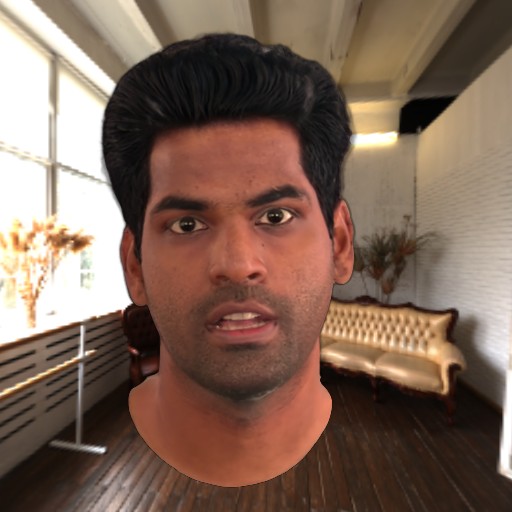} &
        \includegraphics[width=\w,valign=c]{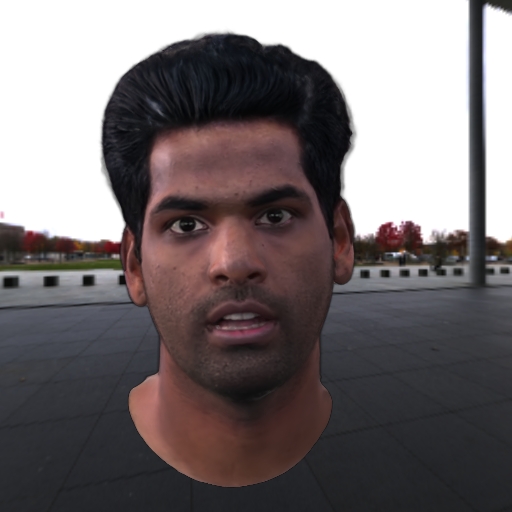} &
        \includegraphics[width=\w,valign=c]{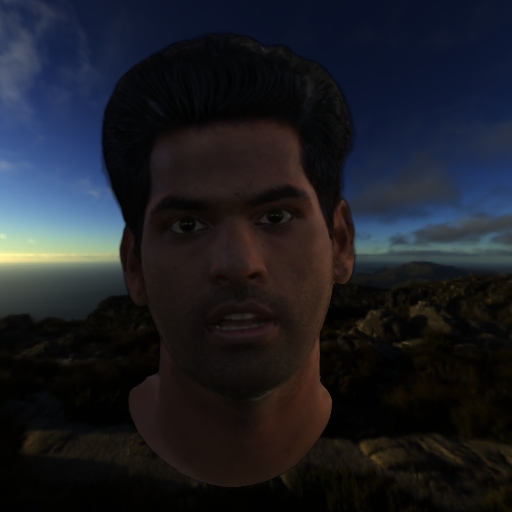} &
        \rotatebox[origin=c]{-90}{Ours}
        \\
        \includegraphics[width=\w,valign=c]{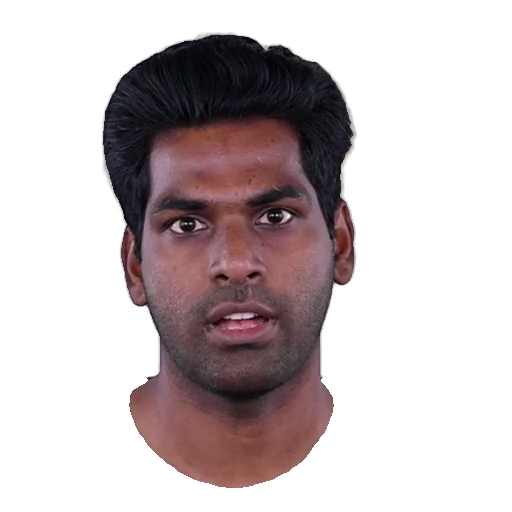} &
        \includegraphics[width=\w,valign=c]{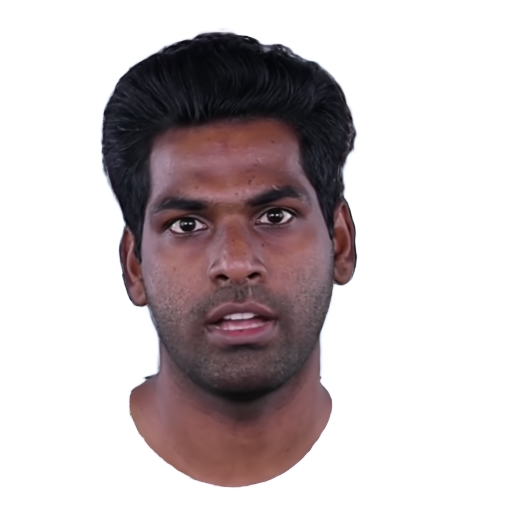} &
        \includegraphics[width=\w,valign=c]{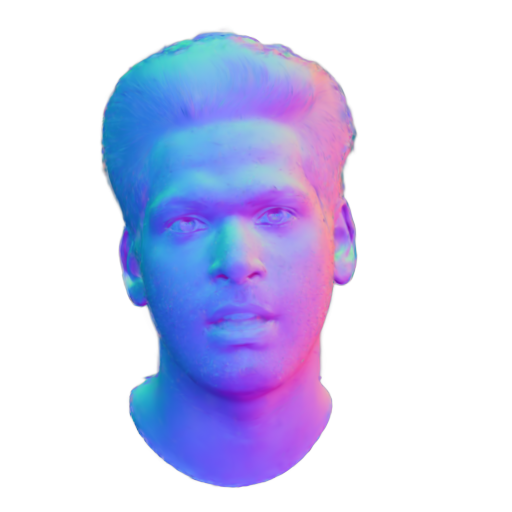} &
        \includegraphics[width=\w,valign=c]{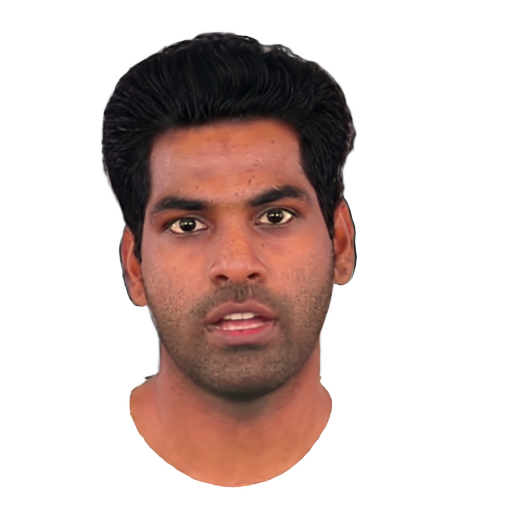} &
        \includegraphics[width=\w,valign=c]{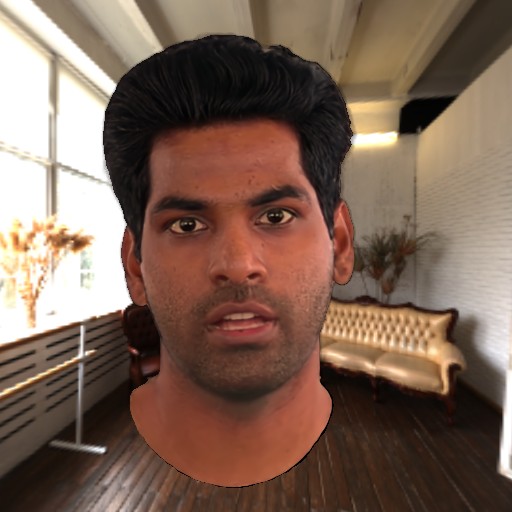} &
        \includegraphics[width=\w,valign=c]{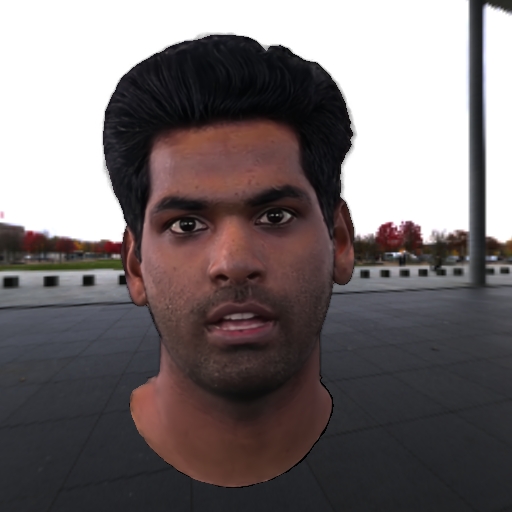} &
        \includegraphics[width=\w,valign=c]{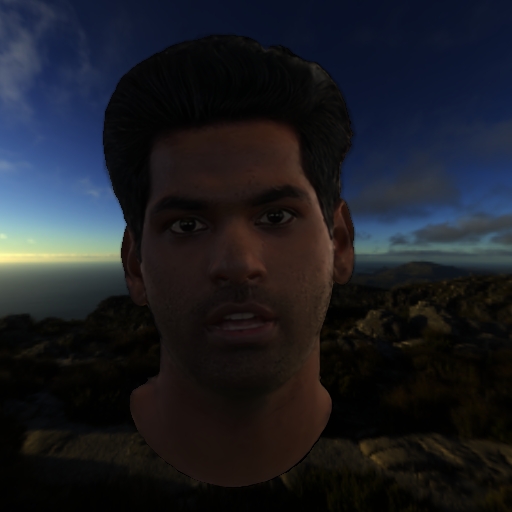} &
        \rotatebox[origin=c]{-90}{HRAvatar}
        \\
        &
        \includegraphics[width=\w,valign=c]{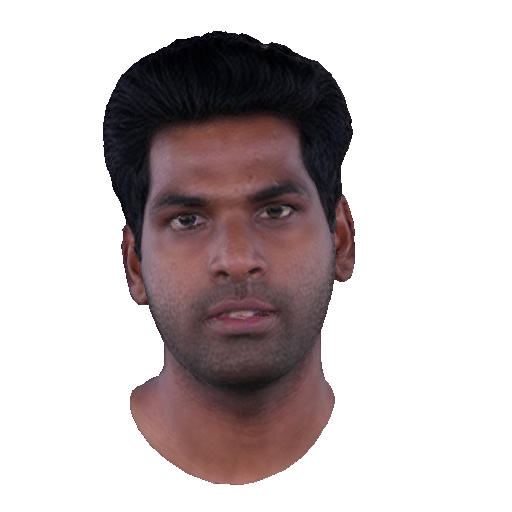} &
        \includegraphics[width=\w,valign=c]{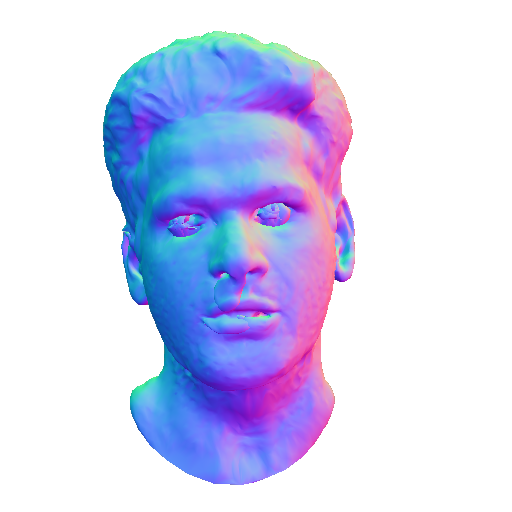} &
        \includegraphics[width=\w,valign=c]{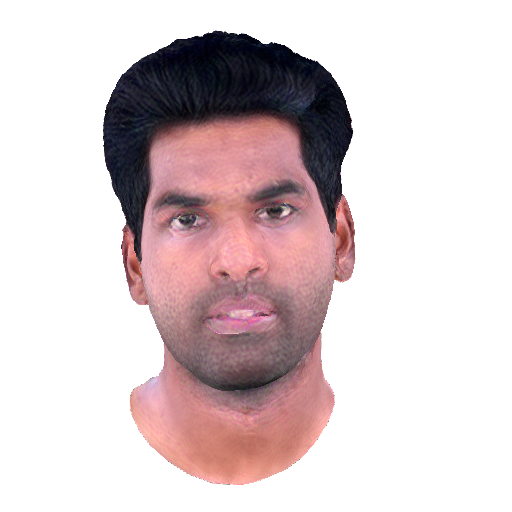} &
        \includegraphics[width=\w,valign=c]{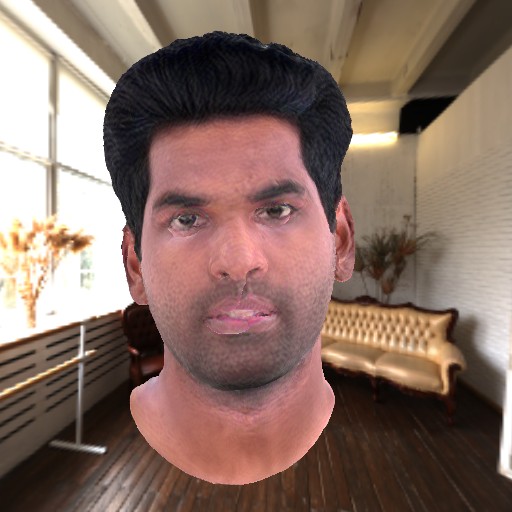} &
        \includegraphics[width=\w,valign=c]{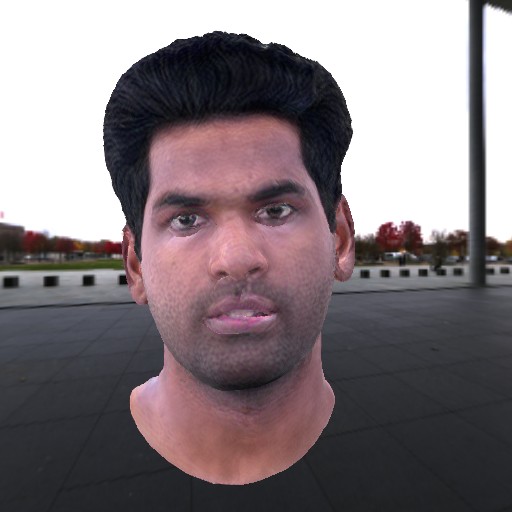} &
        \includegraphics[width=\w,valign=c]{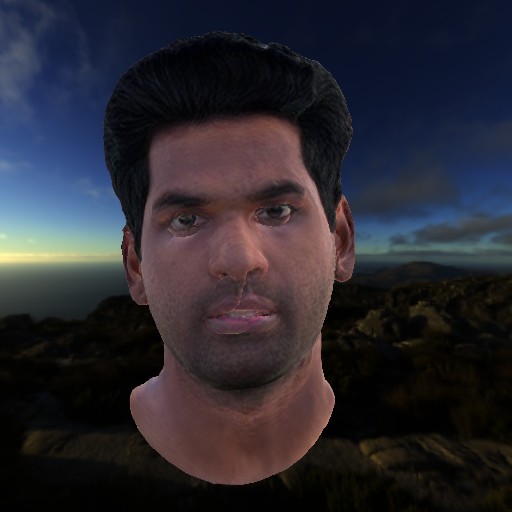} &
        \rotatebox[origin=c]{-90}{FLARE}

        \\
        &
        \includegraphics[width=\w,valign=c]{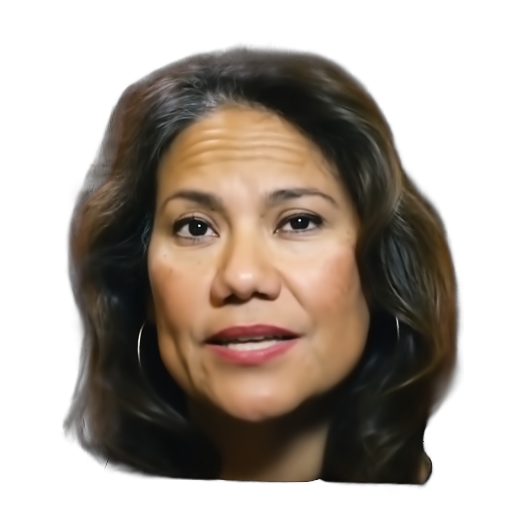} &
        \includegraphics[width=\w,valign=c]{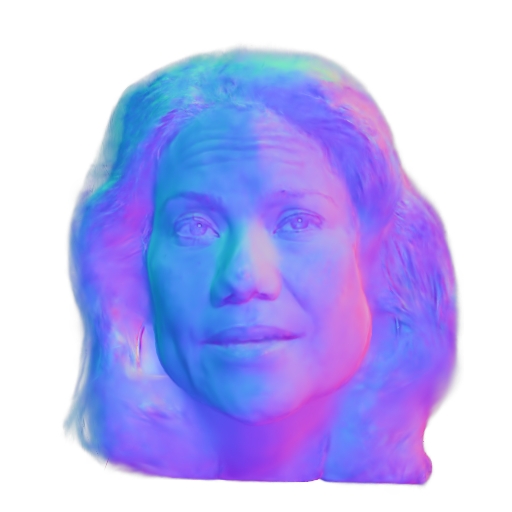} &
        \includegraphics[width=\w,valign=c]{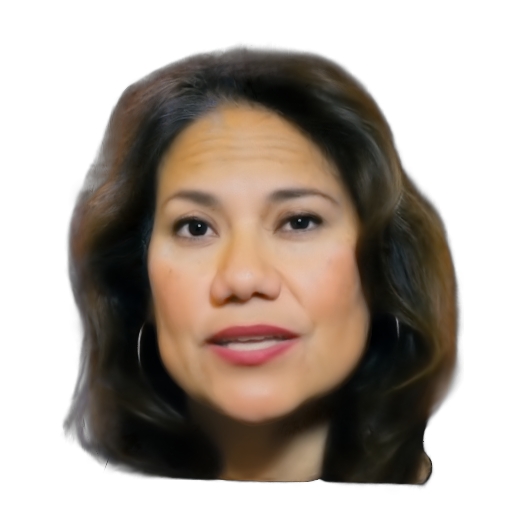} &
        \includegraphics[width=\w,valign=c]{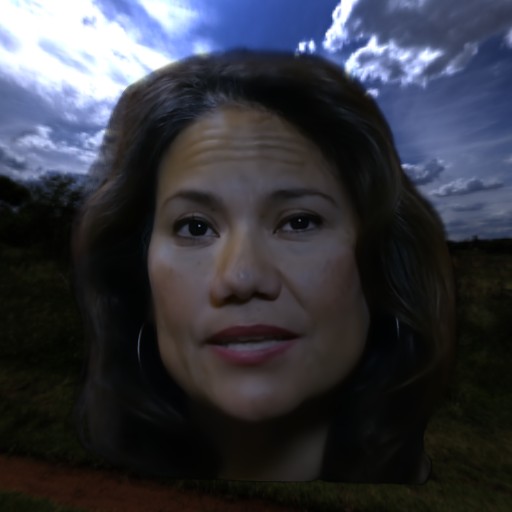} &
        \includegraphics[width=\w,valign=c]{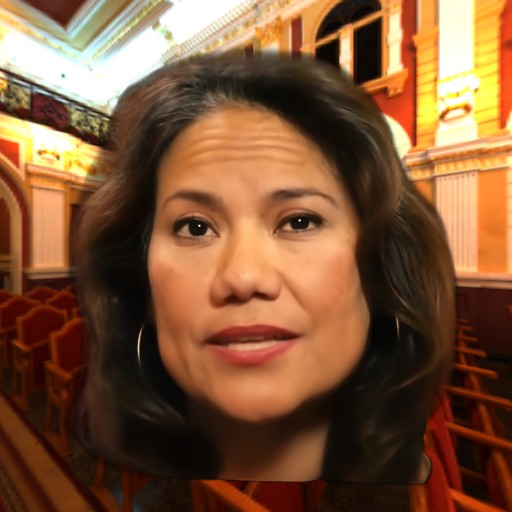} &
        \includegraphics[width=\w,valign=c]{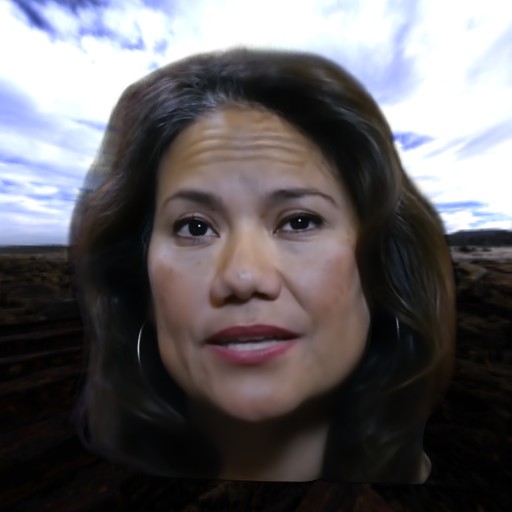} &
        \rotatebox[origin=c]{-90}{Ours}
        \\
        \includegraphics[width=\w,valign=c]{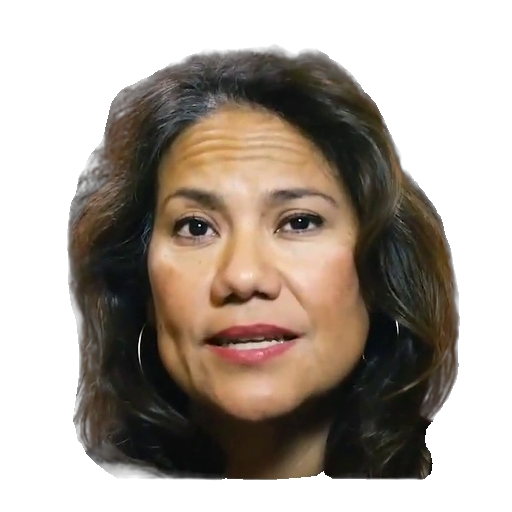} &
        \includegraphics[width=\w,valign=c]{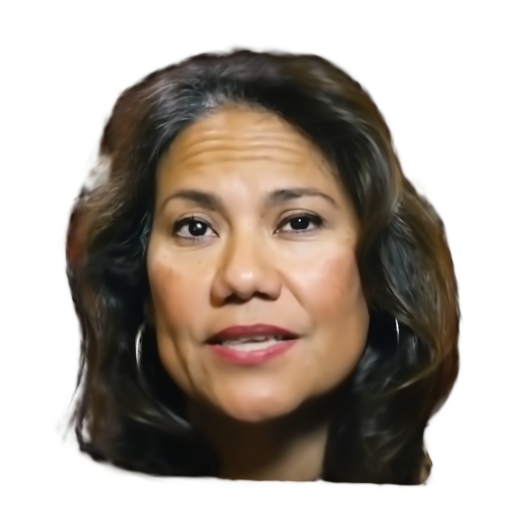} &
        \includegraphics[width=\w,valign=c]{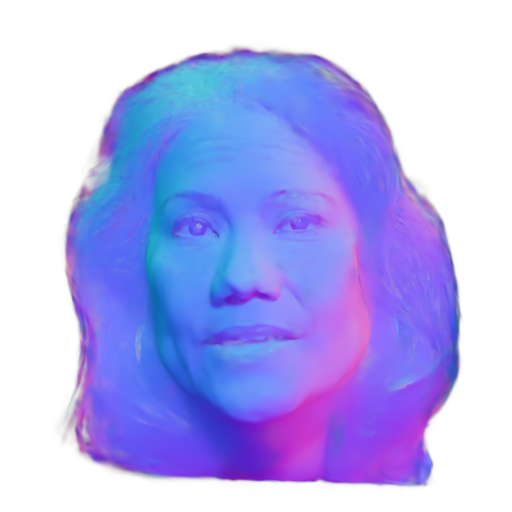} &
        \includegraphics[width=\w,valign=c]{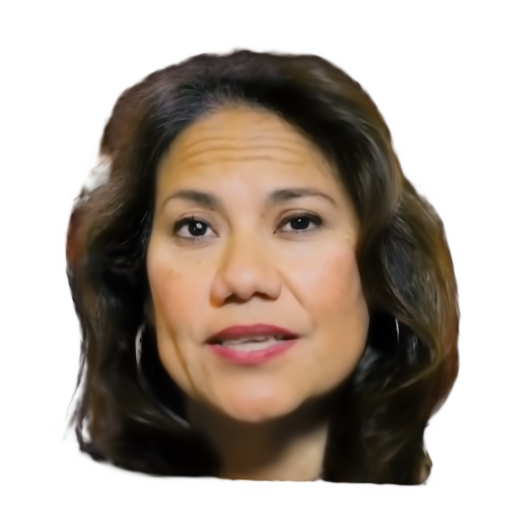} &
        \includegraphics[width=\w,valign=c]{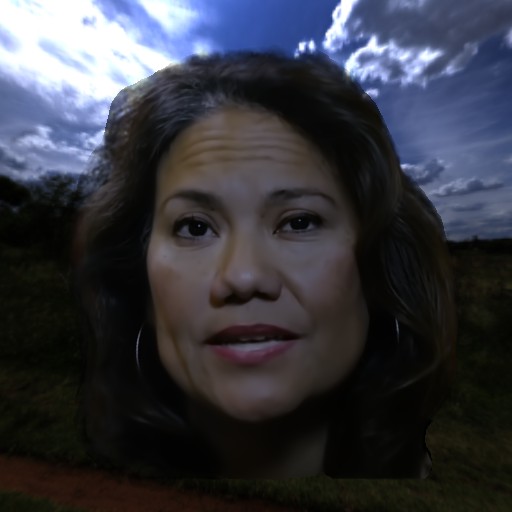} &
        \includegraphics[width=\w,valign=c]{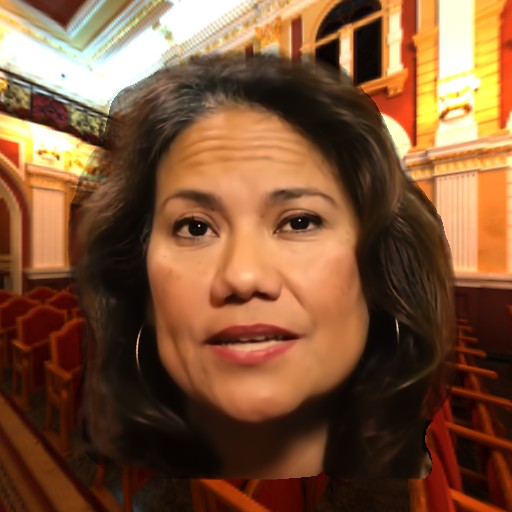} &
        \includegraphics[width=\w,valign=c]{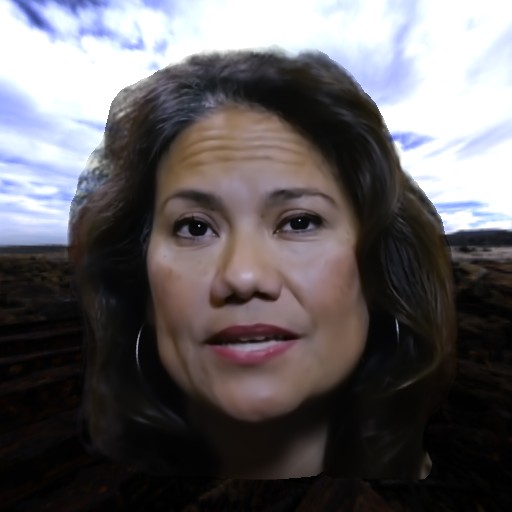} &
        \rotatebox[origin=c]{-90}{HRAvatar}
        \\
        &
        \includegraphics[width=\w,valign=c]{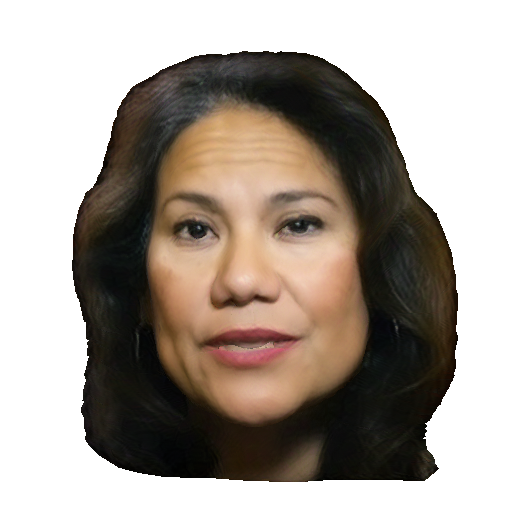} &
        \includegraphics[width=\w,valign=c]{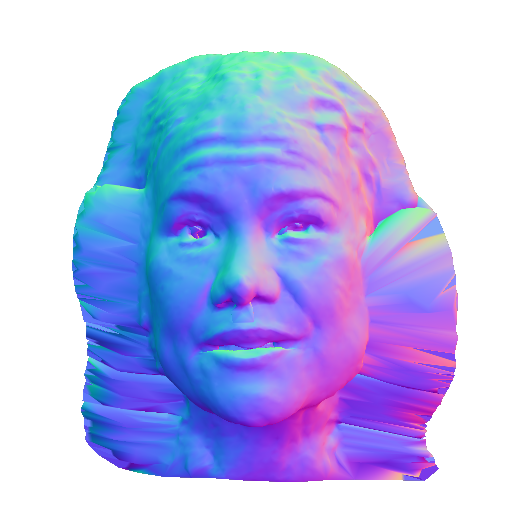} &
        \includegraphics[width=\w,valign=c]{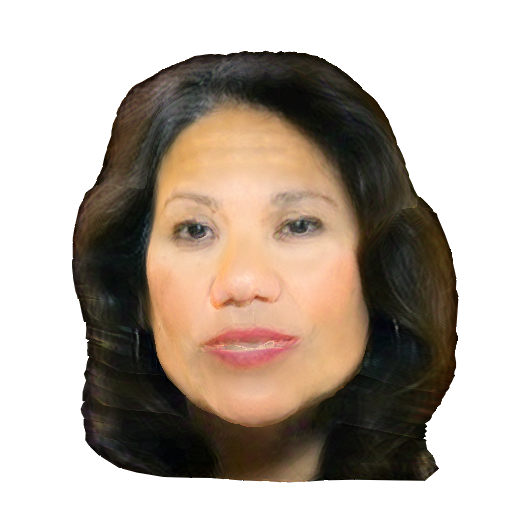} &
        \includegraphics[width=\w,valign=c]{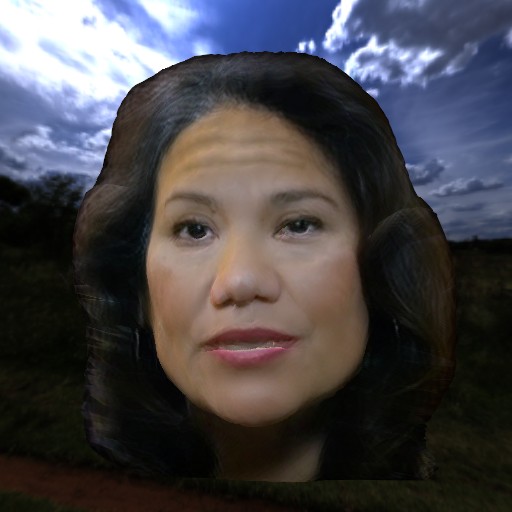} &
        \includegraphics[width=\w,valign=c]{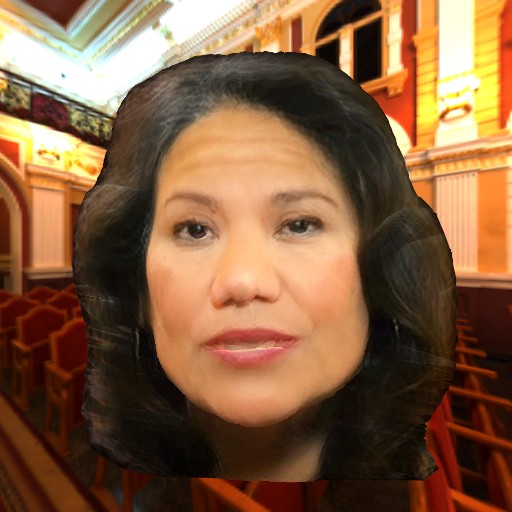} &
        \includegraphics[width=\w,valign=c]{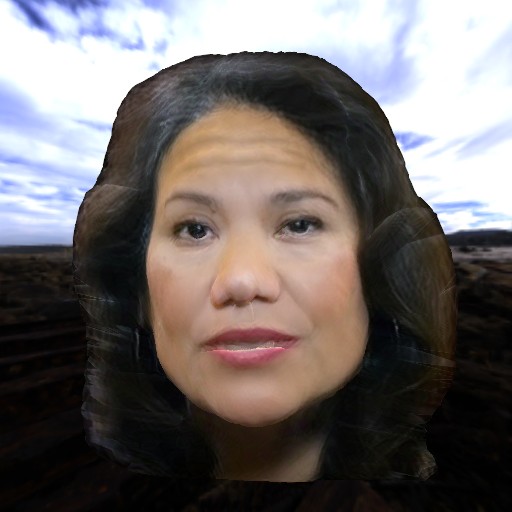} &
        \rotatebox[origin=c]{-90}{FLARE}
    \end{tabular}
    \caption{Visual comparison of our method with HRAvatar~\cite{HRAvatar} and FLARE~\cite{flare} for reconstruction and environment map relighting.}
    \label{fig:exp-comparison-recons-relight}
\end{figure*}

In this section, we validate the quality of our reconstructions in the self-reenactment setting. Unseen frames from the video are tracked to recover FLAME parameters using the tuned encoder, then reconstructed. In Table~\ref{tab:exp-reconstruction} we report average PSNR, SSIM and VGG-based LPIPS metrics for the INSTA and HDTF datasets, compared to state-of-the-art monocular avatar reconstruction methods as reported by the authors of HRAvatar~\cite{HRAvatar}. Our method displays state of the art performance, comparably with HRAvatar, outperforming all baselines. Qualitative reconstruction examples are shown in Figures \ref{fig:result_ours} and \ref{fig:exp-comparison-recons-relight}. Figure~\ref{fig:novel_view} showcases our method's ability to render realistic avatars from novel viewpoints.

Moreover, we assess the effect of primitive count and texture resolution in Figure~\ref{fig:lpips_texres_gaussians_params}. Average LPIPS on the INSTA dataset is reported for various primitive counts and texture resolutions. For fair comparison with HRAvatar which does not have textures, we also compare with equal parameter counts. These results underline the efficiency of our method, enabling on-par reconstruction quality with a significantly smaller model size. The added representation power of textures reduces the number of primitives required through a compact representation of normal and appearance attributes enabled by the UV-space locality of adjacent splats.

In Figure~\ref{fig:exp-ablation-downscale-texture}, we render an avatar with textures down-scaled to lower resolutions at test time without any additional tuning. As shown, our texture-based modeling enables intuitive control of model size, an important capability for bandwidth-adaptive applications. Note that training with lower resolution textures directly would yield better results than down-scaling the textures post-facto, as the optimization process can adapt to the reduced representation power.
 
\subsection{Texture Editing} \label{sec:exp-texediting}

In this section, we show that our texture mapping approach enables several editing use cases that would be very difficult if not impossible to implement with standard Gaussian Splatting rasterization.

Figure~\ref{fig:exp-texture-editing} demonstrates overlay of sharp decals, precise editing of local features and color-shifting face regions. Figure~\ref{fig:exp-texture-editing-pbr} showcases more substantial edits using off-the-shelf PBR materials, rendered under varying illumination to validate the accuracy of our physically-based rendering with conventional material definitions. Our results show that the edited avatars maintain sharpness and consistency across variations in pose, expression and illumination.

In Figure~\ref{fig:exp-ablation-texturing}, we perform an ablation study of several aspects of our method aimed at improving the texture mapping and the quality of reconstructed textures. First, we validate that the UV distortion loss enhances the sharpness of our render. While the reconstruction converges well without the regularization, mapping a texture with high-frequency details reveals some blurriness. Next, we show that the statistical albedo regularization is able to fill-in the holes in the textures, remove distracting artifacts and better preserve the structure of the texture space. Finally, we set our Jacobians (see Section~\ref{sec:method-uvmapping}) to zero, which equates to using a single UV coordinate per splat. This per-primitive mapping yields a discontinuous texture with only a sparse set of texels being used (note that we disable the statistical albedo regularization to better visualize those texels), and cannot be edited at high resolution unless the number of primitives is impractically high.

FATE~\cite{fate} allows texture-based editing of Gaussian Avatars. However, their method relies on per-primitive color, which severely limits the quality of rendering for high-frequency textures. Figure~\ref{fig:fate_comparison} compares renders with the same texture for FATE and GTAvatar. Our method outperforms FATE even while increasing the number of primitives FATE uses well beyond its default.

\subsection{Normal mapping} \label{sec:exp-normal-mapping}

In Figure~\ref{fig:exp-ablation-normal-map}, we demonstrate the impact of our normal mapping on the sharpness of rasterized normals and the quality of relit renders. Furthermore, Figure~\ref{fig:exp-texture-editing-pbr} shows how handcrafted normal maps substantially improve the realism for material editing under dynamic lighting.

\subsection{Relighting} \label{sec:exp-relighting}

Figure~\ref{fig:exp-comparison-recons-relight} presents relighting examples compared with HRAvatar and FLARE. Our method achieves comparable or superior quality relative to both baselines, which employ the same physically-based rendering formulation.
We provide further examples in Figure~\ref{fig:result_ours} to illustrate the relighting capabilities of our method.

\subsection{Rendering time} \label{sec:exp-rendering-time}

Sampling a texture at every ray-splat intersection comes with a computational cost. To minimize this cost, our implementation uses hardware acceleration at test time for more efficient access via CUDA texture objects. In Table~\ref{tab:exp-rendering-time}, we compare rendering speed with other methods, and with variants of our method without hardware acceleration, with more Gaussians and with naive projection instead of the fast UV mapping described in Section~\ref{sec:method-uvmapping}. In the latter, we explicitly perform orthogonal projection to the triangle plane and calculate barycentric coordinates at ray-splat intersection. We report FPS for both static and dynamic geometry, as we found FLAME deformations and, in our case, updating the Jacobians, to be a significant contributor to rendering time. Note that the static geometry case still enables camera movement and relighting. While our method performs slower than the HRAvatar baseline for a given number of primitives, ours requires less (as shown in Figure~\ref{fig:lpips_texres_gaussians}), yielding on par rendering time for comparable or better reconstruction quality. The final trained avatar can be rendered at more than \textbf{170 FPS} with static geometry, or \textbf{80 FPS} with dynamic geometry on a RTX A5000.    

\begin{table}[h]
    \centering
    \newcommand{\tworows}[1]{\multirow{2}{*}{#1}}
    \begin{tabular}{cccc}
        \toprule
        \multirow{2}{*}[1pt]{Method} & FPS & FPS & \multirow{2}{*}[1pt]{LPIPS $\downarrow$} \\[-2pt]
        & (static) & (dynamic) & \\
        \midrule
        
        FLARE~\cite{flare} & 31 & 27 & 0.082 \\
        
        HRAvatar 80k \cite{HRAvatar} & 136 & \ranktwo{88} & 0.062 \\
        
        HRAvatar 10k \cite{HRAvatar} & \ranktwo{165} & \rankone{105} & 0.071 \\
                       
        Ours 10k & \tworows{121} & \tworows{64} & \tworows{} \\[-3pt]
        \small{(no hw. acceleration)} & & & \ranktwo{0.060} \\
        
        Ours 10k & \tworows{152} & \tworows{67} & \tworows{} \\[-3pt]
        \small{(naive projection)} & & & \ranktwo{0.060} \\

        Ours 80k & \tworows{109} & \tworows{64} & \tworows{} \\[-3pt] 
        \small{(more Gaussians)} & & & \rankone{0.058} \\

        \midrule
        Ours 10k & \rankone{175} & \ranktwo{83} & \ranktwo{0.060} \\
        \bottomrule
    \end{tabular}
    \caption{Comparison of rendering speed at inference for static and dynamic geometry on a NVIDIA RTX A5000 GPU, and average LPIPS for unseen frames of the INSTA dataset. Numbers next to method names indicate the number of Gaussians used, when applicable (\textbf{densification and pruning are disabled} for manual control). Despite the additional overhead of texture sampling, our method achieves competitive speed by reducing the number of Gaussians required to achieve a photorealistic reconstruction.}
    \label{tab:exp-rendering-time}
\end{table}

\begin{table}[!hbtp]
    \centering
    \scalebox{0.9}{
    \subfloat[Learning rates.]{
    \begin{tabular}{lr}
        \toprule
        Parameter & LR \\
        \midrule
        Barycentric coords. & \num{1e-3} \\
        Rotation & \num{1e-3} \\
        Scale & \num{5e-3} \\
        Displacement & \num{2e-5} \\
        Opacity & \num{5e-2} \\
        Expression encoder & \num{5e-5} \\
        Material texture & \num{5e-3} \\
        Normal texture & \num{1e-3} \\
        Environment map & \num{2e-2} \\
        FLAME template vertices & \num{1e-5} \\
        FLAME LBS weights & \num{1e-4} \\
        FLAME expr. and pose shapes & \num{1e-6} \\
        FLAME statistical albedo & \num{5e-2} \\
        \\
        \bottomrule
    \end{tabular}
    \label{tab:learning-rates}
    }
    \subfloat[Objective function weights.]{
    \begin{tabular}{lr}
        \toprule
        Weight & Value \\
        \midrule
        $\lambda_{\text{L1}}$ & $0.80$ \\
        $\lambda_{\text{SSIM}}$ & $0.20$ \\
        $\lambda_{\text{mask}}$ & $0.10$ \\
        $\lambda_{\text{diff\_albedo}}$ & $0.25$ \\
        $\lambda_{\text{stat\_albedo}}$ & $0.0001$ \\
        $\lambda_{\text{expr}}$ & $0.01$ \\
        $\lambda_{\text{smooth}}$ & $0.01$ \\
        $\lambda_{\text{normal\_reg}}$ & $0.01$ \\
        $\lambda_{\text{normal\_consist}}$ & $0.05$ \\
        $\lambda_{\text{uv\_dist}}$ & $50$ \\
        $\lambda_{\text{boundary}}$ & $1$ \\
        $\lambda_{\text{bary}}$ & $0.1$ \\
        $\lambda_{\text{lap}}$ & $200$ \\
        $\lambda_{\text{FLAME}}$ & $0.001$ \\
        \bottomrule
    \end{tabular}
    \label{tab:loss-weights}
    }
    }
    \caption{Hyperparameters used for training our method.}
    \label{tab:hyperparameters}
\end{table}

\section{Ethics}

This research aims at pushing forward the precision and authenticity of 3D facial reconstruction for legitimate applications such as visual effects or virtual interactions. We do not condone the use of our work for producing unconsented deepfakes or deceptive content of any kind.  Our focus remains on contributing to scientific progress and industry innovation in alignment with ethical standards. We also encourage ongoing dialogue and the development of regulations to safeguard individual rights as this technology evolves.

\section{Conclusion}

We presented a novel UV-domain Gaussian splatting framework that combines the fidelity and efficiency of EWA volume resampling enabled physics-based inverse rendering with the intuitiveness of texture-based editing, enabling photorealistic, easily editable, and relightable head avatars from monocular videos. Our method achieves state-of-the-art reconstruction and relighting quality while introducing efficient, semantically meaningful solutions for material and geometry control.

Several limitations remain. The use of image-based lighting to represent scene illumination enables intuitive manipulation of the environment, but comes with shortcomings. Most importantly, this approach can only model light at infinity and is unable to represent point lights. Furthermore, effects such as indirect illumination and shadows are not represented, and will thus be baked into other properties, yielding inaccurate physical decomposition.

As part of \textbf{future work}, several avenues remain to further improve our representation. The texture-mapping contribution could be integrated in a Gaussian rasterizer that enables more advanced lighting representations, \eg through ray-tracing \cite{3dgrt2024, xie2024envgs, IRGS, SVG-IR}. 
Reducing aliasing artifacts is also a key challenge: incorporating anti-aliasing strategies, either through trilinear texture filtering in the UV domain or directly within the 2DGS splatting stage \cite{AA2DGS} could enhance visual smoothness and stability of avatars and edits. In addition, extending this representation to the better-constrained multi-view setting could enable further uses. Such extensions would make our method even more suitable for production-grade applications.

\section{Acknowledgements}

This project was funded in part by the Brittany region of France under the Inno R\&D funding scheme (grant number 24005536).

\printbibliography

@article{xie2024envgs,
    title={EnvGS: Modeling View-Dependent Appearance with Environment Gaussian},
    author={Xie, Tao and Chen, Xi and Xu, Zhen and Xie, Yiman and Jin, Yudong and Shen, Yujun and Peng, Sida and Bao, Hujun and Zhou, Xiaowei},
    journal={arXiv preprint arXiv:2412.15215},
    year={2024}
}

@article{3dgrt2024,
    author = {Nicolas Moenne-Loccoz and Ashkan Mirzaei and Or Perel and Riccardo de Lutio and Janick Martinez Esturo and Gavriel State and Sanja Fidler and Nicholas Sharp and Zan Gojcic},
    title = {3D Gaussian Ray Tracing: Fast Tracing of Particle Scenes},
    journal = {ACM Transactions on Graphics and SIGGRAPH Asia},
    year = {2024},
}

@article{facetunegan,
    author = {Olivier, Nicolas and Baert, Kelian and Danieau, Fabien and Multon, Franck and Avril, Quentin},
    title = {FaceTuneGAN: Face autoencoder for convolutional expression transfer using neural generative adversarial networks},
    year = {2023},
    issue_date = {Feb 2023},
    publisher = {Pergamon Press, Inc.},
    address = {USA},
    volume = {110},
    number = {C},
    issn = {0097-8493},
    url = {https://doi.org/10.1016/j.cag.2022.12.004},
    doi = {10.1016/j.cag.2022.12.004},
    journal = {Comput. Graph.},
    month = feb,
    pages = {69–85},
    numpages = {17}
}

@misc{robustvideomatting,
      title={Robust High-Resolution Video Matting with Temporal Guidance}, 
      author={Shanchuan Lin and Linjie Yang and Imran Saleemi and Soumyadip Sengupta},
      year={2021},
      eprint={2108.11515},
      archivePrefix={arXiv},
      primaryClass={cs.CV}
}

@article{chen2024intrinsicanything,
    title     = {IntrinsicAnything: Learning Diffusion Priors for Inverse Rendering Under Unknown Illumination},
    author    = {Xi, Chen and Sida, Peng and Dongchen, Yang and Yuan, Liu and Bowen, Pan and Chengfei, Lv and Xiaowei, Zhou.},
    journal   = {arxiv: 2404.11593},
    year      = {2024},
}

@inproceedings{mega,
  title={Mega: Hybrid mesh-gaussian head avatar for high-fidelity rendering and head editing},
  author={Wang, Cong and Kang, Di and Sun, Heyi and Qian, Shenhan and Wang, Zixuan and Bao, Linchao and Zhang, Song-Hai},
  booktitle={Proceedings of the Computer Vision and Pattern Recognition Conference},
  pages={26274--26284},
  year={2025}
}

@inproceedings{zhang2021flow,
  title={Flow-Guided One-Shot Talking Face Generation With a High-Resolution Audio-Visual Dataset},
  author={Zhang, Zhimeng and Li, Lincheng and Ding, Yu and Fan, Changjie},
  booktitle={Proceedings of the IEEE/CVF Conference on Computer Vision and Pattern Recognition},
  pages={3661--3670},
  year={2021}
}

@inproceedings{gstex,
  title={Gstex: Per-primitive texturing of 2d gaussian splatting for decoupled appearance and geometry modeling},
  author={Rong, Victor and Chen, Jingxiang and Bahmani, Sherwin and Kutulakos, Kiriakos N and Lindell, David B},
  booktitle={2025 IEEE/CVF Winter Conference on Applications of Computer Vision (WACV)},
  pages={3508--3518},
  year={2025},
  organization={IEEE}
}

@misc{texturedgaussiansenhanced3d,
      title={Textured Gaussians for Enhanced 3D Scene Appearance Modeling}, 
      author={Brian Chao and Hung-Yu Tseng and Lorenzo Porzi and Chen Gao and Tuotuo Li and Qinbo Li and Ayush Saraf and Jia-Bin Huang and Johannes Kopf and Gordon Wetzstein and Changil Kim},
      year={2025},
      eprint={2411.18625},
      archivePrefix={arXiv},
      primaryClass={cs.CV},
      url={https://arxiv.org/abs/2411.18625}, 
}

@inproceedings{chen2024gaussianeditor,
  title={Gaussianeditor: Swift and controllable 3d editing with gaussian splatting},
  author={Chen, Yiwen and Chen, Zilong and Zhang, Chi and Wang, Feng and Yang, Xiaofeng and Wang, Yikai and Cai, Zhongang and Yang, Lei and Liu, Huaping and Lin, Guosheng},
  booktitle={Proceedings of the IEEE/CVF conference on computer vision and pattern recognition},
  pages={21476--21485},
  year={2024}
}

@inproceedings{yu2024cogs,
  title={Cogs: Controllable gaussian splatting},
  author={Yu, Heng and Julin, Joel and Milacski, Zolt{\'a}n {\'A} and Niinuma, Koichiro and Jeni, L{\'a}szl{\'o} A},
  booktitle={Proceedings of the IEEE/CVF Conference on Computer Vision and Pattern Recognition},
  pages={21624--21633},
  year={2024}
}

@inproceedings{srinivasan2024nuvo,
  title={Nuvo: Neural uv mapping for unruly 3d representations},
  author={Srinivasan, Pratul P and Garbin, Stephan J and Verbin, Dor and Barron, Jonathan T and Mildenhall, Ben},
  booktitle={European Conference on Computer Vision},
  pages={18--34},
  year={2024},
  organization={Springer}
}

@article{zhan2023general,
  title={General neural gauge fields},
  author={Zhan, Fangneng and Liu, Lingjie and Kortylewski, Adam and Theobalt, Christian},
  journal={arXiv preprint arXiv:2305.03462},
  year={2023}
}

@inproceedings{xiang2021neutex,
  title={Neutex: Neural texture mapping for volumetric neural rendering},
  author={Xiang, Fanbo and Xu, Zexiang and Hasan, Milos and Hold-Geoffroy, Yannick and Sunkavalli, Kalyan and Su, Hao},
  booktitle={Proceedings of the IEEE/CVF Conference on Computer Vision and Pattern Recognition},
  pages={7119--7128},
  year={2021}
}

@inproceedings{fate,
  title={Fate: Full-head gaussian avatar with textural editing from monocular video},
  author={Zhang, Jiawei and Wu, Zijian and Liang, Zhiyang and Gong, Yicheng and Hu, Dongfang and Yao, Yao and Cao, Xun and Zhu, Hao},
  booktitle={Proceedings of the Computer Vision and Pattern Recognition Conference},
  pages={5535--5545},
  year={2025}
}

@inproceedings{flashavatar,
    author    = {Jun Xiang and Xuan Gao and Yudong Guo and Juyong Zhang},
    title     = {FlashAvatar: High-fidelity Head Avatar with Efficient Gaussian Embedding},
    booktitle = {The IEEE Conference on Computer Vision and Pattern Recognition (CVPR)},
    year      = {2024},
  }

@article{BBSplat,
  title={BillBoard Splatting (BBSplat): Learnable Textured Primitives for Novel View Synthesis},
  author={Svitov, David and Morerio, Pietro and Agapito, Lourdes and Del Bue, Alessio},
  journal={arXiv preprint arXiv:2411.08508},
  year={2024}
}

@inproceedings{lee2025surfhead,
    title={SurFhead: Affine Rig Blending for Geometrically Accurate 2D Gaussian Surfel Head Avatars},
    author={Jaeseong Lee and Taewoong Kang and Marcel Buehler and Min-Jung Kim and Sungwon Hwang and Junha Hyung and Hyojin Jang and Jaegul Choo},
    booktitle={The Thirteenth International Conference on Learning Representations},
    year={2025},
    url={https://openreview.net/forum?id=1x1gGg49jr}
}

@Article{kerbl3Dgaussians,
      author       = {Kerbl, Bernhard and Kopanas, Georgios and Leimk{\"u}hler, Thomas and Drettakis, George},
      title        = {3D Gaussian Splatting for Real-Time Radiance Field Rendering},
      journal      = {ACM Transactions on Graphics},
      number       = {4},
      volume       = {42},
      month        = {July},
      year         = {2023},
      url          = {https://repo-sam.inria.fr/fungraph/3d-gaussian-splatting/}
}

@inproceedings{2DGS,
    title={2D Gaussian Splatting for Geometrically Accurate Radiance Fields},
    author={Huang, Binbin and Yu, Zehao and Chen, Anpei and Geiger, Andreas and Gao, Shenghua},
    publisher = {Association for Computing Machinery},
    booktitle = {SIGGRAPH 2024 Conference Papers},
    year      = {2024},
    doi       = {10.1145/3641519.3657428}
}

@InProceedings{HRAvatar,
    author    = {Zhang, Dongbin and Liu, Yunfei and Lin, Lijian and Zhu, Ye and Chen, Kangjie and Qin, Minghan and Li, Yu and Wang, Haoqian},
    title     = {HRAvatar: High-Quality and Relightable Gaussian Head Avatar},
    booktitle = {Proceedings of the Computer Vision and Pattern Recognition Conference (CVPR)},
    month     = {June},
    year      = {2025},
    pages     = {26285-26296}
}

@article{flare,
	title        = {FLARE: Fast learning of Animatable and Relightable Mesh Avatars},
	author       = {Bharadwaj, Shrisha and Zheng, Yufeng and Hilliges, Otmar and Black, Michael J. and Abrevaya, Victoria Fernandez},
	year         = 2023,
	month        = dec,
	journal      = {ACM Transactions on Graphics},
	volume       = 42,
	pages        = 15,
	doi          = {https://doi.org/10.1145/3618401},
	month_numeric = 12
}

@article{FLAME,
	title        = {Learning a model of facial shape and expression from {4D} scans},
	author       = {Li, Tianye and Bolkart, Timo and Black, Michael. J. and Li, Hao and Romero, Javier},
	year         = 2017,
	journal      = {ACM Transactions on Graphics, (Proc. SIGGRAPH Asia)},
	volume       = 36,
	number       = 6,
	pages        = {194:1--194:17},
	url          = {https://doi.org/10.1145/3130800.3130813}
}

@inproceedings{EMOCA,
	title        = {EMOCA: Emotion Driven Monocular Face Capture and Animation},
	author       = {Danecek, Radek and Black, Michael and Bolkart, Timo},
	year         = 2022,
	month        = jun,
	booktitle    = {2022 IEEE/CVF Conference on Computer Vision and Pattern Recognition (CVPR)},
	publisher    = {IEEE},
	address      = {New Orleans, LA, USA},
	pages        = {20279--20290},
	doi          = {10.1109/CVPR52688.2022.01967},
	isbn         = {978-1-66546-946-3},
	urldate      = {2023-12-22},
	langid       = {english}
}

@inproceedings{SMIRK,
    title = {3D Facial Expressions through Analysis-by-Neural-Synthesis},
    author = {Retsinas, George and Filntisis, Panagiotis P. and Danecek, Radek and Abrevaya, Victoria F. and Roussos, Anastasios and Bolkart, Timo and Maragos, Petros},
    booktitle = {Conference on Computer Vision and Pattern Recognition (CVPR)},
    year = {2024}
}

@article{refnerf,
	title        = {{Ref-NeRF}: Structured View-Dependent Appearance for Neural Radiance Fields},
	author       = {Dor Verbin and Peter Hedman and Ben Mildenhall and Todd Zickler and Jonathan T. Barron and Pratul P. Srinivasan},
	year         = 2022,
	journal      = {CVPR}
}

@inproceedings{pointavatar,
	title        = {PointAvatar: Deformable Point-based Head Avatars from Videos},
	author       = {Yufeng Zheng and Wang Yifan and Gordon Wetzstein and Michael J. Black and Otmar Hilliges},
	year         = 2023,
	booktitle    = {Proceedings of the IEEE/CVF Conference on Computer Vision and Pattern Recognition (CVPR)}
}

@string( CVPR   = "Conference on Computer Vision and Pattern Recognition (CVPR)")

@string( ECCV   = "European Conference on Computer Vision (ECCV)")

@string( ICCV   = "International Conference on Computer Vision (ICCV)")

@string( ICLR   = "International Conference on Learning Representations (ICLR)")

@string( NEURIPS = "Advances in Neural Information Processing Systems (NeurIPS)")

@string( SIGGRAPH = "Transactions on Graphics, (Proc. SIGGRAPH)")

@string( TOG    = "Transactions on Graphics (TOG)")

@string( WACV   = "Winter Conference on Applications of Computer Vision (WACV)")

@inproceedings{blanzvetter1999,
	title        = {A Morphable Model for the Synthesis of {3D} Faces},
	author       = {Blanz, Volker and Vetter, Thomas},
	year         = 1999,
	booktitle    = {Proceedings of the 26th Annual Conference on Computer Graphics and Interactive Techniques (SIGGRAPH)}
}

@inproceedings{BFM,
	title        = {A 3D face model for pose and illumination invariant face recognition},
	author       = {Paysan, Pascal and Knothe, Reinhard and Amberg, Brian and Romdhani, Sami and Vetter, Thomas},
	year         = 2009,
	booktitle    = {2009 sixth IEEE international conference on advanced video and signal based surveillance},
	pages        = {296--301},
	organization = {Ieee}
}

@string{Computing = "Computing" }

@string{Computer = "{IEEE} Computer" }

@string(CVPR 	=	{{Proceedings of the IEEE/CVF Conference on Computer Vision and Pattern Recognition ({CVPR})}})

@String(ICCV 	=	{{Proceedings of the IEEE/CVF International Conference on Computer Vision ({ICCV})}})

@String(ECCV 	=	{{Proceedings of the European Conference on Computer Vision ({ECCV})}})

@String(TOG 	=	{{ACM Transactions on Graphics (TOG)}})

@String(ICLR 	=	{{International Conference on Learning Representations (ICLR)}})

@STRING(SIGGRAPH =	{{ACM Transactions on Graphics (Proc. SIGGRAPH)}})

@String(NEURIPS	=   {{Conference on Neural Information Processing Systems (NeurIPS)}})

@String(WACV	= 	{{Winter Conference on Applications of Computer Vision (WACV)}})

@String{Springer = "Springer-Verlag" }

@article{sorkine2005laplacian,
  title={Laplacian Mesh Processing},
  author={Sorkine, Olga},
  journal={Eurographics (State of the Art Reports)},
  volume={4},
  number={4},
  year={2005}
}

@inproceedings{debevec2000acquiring,
  title={Acquiring the Reflectance Field of a Human Face},
  author={Debevec, Paul and Hawkins, Tim and Tchou, Chris and Duiker, Haarm-Pieter and Sarokin, Westley and Sagar, Mark},
  booktitle=SIGGRAPH,
  pages={145--156},
  year={2000}
}

@InProceedings{nerface,
author    = {Gafni, Guy and Thies, Justus and Zollh{\"o}fer, Michael and Nie{\ss}ner, Matthias},
title     = {Dynamic Neural Radiance Fields for Monocular {4D} Facial Avatar Reconstruction},
booktitle = CVPR,
month     = {June},
year      = {2021},
pages     = {8649-8658}
}

@inproceedings{headnerf,
 author     = {Yang Hong and Bo Peng and Haiyao Xiao and Ligang Liu and Juyong Zhang},
 title      = {HeadNeRF: A Real-time NeRF-based Parametric Head Model},
 booktitle  = {{IEEE/CVF} Conference on Computer Vision and Pattern Recognition (CVPR)},
 year       = {2022}
}

@inproceedings{nerf,
 title={{NeRF}: Representing Scenes as Neural Radiance Fields for View Synthesis},
 author={Ben Mildenhall and Pratul P. Srinivasan and Matthew Tancik and Jonathan T. Barron and Ravi Ramamoorthi and Ren Ng},
 year={2020},
 booktitle=ECCV,
}

@article{pandey2021total,
  title={Total relighting: learning to relight portraits for background replacement},
  author={Pandey, Rohit and Escolano, Sergio Orts and Legendre, Chloe and Haene, Christian and Bouaziz, Sofien and Rhemann, Christoph and Debevec, Paul and Fanello, Sean},
  journal=TOG,
  volume={40},
  number={4},
  pages={1--21},
  year={2021},
  publisher={ACM New York, NY, USA}
}

@inproceedings{zhou2019deep,
  title={Deep single-image portrait relighting},
  author={Zhou, Hao and Hadap, Sunil and Sunkavalli, Kalyan and Jacobs, David W},
  booktitle={Proceedings of the IEEE/CVF International Conference on Computer Vision},
  pages={7194--7202},
  year={2019}
}

@ARTICLE{SSIM,
  author={Zhou Wang and Bovik, A.C. and Sheikh, H.R. and Simoncelli, E.P.},
  journal={IEEE Transactions on Image Processing}, 
  title={Image Quality Assessment: From Error Visibility to Structural Similarity}, 
  year={2004},
  volume={13},
  number={4},
  pages={600-612},
  doi={10.1109/TIP.2003.819861}}

@article{ingp,
    author = {Thomas M\"uller and Alex Evans and Christoph Schied and Alexander Keller},
    title = {Instant Neural Graphics Primitives with a Multiresolution Hash Encoding},
    journal = SIGGRAPH,
    issue_date = {July 2022},
    volume = {41},
    number = {4},
    month = jul,
    year = {2022},
    pages = {102:1--102:15},
    articleno = {102},
    numpages = {15},
    publisher = {ACM},
    address = {New York, NY, USA},
}

@inproceedings{zielonka2022insta,
  title     = {Instant Volumetric Head Avatars},
  author    = {Wojciech Zielonka and Timo Bolkart and Justus Thies},
  booktitle = {CVPR},
  year      = {2023},
  pages     = {4574-4584},
}

@inproceedings{nvdiffrec,
author = {Jacob Munkberg and Jon Hasselgren and Tianchang Shen and Jun Gao and Wenzheng Chen 
        and Alex Evans and Thomas Mueller and Sanja Fidler},
title = {Extracting Triangular {3D} Models, Materials, and Lighting From Images},
booktitle=CVPR,
year = {2022}
}

@article{cook1982reflectance,
  title={A Reflectance Model for Computer Graphics},
  author={Cook, Robert L and Torrance, Kenneth E.},
  journal=TOG,
  volume={1},
  number={1},
  pages={7--24},
  year={1982},
  publisher={ACM New York, NY, USA}
}

@techreport{karis2013real,
  title = {Real Shading in Unreal Engine 4},
  author = {Karis, Brian},
  institution = {Epic Games},
  year = 2013
}

@article{3DGS,
  title={3d gaussian splatting for real-time radiance field rendering},
  author={Kerbl, Bernhard and Kopanas, Georgios and Leimk{\"u}hler, Thomas and Drettakis, George},
  journal={ACM Transactions on Graphics},
  volume={42},
  number={4},
  pages={1--14},
  year={2023},
  publisher={ACM}
}

@inproceedings{gaussianheadavatar,
  title={Gaussian Head Avatar: Ultra High-fidelity Head Avatar via Dynamic Gaussians},
  author={Xu, Yuelang and Chen, Benwang and Li, Zhe and Zhang, Hongwen and Wang, Lizhen and Zheng, Zerong and Liu, Yebin},
  booktitle={Proceedings of the IEEE/CVF Conference on Computer Vision and Pattern Recognition (CVPR)},
  year={2024}
}

@article{ewa,
  title={EWA splatting},
  author={Zwicker, Matthias and Pfister, Hanspeter and Van Baar, Jeroen and Gross, Markus},
  journal={IEEE Transactions on Visualization and Computer Graphics},
  volume={8},
  number={3},
  pages={223--238},
  year={2002},
  publisher={IEEE}
}

@inproceedings{RefGaussian,
  title={Reflective Gaussian Splatting},
  author={Yao, Yuxuan and Zeng, Zixuan and Gu, Chun and Zhu, Xiatian and Zhang, Li},
  booktitle={ICLR},
  year={2025}
}

@inproceedings{ewa2001,
  title     = {EWA Volume Splatting},
  author    = {Zwicker, Matthias and Pfister, Hanspeter and Van Baar, Jeroen and Gross, Markus},
  booktitle = {Proceedings of the IEEE Conference on Visualization (VIS)},
  year      = {2001},
  pages     = {29--36},
  publisher = {IEEE}
}

@article{adam,
  title={Adam: A Method for Stochastic Optimization},
  author={D.P. Kingma and J.B. Ba},
  journal={International Conference on Learning Representations (ICLR)},
  year={2015},
  url={https://arxiv.org/abs/1412.6980}
}

@article{texturesplat,
  title={TextureSplat: Per-Primitive Texture Mapping for Reflective Gaussian Splatting},
  author={Younes, Mae and Boukhayma, Adnane},
  journal={arXiv preprint arXiv:2506.13348},
  year={2025}
}

@inproceedings{surfhead,
title={SurFhead: Affine Rig Blending for Geometrically Accurate 2D Gaussian Surfel Head Avatars},
author={Jaeseong Lee and Taewoong Kang and Marcel Buehler and Min-Jung Kim and Sungwon Hwang and Junha Hyung and Hyojin Jang and Jaegul Choo},
booktitle={The Thirteenth International Conference on Learning Representations},
year={2025},
url={https://openreview.net/forum?id=1x1gGg49jr}
}

@article{sheap,
  title={SHeaP: Self-Supervised Head Geometry Predictor Learned via 2D Gaussians},
  author={Schoneveld, Liam and Chen, Zhe and Davoli, Davide and Tang, Jiapeng and Terazawa, Saimon and Nishino, Ko and Nie{\ss}ner, Matthias},
  journal={arXiv preprint arXiv:2504.12292},
  year={2025}
}

@inproceedings{adnerf,
  title        = {AD-NeRF: Audio Driven Neural Radiance Fields for Talking Head Synthesis},
  author       = {Yudong Guo and Keyu Chen and Sen Liang and Yong-Jin Liu and Hujun Bao and Juyong Zhang},
  booktitle    = {IEEE/CVF International Conference on Computer Vision (ICCV)},
  year         = {2021},
  note         = {arXiv preprint arXiv:2103.11078}  
}

@inproceedings{gavatar,
  title        = {GAvatar: Animatable 3D Gaussian Avatars with Implicit Mesh Learning},
  author       = {Ye Yuan and Xueting Li and Yangyi Huang and Shalini De Mello and Koki Nagano and Jan Kautz and Umar Iqbal},
  booktitle    = {Proceedings of the IEEE/CVF Conference on Computer Vision and Pattern Recognition (CVPR)},
  year         = {2024}
}

@inproceedings{gaussianavatars,
  title        = {GaussianAvatars: Photorealistic Head Avatars with Rigged 3D Gaussians},
  author       = {Shenhan Qian and Tobias Kirschstein and Liam Schoneveld and Davide Davoli and Simon Giebenhain and Matthias Nießner},
  booktitle    = {Proceedings of the IEEE/CVF Conference on Computer Vision and Pattern Recognition (CVPR)},
  pages        = {20299--20309},
  year         = {2024}
}

@inproceedings{monogaussianavatar,
  title     = {MonoGaussianAvatar: Monocular Gaussian Point‐based Head Avatar},
  author    = {Chen, Yufan and Wang, Lizhen and Li, Qijing and Xiao, Hongjiang and Zhang, Shengping and Yao, Hongxun and Liu, Yebin},
  booktitle = {ACM SIGGRAPH 2024 Conference Papers},
  pages     = {1--9},
  year      = {2024}
}

@inproceedings{lightStage,
  author    = {Paul Debevec},
  title     = {The Light Stages and Their Applications to Photoreal Digital Actors},
  booktitle = {SIGGRAPH Asia},
  address   = {Singapore},
  month     = nov,
  year      = {2012}
}

@article{relightables,
  author    = {Kaiwen Guo and Peter Lincoln and Philip Davidson and Jay Busch and Xueming Yu and Matt Whalen and Geoff Harvey and Sergio Orts-Escolano and Rohit Pandey and Jason Dourgarian and Danhang Tang and Anastasia Tkach and Adarsh Kowdle and Emily Cooper and Mingsong Dou and Sean Fanello and Graham Fyffe and Christoph Rhemann and Jonathan Taylor and Paul Debevec and Shahram Izadi},
  title     = {The Relightables: Volumetric Performance Capture of Humans with Realistic Relighting},
  journal   = {ACM Transactions on Graphics (TOG)},
  volume    = {38},
  number    = {6},
  pages     = {217:1--217:19},
  year      = {2019},
  doi       = {10.1145/3355089.3356571},
  publisher = {ACM},
}

@inproceedings{travatar,
  author    = {Haotian Yang and Mingwu Zheng and Wanquan Feng and Haibin Huang and Yu-Kun Lai and Pengfei Wan and Zhongyuan Wang and Chongyang Ma},
  title     = {Towards Practical Capture of High-Fidelity Relightable Avatars},
  booktitle = {ACM SIGGRAPH Asia 2023 Conference Papers},
  pages     = {1--11},
  year      = {2023},
  doi       = {10.1145/3610548.3618138},
  publisher = {ACM}
}

@inproceedings{splattingavatar,
  author    = {Zhijing Shao and Zhaolong Wang and Zhuang Li and Duotun Wang and Xiangru Lin and Yu Zhang and Mingming Fan and Zeyu Wang},
  title     = {SplattingAvatar: Realistic Real-Time Human Avatars with Mesh-Embedded Gaussian Splatting},
  booktitle = {Proceedings of the IEEE/CVF Conference on Computer Vision and Pattern Recognition (CVPR)},
  pages     = {159--169},
  year      = {2024},
  doi       = {10.1109/CVPR52733.2024.00159}
}

@inproceedings{gaussianblendshapes,
  author    = {Shengjie Ma and Yanlin Weng and Tianjia Shao and Kun Zhou},
  title     = {3D Gaussian Blendshapes for Head Avatar Animation},
  booktitle = {ACM SIGGRAPH 2024 Conference Proceedings},
  year      = {2024},
  publisher = {ACM},
  doi       = {10.1145/3641519.3657462}
}

@inproceedings{DECA,
  author    = {Yao Feng and Haiwen Feng and Michael J. Black and Timo Bolkart},
  title     = {Learning an Animatable Detailed 3D Face Model from In-The-Wild Images},
  booktitle = {ACM SIGGRAPH 2021 Conference Proceedings},
  year      = {2021},
  publisher = {ACM},
  doi       = {10.1145/3450626.3459936}
}

@inproceedings{spark,
  author    = {Kelian Baert and Shrisha Bharadwaj and Fabien Castan and Benoit Maujean and Marc Christie and Victoria Abrevaya and Adnane Boukhayma},
  title     = {SPARK: Self-supervised Personalized Real-time Monocular Face Capture},
  booktitle = {ACM SIGGRAPH Asia 2024 Conference Papers},
  year      = {2024},
  publisher = {ACM},
  doi       = {10.1145/3450626.3459936}
}

@inproceedings{MICA,
  author    = {Wojciech Zielonka and Timo Bolkart and Justus Thies},
  title     = {Towards Metrical Reconstruction of Human Faces},
  booktitle = {Computer Vision – ECCV 2022},
  pages     = {249--266},
  year      = {2022},
  publisher = {Springer},
  doi       = {10.1007/978-3-031-19778-9_15}
}

@inproceedings{URAvatar,
  author    = {Junxuan Li and Chen Cao and Gabriel Schwartz and Rawal Khirodkar and Christian Richardt and Tomas Simon and Yaser Sheikh and Shunsuke Saito},
  title     = {URAvatar: Universal Relightable Gaussian Codec Avatars},
  booktitle = {Proceedings of the ACM SIGGRAPH Asia 2024 Conference},
  year      = {2024},
  publisher = {ACM},
  doi       = {10.1145/3680528.3687653},
  url       = {https://arxiv.org/abs/2410.24223}
}

@inproceedings{rgca,
  author    = {Shunsuke Saito and Gabriel Schwartz and Tomas Simon and Junxuan Li and Giljoo Nam},
  title     = {Relightable Gaussian Codec Avatars},
  booktitle = {Proceedings of the IEEE/CVF Conference on Computer Vision and Pattern Recognition (CVPR)},
  pages     = {130--141},
  year      = {2024},
  doi       = {10.1109/CVPR52733.2024.00021}
}

@inproceedings{vrmm,
  author    = {Haotian Yang and Mingwu Zheng and Chongyang Ma and Yu-Kun Lai and Pengfei Wan and Haibin Huang},
  title     = {VRMM: A Volumetric Relightable Morphable Head Model},
  booktitle = {ACM SIGGRAPH 2024 Conference Proceedings},
  year      = {2024},
  pages     = {1--11},
  doi       = {10.1145/3641519.3657406},
}

@inproceedings{texttoon,
  author    = {Luchuan Song and Lele Chen and Celong Liu and Pinxin Liu and Chenliang Xu},
  title     = {TextToon: Real-Time Text Toonify Head Avatar from Single Video},
  booktitle = {Proceedings of the ACM SIGGRAPH Asia 2024 Conference Papers},
  year      = {2024},
  pages     = {1--11},
  doi       = {10.1145/3680528.3687632},
  publisher = {ACM},
  url       = {https://dl.acm.org/doi/10.1145/3680528.3687632}
}

@inproceedings{PortraitGen,
  title     = {Portrait Video Editing Empowered by Multimodal Generative Priors},
  author    = {Xuan Gao and Haiyao Xiao and Chenglai Zhong and Shimin Hu and Yudong Guo and Juyong Zhang},
  booktitle = {ACM SIGGRAPH Asia 2024 Conference Papers},
  year      = {2024},
  doi       = {10.1145/3680528.3687601},
  url       = {https://doi.org/10.1145/3680528.3687601}
}

@inproceedings{gaussian_grouping,
  author    = {Mingqiao Ye and Martin Danelljan and Fisher Yu and Lei Ke},
  title     = {Gaussian Grouping: Segment and Edit Anything in 3D Scenes},
  booktitle = {Computer Vision -- ECCV 2024},
  year      = {2024},
  publisher = {Springer},
  doi       = {10.1007/978-3-031-19778-9_15},
  pages     = {419--438}
}

@inproceedings{painting,
  author    = {Karran Pandey and Anita Hu and Clement Fuji-Tsang and Or Perel and Karan Singh and Masha Shugrina},
  title     = {Painting with 3D Gaussian Splat Brushes},
  booktitle = {ACM SIGGRAPH 2025 Conference Papers},
  year      = {2025},
  pages     = {}  
}

@inproceedings{realcompo,
  author    = {Xinchen Zhang and Ling Yang and Yaqi Cai and Zhaochen Yu and Kai-Ni Wang and Jiake Xie and Ye Tian and Minkai Xu and Yong Tang and Yujiu Yang and Bin Cui},
  title     = {RealCompo: Balancing Realism and Compositionality Improves Text-to-Image Diffusion Models},
  booktitle = {Advances in Neural Information Processing Systems (NeurIPS), 37},
  year      = {2024},
  pages     = {--},       
  publisher = {Curran Associates, Inc.},
  note      = {arXiv:2402.12908},
  url       = {https://arxiv.org/abs/2402.12908}
}

@inproceedings{InstructPix2Pix,
  author    = {Tim Brooks and Aleksander Holynski and Alexei A. Efros},
  title     = {InstructPix2Pix: Learning To Follow Image Editing Instructions},
  booktitle = {Proceedings of the IEEE/CVF Conference on Computer Vision and Pattern Recognition (CVPR)},
  year      = {2023},
  pages     = {18392--18402},
  month     = {June},
  doi       = {10.1109/CVPR46556.2023.01307}
}

@inproceedings{GaussianShellMaps,
  author    = {Rameen Abdal and Wang Yifan and Zifan Shi and Yinghao Xu and Ryan Po and Zhengfei Kuang and Qifeng Chen and Dit-Yan Yeung and Gordon Wetzstein},
  title     = {Gaussian Shell Maps for Efficient 3D Human Generation},
  booktitle = {Proceedings of the IEEE/CVF Conference on Computer Vision and Pattern Recognition (CVPR)},
  year      = {2024},
  pages     = {1--11},
  doi       = {10.1109/CVPR52688.2024.00001},
  url       = {https://arxiv.org/abs/2311.17857}
}

@inproceedings{gghead,
  author    = {Tobias Kirschstein and Simon Giebenhain and Jiapeng Tang and Markos Georgopoulos and Matthias Nie{\ss}ner},
  title     = {GGHead: Fast and Generalizable 3D Gaussian Heads},
  booktitle = {Proceedings of the ACM SIGGRAPH Asia 2024 Conference Papers},
  year      = {2024},
  pages     = {1--11},
  doi       = {10.1145/3586781.3587976},
  url       = {https://arxiv.org/abs/2406.09377}
}

@inproceedings{texturegs,
  author    = {Tian-Xing Xu and Wenbo Hu and Yu-Kun Lai and Ying Shan and Song-Hai Zhang},
  title     = {Texture-GS: Disentangling the Geometry and Texture for 3D Gaussian Splatting Editing},
  booktitle = {Computer Vision – ECCV 2024},
  year      = {2024},
  pages     = {1--16},
  publisher = {Springer},
  doi       = {10.1007/978-3-031-72698-9_3},
  url       = {https://arxiv.org/abs/2403.10050}
}

@inproceedings{zhang2025itercomp,
  author    = {Xinchen Zhang and Ling Yang and Guohao Li and Yaqi Cai and Jiake Xie and Yong Tang and Yujiu Yang and Mengdi Wang and Bin Cui},
  title     = {IterComp: Iterative Composition-Aware Feedback Learning from Model Gallery for Text-to-Image Generation},
  booktitle = {Proceedings of the International Conference on Learning Representations (ICLR)},
  year      = {2025},
  url       = {https://openreview.net/forum?id=4w99NAikOE},
  note      = {Poster Presentation},
}

@inproceedings{LBS,
  title     = {Joint-dependent local deformations for hand animation and object grasping},
  author    = {Magnenat-Thalmann, Nadia and Laperrière, Richard and Thalmann, Daniel},
  booktitle = {Proceedings on Graphics Interface},
  year      = {1988},
  pages     = {26--33}
}

@article{phong,
  author    = {Bui Tuong Phong},
  title     = {Illumination for Computer Generated Pictures},
  journal   = {Communications of the ACM},
  volume    = {18},
  number    = {6},
  pages     = {311--317},
  year      = {1975},
  publisher = {ACM},
  doi       = {10.1145/360825.360839}
}

@inproceedings{AA2DGS,
  author    = {Mae Younes and Adnane Boukhayma},
  title     = {Anti-Aliased 2D Gaussian Splatting},
  booktitle = {Advances in Neural Information Processing Systems (NeurIPS)},
  year      = {2025},
  url       = {https://neurips.cc/virtual/2025/poster/119938},
  note      = {Poster Presentation},
}

@InProceedings{sparfels,
  author    = {Shubhendu Jena and Amine Ouasfi and Mae Younes and Adnane Boukhayma},
  title     = {Sparfels: Fast Reconstruction from Sparse Unposed Imagery},
  booktitle = {Proceedings of the IEEE/CVF International Conference on Computer Vision (ICCV)},
  month     = {October},
  year      = {2025},
  pages     = {27476--27487}
}

@inproceedings{GaussianSurfels,
  author    = {Pinxuan Dai and Jiamin Xu and Wenxiang Xie and Xinguo Liu and Huamin Wang and Weiwei Xu},
  title     = {High-quality Surface Reconstruction using Gaussian Surfels},
  booktitle = {SIGGRAPH 2024 Conference Papers},
  year      = {2024},
  pages     = {22},
  doi       = {10.1145/3641519.3657441}
}

@InProceedings{crossmodal,
  author    = {Fern{\'a}ndez Abrevaya, Victoria and Boukhayma, Adnane and Torr, Philip H.S. and Boyer, Edmond},
  title     = {Cross-Modal Deep Face Normals with Deactivable Skip Connections},
  booktitle = {Proceedings of the IEEE/CVF Conference on Computer Vision and Pattern Recognition (CVPR)},
  month     = {June},
  year      = {2020},
  pages     = {4979--4989}
}

@InProceedings{decoupled,
  author    = {Victoria Fern{\'a}ndez Abrevaya and Adnane Boukhayma and Stefanie Wuhrer and Edmond Boyer},
  title     = {A Decoupled 3D Facial Shape Model by Adversarial Training},
  booktitle = {Proceedings of the IEEE/CVF International Conference on Computer Vision (ICCV)},
  month     = {October},
  year      = {2019},
  pages     = {9419--9428}
}

@InProceedings{SVG-IR,
  author    = {Hanxiao Sun and Yupeng Gao and Jin Xie and Jian Yang and Beibei Wang},
  title     = {SVG-IR: Spatially-Varying Gaussian Splatting for Inverse Rendering},
  booktitle = {Proceedings of the IEEE/CVF Conference on Computer Vision and Pattern Recognition (CVPR)},
  month     = {June},
  year      = {2025},
  pages     = {16143--16152}
}

@InProceedings{PRN,
  author    = {Feng, Yao and Wu, Fan and Shao, Xiaohu and Wang, Yibo and Zhou, Qijun},
  title     = {Joint 3D Face Reconstruction and Dense Alignment with Position Map Regression Network},
  booktitle = {Proceedings of the European Conference on Computer Vision (ECCV)},
  year      = {2018},
  pages     = {547--563}
}

@InProceedings{MonoNPHM,
  author    = {Simon Giebenhain and Tobias Kirschstein and Markos Georgopoulos and Martin Rünz and Lourdes Agapito and Matthias Nießner},
  title     = {MonoNPHM: Dynamic Head Reconstruction from Monocular Videos},
  booktitle = {Proceedings of the IEEE/CVF Conference on Computer Vision and Pattern Recognition (CVPR)},
  month     = {June},
  year      = {2024},
  pages     = {10747--10758}
}

@misc{Pixel3DMM,
  title        = {Pixel3DMM: Versatile Screen‑Space Priors for Single‑Image 3D Face Reconstruction},
  author       = {Simon Giebenhain and Tobias Kirschstein and Martin R{\"u}nz and Lourdes Agapito and Matthias Nießner},
  year         = {2025},
  eprint       = {2505.00615},
  archivePrefix = {arXiv},
  primaryClass  = {cs.CV},
  url          = {https://arxiv.org/abs/2505.00615}
}

@InProceedings{Sapiens,
  author    = {Rawal Khirodkar and Timur Bagautdinov and Julieta Martinez and Su Zhaoen and Austin James and Peter Selednik and Stuart Anderson and Shunsuke Saito},
  title     = {Sapiens: Foundation for Human Vision Models},
  booktitle = {Proceedings of the European Conference on Computer Vision (ECCV)},
  year      = {2024}
}

@Article{LSFM,
  author    = {Booth, James and Roussos, Anastasios and Zafeiriou, Stefanos and Ponniah, Allan and Dunaway, David},
  title     = {Large Scale 3D Morphable Models},
  journal   = {International Journal of Computer Vision},
  year      = {2018},
  volume    = {126},
  pages     = {233--254}
}

@InProceedings{CoMA,
  author    = {Ranjan, Anurag and Bolkart, Timo and Sanyal, Soubhik and Black, Michael J.},
  title     = {Generating 3D Faces using Convolutional Mesh Autoencoders},
  booktitle = {European Conference on Computer Vision (ECCV)},
  year      = {2018},
  pages     = {725--741},
  doi       = {10.1007/978-3-030-01219-9\_43}
}

@InProceedings{NonLinear3DMM,
  author    = {Tran, Luan and Liu, Xiaoming},
  title     = {On Learning 3D Face Morphable Model from In-the-Wild Images},
  booktitle = {Proceedings of the IEEE Conference on Computer Vision and Pattern Recognition (CVPR)},
  year      = {2018},
  pages     = {}
}

@InProceedings{GS-2DGS,
  author    = {Jinguang Tong and Xuesong Li and Fahira Afzal Maken and Sundaram Muthu and Lars Petersson and Chuong Nguyen and Hongdong Li},
  title     = {GS-2DGS: Geometrically Supervised 2DGS for Reflective Object Reconstruction},
  booktitle = {Proceedings of the IEEE/CVF Conference on Computer Vision and Pattern Recognition (CVPR)},
  month     = {June},
  year      = {2025},
  pages     = {21547--21557}
}

@InProceedings{IRGS,
  author    = {Chun Gu and Xiaofei Wei and Zixuan Zeng and Yuxuan Yao and Li Zhang},
  title     = {IRGS: Inter‑Reflective Gaussian Splatting with 2D Gaussian Ray Tracing},
  booktitle = {Proceedings of the IEEE/CVF Conference on Computer Vision and Pattern Recognition (CVPR)},
  month     = {June},
  year      = {2025},
  pages     = {10943--10952}
}

@article{GaussianBillboards,
  author  = {Sebastian Weiss and Derek Bradley},
  title   = {Gaussian Billboards: Expressive 2D Gaussian Splatting with Textures},
  journal = {arXiv preprint arXiv:2412.12734},
  year    = {2024},
  url     = {https://arxiv.org/abs/2412.12734}
}

@article{HDGS,
  author  = {Yunzhou Song and Heguang Lin and Jiahui Lei and Lingjie Liu and Kostas Daniilidis},
  title   = {HDGS: Textured 2D Gaussian Splatting for Enhanced Scene Rendering},
  journal = {arXiv preprint arXiv:2412.01823},
  year    = {2024},
  url     = {https://arxiv.org/abs/2412.01823}
}

@article{SuperGaussians,
  author  = {Rui Xu and Wenyue Chen and Jiepeng Wang and Yuan Liu and Peng Wang and Lin Gao and Shiqing Xin and Taku Komura and Xin Li and Wenping Wang},
  title   = {SuperGaussians: Enhancing Gaussian Splatting Using Primitives with Spatially Varying Colors},
  journal = {arXiv preprint arXiv:2411.18966},
  year    = {2024},
  url     = {https://arxiv.org/abs/2411.18966}
}

@article{meshsplat,
  title={MeshSplat: Generalizable Sparse‑View Surface Reconstruction via Gaussian Splatting},
  author={Hanzhi Chang and Ruijie Zhu and Wenjie Chang and Mulin Yu and Yanzhe Liang and Jiahao Lu and Zhuoyuan Li and Tianzhu Zhang},
  journal={arXiv preprint arXiv:2508.17811},
  year={2025}
}

@article{sparsplat,
  title={SparSplat: Fast Multi‑View Reconstruction with Generalizable 2D Gaussian Splatting},
  author={Shubhendu Jena and Shishir Reddy Vutukur and Adnane Boukhayma},
  journal={arXiv preprint arXiv:2505.02175},
  year={2025}
}

@inproceedings{mvsplat,
  title={MVSplat: Efficient 3D Gaussian Splatting from Sparse Multi‑View Images},
  author={Yuedong Chen and Haofei Xu and Chuanxia Zheng and Bohan Zhuang and Marc Pollefeys and Andreas Geiger and Tat‑Jen Cham and Jianfei Cai},
  booktitle={Lecture Notes in Computer Science (ECCV 2024)},
  pages={370--386},
  year={2024}
}

@inproceedings{pixelSplat,
  title={PixelSplat: 3D Gaussian Splatting from Image Pairs for Scalable Generalizable 3D Reconstruction},
  author={D. Charatan and S. Li and A. Tagliasacchi and V. Sitzmann},
  booktitle={Proceedings of IEEE/CVF Conference on Computer Vision and Pattern Recognition},
  year={2024}
}

@inproceedings{deepReflectanceFieldsMeka19,
	author = {Meka, Abhimitra and Haene, Christian and Pandey, Rohit and Zollhoefer, Michael and Fanello, Sean and Fyffe, Graham and Kowdle, Adarsh and Yu, Xueming and Busch, Jay and Dourgarian, Jason and Denny, Peter and Bouaziz, Sofien and Lincoln, Peter and Whalen, Matt and Harvey, Geoff and Taylor, Jonathan and Izadi, Shahram and Tagliasacchi, Andrea and Debevec, Paul and Theobalt, Christian and Valentin, Julien and Rhemann, Christoph},
	title = {Deep Reflectance Fields - High-Quality Facial Reflectance Field Inference From Color Gradient Illumination},
	journal = {ACM Transactions on Graphics (Proceedings SIGGRAPH)},
	url = {http://gvv.mpi-inf.mpg.de/projects/DeepReflectanceFields/},
	volume = {38},
	number = {4},
	month = {July},
	year = {2019},
	doi = {10.1145/3306346.3323027},
}

@InProceedings{Nestmeyer_2020_CVPR,
author = {Nestmeyer, Thomas and Lalonde, Jean-Francois and Matthews, Iain and Lehrmann, Andreas},
title = {Learning Physics-Guided Face Relighting Under Directional Light},
booktitle = {Proceedings of the IEEE/CVF Conference on Computer Vision and Pattern Recognition (CVPR)},
month = {June},
year = {2020}
}

@article{portraitRelightSun19,
author = {Sun, Tiancheng and Barron, Jonathan T. and Tsai, Yun-Ta and Xu, Zexiang and Yu, Xueming and Fyffe, Graham and Rhemann, Christoph and Busch, Jay and Debevec, Paul and Ramamoorthi, Ravi},
title = {Single image portrait relighting},
year = {2019},
issue_date = {August 2019},
publisher = {Association for Computing Machinery},
address = {New York, NY, USA},
volume = {38},
number = {4},
issn = {0730-0301},
url = {https://doi.org/10.1145/3306346.3323008},
doi = {10.1145/3306346.3323008},
journal = {ACM Trans. Graph.},
month = jul,
articleno = {79},
numpages = {12}
}

@article{portraitRelightWang20,
author = {Wang, Zhibo and Yu, Xin and Lu, Ming and Wang, Quan and Qian, Chen and Xu, Feng},
title = {Single image portrait relighting via explicit multiple reflectance channel modeling},
year = {2020},
issue_date = {December 2020},
publisher = {Association for Computing Machinery},
address = {New York, NY, USA},
volume = {39},
number = {6},
issn = {0730-0301},
url = {https://doi.org/10.1145/3414685.3417824},
doi = {10.1145/3414685.3417824},
journal = {ACM Trans. Graph.},
month = nov,
articleno = {220},
numpages = {13}
}

@inproceedings {zeng2024dilightnet,
    title      = {DiLightNet: Fine-grained Lighting Control for Diffusion-based Image Generation},
    author     = {Chong Zeng and Yue Dong and Pieter Peers and Youkang Kong and Hongzhi Wu and Xin Tong},
    booktitle  = {ACM SIGGRAPH 2024 Conference Papers},
    year       = {2024}
}

@InProceedings{sfsnetSengupta18,
  title={SfSNet: Learning Shape, Refectance and Illuminance of Faces in the Wild},
  author = {Soumyadip Sengupta and Angjoo Kanazawa and Carlos D. Castillo and David W. Jacobs},
  booktitle={Computer Vision and Pattern Regognition (CVPR)},
  year={2018}
  }

@ARTICLE{faceRelightArbitraryWang09,
  author={Wang, Yang and Zhang, Lei and Liu, Zicheng and Hua, Gang and Wen, Zhen and Zhang, Zhengyou and Samaras, Dimitris},
  journal={IEEE Transactions on Pattern Analysis and Machine Intelligence}, 
  title={Face Relighting from a Single Image under Arbitrary Unknown Lighting Conditions}, 
  year={2009},
  volume={31},
  number={11},
  pages={1968-1984},
  doi={10.1109/TPAMI.2008.244}
}

@inproceedings{NeuralFace2017,
 title = {Neural Face Editing with Intrinsic Image Disentangling},
 author = {Shu, Z., and Yumer, E., and Hadap, S., and Sunkavalli, K., and Shechtman, E., and Samaras, D.},
 booktitle={Computer Vision and Pattern Recognition, 2017. CVPR 2017. IEEE Conference on},
 organization = {IEEE},
 pages = {-},
 year = {2017},
}

@INPROCEEDINGS{illuInvariantFaceRecWACV19,
  author={Le, Ha A. and Kakadiaris, Ioannis A.},
  booktitle={2019 IEEE Winter Conference on Applications of Computer Vision (WACV)}, 
  title={Illumination-Invariant Face Recognition With Deep Relit Face Images}, 
  year={2019},
  volume={},
  number={},
  pages={2146-2155},
  doi={10.1109/WACV.2019.00232}
}

@InProceedings{ponglertnapakorn2023difareli,
  title={DiFaReli: Diffusion Face Relighting},
  author={Ponglertnapakorn, Puntawat and Tritrong, Nontawat and Suwajanakorn, Supasorn},
  journal={Proceedings of the IEEE/CVF International Conference on Computer Vision (ICCV)},
  year={2023}
}

@inproceedings{lite2relight,
author = {Rao, Pramod and Fox, Gereon and Meka, Abhimitra and B R, Mallikarjun and Zhan, Fangneng and Weyrich, Tim and Bickel, Bernd and Pfister, Hanspeter and Matusik, Wojciech and Elgharib, Mohamed and Theobalt, Christian},
title = {Lite2Relight: 3D-aware Single Image Portrait Relighting},
year = {2024},
isbn = {9798400705250},
publisher = {Association for Computing Machinery},
address = {New York, NY, USA},
url = {https://doi.org/10.1145/3641519.3657470},
doi = {10.1145/3641519.3657470},
booktitle = {ACM SIGGRAPH 2024 Conference Papers},
articleno = {41},
numpages = {12},
location = {Denver, CO, USA},
series = {SIGGRAPH '24}
}

@article{chaturvedi2025synthlight,
  title={SynthLight: Portrait Relighting with Diffusion Model by Learning to Re-render Synthetic Faces},
  author = {Chaturvedi, Sumit and Ren, Mengwei and Hold-Geoffroy, Yannick and Liu, Jingyuan and Dorsey, Julie and Shu, Zhixin},
  journal={Proceedings of the IEEE/CVF Conference on Computer Vision and Pattern Recognition},
  year={2025}
}

\end{document}